\documentclass[11pt]{article}

\PassOptionsToPackage{table}{xcolor}

\usepackage[preprint]{acl}

\usepackage{times}
\usepackage{latexsym}
\usepackage{tikz}

\usetikzlibrary{arrows.meta,fit,calc,positioning}

\usepackage[T1]{fontenc}

\usepackage[utf8]{inputenc}

\usepackage{microtype}

\usepackage{inconsolata}
\usepackage{mwe}
\usepackage{amsmath}
\usepackage{amssymb}
\usepackage{graphicx}
\usepackage{booktabs}
\usepackage[table]{xcolor}
\usepackage{array}
\usepackage{algorithm}
\usepackage{algpseudocode}
\usepackage{makecell}

\newcommand{\model}{\textsc{LoRAWeaver }}
\usepackage{subcaption}
\usepackage{caption}

\title{LoRAWeaver: Directional Adapter Routing in LoRA Representation Space}

\author{
  \textbf{Inish Melroy D Souza},
  \textbf{Tom Jackson},
  \textbf{Vitor Castro},
  \textbf{Rebecca Higginson},
  \textbf{Andrea Soltoggio}
\\
  \small{
    \texttt{\{I.M.D-Souza, T.W.Jackson, V.M.Q.Castro, R.L.Higginson, A.Soltoggio\}@lboro.ac.uk}
  }
\\
  Loughborough University, UK
}

\begin{document}

\maketitle

\begin{abstract}
Large language models are expensive to update through full retraining. Parameter-efficient fine-tuning reduces the cost of adaptation, but repeated adaptation creates a growing pool of task-specific LoRA adapters. This pool complicates adapter selection during inference. Existing routing and encoder-based methods often require retraining when new adapters are added, which limits their use in dynamic settings. This paper treats adapter routing as a representation alignment problem in the adapter-output space. The proposed training-free method selects adapters by matching adapter-conditioned query representations against task prototypes using cosine similarity. Experiments with LLaMA-2-7B and LLaMA-2-13B on a 48-task FLAN-v2 benchmark and on out-of-distribution tasks show that the method outperforms training-free baselines in the cross-task setting and approaches a trained retrieval encoder without learning a router. An ablation confirms the benefit of directional alignment in the adapter-output space.
\end{abstract}

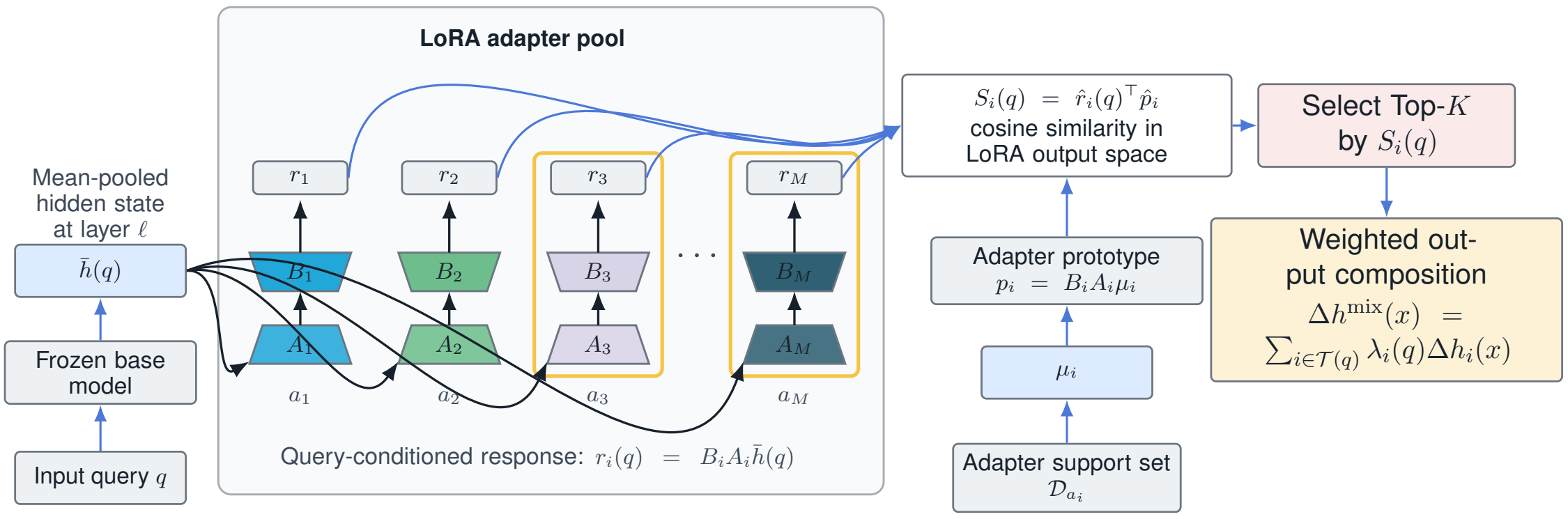
\begin{figure*}[t]
\centering
\resizebox{\textwidth}{!}{\definecolor{Ink}{HTML}{17212B}
\definecolor{Slate}{HTML}{3C4856}
\definecolor{SoftBlue}{HTML}{DCEBFF}
\definecolor{LineBlue}{HTML}{4C78D8}
\definecolor{AdapterA}{HTML}{22A7D9}
\definecolor{AdapterB}{HTML}{6DBE8C}
\definecolor{AdapterC}{HTML}{D8D4E8}
\definecolor{AdapterM}{HTML}{2F6073}
\definecolor{SoftGray}{HTML}{EEF1F4}
\definecolor{PanelGray}{HTML}{F8FAFC}
\definecolor{SelectGold}{HTML}{F6C445}
\definecolor{ScoreRed}{HTML}{D75F5F}

\tikzset{
  >=Latex,
  font=\sffamily,
  flow/.style={
    -{Latex[length=2.6mm,width=2.0mm]},
    draw=Ink,
    line width=0.9pt
  },
  softflow/.style={
    -{Latex[length=2.3mm,width=1.8mm]},
    draw=LineBlue,
    line width=0.85pt
  },
  panel/.style={
    rounded corners=5pt,
    draw=Slate!45,
    line width=0.8pt,
    fill=PanelGray
  },
  box/.style={
    rectangle,
    rounded corners=2pt,
    draw=Slate!75,
    fill=SoftGray,
    line width=0.75pt,
    inner sep=4pt,
    align=center,
    text=Ink
  },
  smallbox/.style={
    box,
    font=\sffamily\footnotesize,
    minimum width=1.30cm,
    minimum height=0.45cm
  },
  sourcebox/.style={
    box,
    font=\sffamily\small,
    minimum width=2.35cm,
    minimum height=0.72cm
  },
  scorebox/.style={
    box,
    fill=white,
    font=\sffamily\small,
    text width=3.4cm
  },
  note/.style={
    font=\sffamily\footnotesize,
    align=center,
    text=Slate
  },
  title/.style={
    font=\sffamily\bfseries\small,
    align=center,
    text=Ink
  }
}

\newcommand{\adapterstack}[5]{%
  \begin{scope}[shift={#1}]
    \path[draw=Slate!75, line width=0.8pt, fill=#5]
      (-0.70,1.05) -- (0.70,1.05) -- (0.52,0.52) -- (-0.52,0.52) -- cycle;
    \node[font=\sffamily\bfseries\footnotesize, text=Ink] at (0,0.78) {$B_{#2}$};
    \path[draw=Slate!75, line width=0.8pt, fill=#5!88]
      (-0.52,0.05) -- (0.52,0.05) -- (0.70,-0.48) -- (-0.70,-0.48) -- cycle;
    \node[font=\sffamily\bfseries\footnotesize, text=Ink] at (0,-0.22) {$A_{#2}$};
    \draw[flow] (0,0.05) -- (0,0.52);
    \draw[flow] (0,1.05) -- (0,1.78);
    \node[smallbox] (rvec#2) at (0,2.08) {$#3_{#2}$};
    \node[note] at (0,-0.95) {$a_{#2}$};
    \ifnum#4=1
      \draw[draw=SelectGold, line width=1.4pt, rounded corners=3pt]
        (-0.88,-0.65) rectangle (0.88,2.42);
    \fi
  \end{scope}%
}

\centering
\begin{tikzpicture}[x=1cm,y=1cm]

\node[sourcebox] (query) at (0,0) {Input query $q$};
\node[sourcebox, text width=2.35cm] (base) at (0,1.45) {Frozen base\\model};
\node[sourcebox, fill=SoftBlue] (hbar) at (0,2.85) {$\bar{h}(q)$};
\node[note, text width=3.0cm] at (0,3.75) {Mean-pooled hidden state\\at layer $\ell$};

\draw[softflow] (query) -- (base);
\draw[softflow] (base) -- (hbar);

\node[sourcebox, text width=2.85cm] (support) at (13.3,0) {Adapter support set\\$\mathcal{D}_{a_i}$};
\node[sourcebox, fill=SoftBlue] (mu) at (13.3,1.45) {$\mu_i$};
\node[sourcebox, text width=3.45cm] (proto) at (13.3,2.85)
  {Adapter prototype\\$p_i=B_iA_i\mu_i$};
\draw[softflow] (support) -- (mu);
\draw[softflow] (mu) -- (proto);

\node[panel, fit={(1.75,-0.10) (10.65,6.35)}] (library) {};
\node[title] at (5.85,6.02) {LoRA adapter pool
};

\adapterstack{(2.75,2.05)}{1}{r}{0}{AdapterA}
\adapterstack{(4.80,2.05)}{2}{r}{0}{AdapterB}
\adapterstack{(6.85,2.05)}{3}{r}{1}{AdapterC}
\node[font=\sffamily\bfseries\Large, text=Slate] at (8.25,3.05) {$\cdots$};
\adapterstack{(9.55,2.05)}{M}{r}{1}{AdapterM}

\draw[flow] (hbar.east) to[out=5,in=210] (2.05,1.60);
\draw[flow] (hbar.east) to[out=10,in=235] (4.10,1.60);
\draw[flow] (hbar.east) to[out=16,in=245] (6.15,1.60);
\draw[flow] (hbar.east) to[out=22,in=250] (8.85,1.60);

\node[note, text width=9.4cm] at (6.00,0.28)
  {Query-conditioned response: $r_i(q)=B_iA_i\bar{h}(q)$};

\node[scorebox, text width=4.25cm] (score) at (13.3,4.85)
  {$S_i(q)=\hat{r}_i(q)^\top\hat{p}_i$\\[2pt]
   cosine similarity in LoRA output space};
\draw[softflow] (proto.north) -- (score.south);

\draw[softflow] (rvec1.east) to[out=85,in=200] (score.west);
\draw[softflow] (rvec2.east) to[out=75,in=205] (score.west);
\draw[softflow] (rvec3.east) to[out=70,in=210] (score.west);
\draw[softflow] (rvecM.east) to[out=58,in=215] (score.west);

\node[box, fill=ScoreRed!13, text width=3.25cm] (topk) at (17.7,4.85)
  {Select Top-$K$\\by $S_i(q)$};
\draw[softflow] (score) -- (topk);

\node[box, fill=SelectGold!22, text width=4.55cm] (mix) at (17.7,2.45)
  {Weighted output composition\\[2pt]
   $\Delta h^{\mathrm{mix}}(x)=
   \sum_{i\in\mathcal{T}(q)}\lambda_i(q)\Delta h_i(x)$};
\draw[softflow] (topk) -- (mix);

\end{tikzpicture}}
\caption{LoRAWeaver. For each query, token-level activations are mean-pooled to form $\bar{h}$, which is passed through each candidate LoRA adapter to obtain query-conditioned responses. These responses are compared with precomputed adapter prototypes $p_i$, estimated by averaging representations from samples of the corresponding adapter's training dataset. Retrieval is performed independently at each layer, allowing layer-specific adapter selection. Gold outlines denote adapters selected by Top-$K$ retrieval.}
\label{fig:lora-retrieval}
\end{figure*}

\section{Introduction}

Large language models often require adaptation when tasks, domains, or data distributions change \cite{10.1145/3705725}. Training separate models for each setting increases computation and complexity, possibly resulting in inefficient resource utilisation, poor sustainability, and increased carbon emissions. With the progressive diffusion of AI systems and the resulting energy footprint, efficient model adaptation, and the sharing and reuse of learned knowledge, have become key research challenges \citep{soltoggio2024collective, bell2025future}. Parameter-efficient fine-tuning methods reduce the cost of adaptation by updating only a limited number of trainable parameters. Low-Rank Adaptation (LoRA) is one such method that learns low-rank updates for selected model weights and keeps the backbone network frozen \citep{hu2022lora}. Community platforms, including Hugging Face \citep{huggingface2025} and ModelScope \citep{modelscope2025}, now host large collections of task-specific LoRA adapters.  These collections, given heterogeneous inputs and an evolving adapter pool, require a mechanism for selecting and composing adapters without retraining the backbone. Consequently, the reuse, recycling, and merging of adapters and modular components has become an active area of research \citep{yadav2025tmlr-survey, liu2026appealrealityrecyclingloras, nath2026policy}.

Existing methods address adapter composition through static merging, supervised dynamic selection, or training-free selection. Static merging combines adapters offline into a single model, but the merged model uses the same adapter combination for all queries \citep{modelsoups, ilharco2023iclrediting, yadav2023ties, yu2024language, chronopoulou2023adaptersoup}. Supervised dynamic selection learns a router over experts, but changes to the adapter pool typically require router retraining \citep{wang2024loraflow, zhao2024loraretriever}. These limitations motivate training-free selection methods, which avoid offline merging and router retraining.

Prior training-free methods mainly score adapters either by the magnitude of the adapter response, or by similarity computed in the base-model representation space, before the query has been passed through any adapter. Arrow \cite{ostapenko2024arrow}, spectR \cite{fleshman2025spectr}, and LoGo \cite{lee2025logo} score adapters by magnitude or confidence signals such as activation norm (Arrow, spectR) or output entropy (LoGo), rather than direction.

This paper hypothesises that, in the adapter-transformed space, the quantity that separates a relevant adapter from an irrelevant one is the direction of the transformed query rather than the magnitude of the adapter response. Therefore, each adapter is summarised by a single prototype vector, which is computed by passing support examples from the adapter's training distribution through the adapter itself. At inference, every candidate adapter transforms the query, and the adapter whose transformed query aligns most closely with its prototype, measured by cosine similarity, is selected. No router is trained and no model is merged. The resulting training-free method is called LoRAWeaver. Figure \ref{fig:lora-retrieval} illustrates the LoRAWeaver routing method.

LoRAWeaver makes one adapter decision per layer for each input sequence, which prevents token-wise adapter changes during inference: within a layer, the selected adapter is fixed for all tokens of the sequence. Across layers the decisions are independent, so every layer selects the adapter whose prototype best matches the transformed query at that depth. Crucially, because earlier and later layers expose different representational structure, the selection adapts to the geometry available at each depth. In this formulation, each LoRA module serves two roles at once: as an adaptation mechanism and as a task-specific retrieval space.

LoRAWeaver is evaluated with LLaMA-2-7B and LLaMA-2-13B on a 48-task FLAN-v2 mixed-task benchmark and on out-of-distribution tasks spanning reasoning, knowledge, mathematics, and code (BBH, MMLU, GSM8K, MATH, AQuA, and CodeXGLUE).  In the cross-task setting, where each query's own task adapter is withheld, LoRAWeaver obtains the highest average performance among training-free methods on both backbones, suggesting the method retrieves adapters that transfer across tasks. The method's performance also approaches a trained retrieval encoder (LoRARetriever) without learning any router or merging any weights. The main hypothesis is validated with an ablation that isolates the scoring function and shows that directional similarity in the adapter-transformed space, rather than response magnitude, drives this performance. Together, these results establish it as an effective training-free retrieval method for multi-adapter inference.

\section{Related Work}

\paragraph{Static LoRA Composition.}

Static LoRA composition methods merge multiple adapters offline into a single set of weights that is then applied to every query \citep{modelsoups, ilharco2023iclrediting, yadav2023ties, yu2024language, chronopoulou2023adaptersoup, huang2024lorahub}. Because the merge is performed once and ahead of time, these methods add no routing cost at inference and require no per-query decision, which makes them simple to deploy. However, a single fused adapter assigns one mixture to inputs from every domain, averaging domain-specific behaviour and leaving no way to adjust the combination for an individual query.

\paragraph{Supervised LoRA Routing.}

Supervised dynamic routing methods use MoE-inspired routers, gates, retrievers, or classifiers to predict adapter fusion weights per query. LoRA-Flow~\cite{wang2024loraflow} learns layer-wise and token-wise fusion gates over predefined adapters. MeteoRA~\cite{xu2024meteora} uses an MoE-style gating network over task-specific LoRA experts. LoRARetriever~\cite{zhao2024loraretriever} retrieves adapters using a trained retriever model. MoLE~\cite{wu2024mole}, LoRAMoE~\cite{dou2024loramoe}, DLP-LoRA~\cite{zhang2024dlplora}, MixLoRA~\cite{li2024mixlora}, HydraLoRA~\cite{tian2024hydralora}, MoLA~\cite{gao2024mola}, and RAMoLE~\cite{zhao2024ramole} also use trained routing, classification, or gating modules for adapter selection and fusion. These methods provide query-level adaptation, but their routing modules are tied to a predefined adapter or domain set, so adding adapters to the library requires retraining or updating the router. PHATGOOSE~\cite{pmlr-v235-muqeeth24a} avoids this specific issue by training a lightweight gate independently per adapter, but still requires a trained gating vector for every adapter before it can be used.

\paragraph{Training-Free Routing Methods.}

Training-free routing methods select adapters without a learned router. qa-FLoRA~\cite{shukla2025qaflora} scores adapters layer-wise by computing the KL divergence between the adapter-conditioned and base model vocabulary distributions. Arrow~\cite{ostapenko2024arrow} and spectR~\cite{fleshman2025spectr} perform routing by maximising the norm of transformed LoRA activations. Similarly, LoGo~\cite{lee2025logo} utilises activation norms, alongside entropy-based scores from the final layer, to perform instance-level routing.

Although these methods differ, they share a common basis for selection: each ranks adapters by a magnitude or divergence quantity, such as activation norm, output entropy, weight geometry, or distributional distance. The direction of the adapter response in the adapter-output space, as implemented by LoRAWeaver, indicates whether an adapter moves the query towards its own training distribution, and to the best of our knowledge, has not been explored extensively in prior work.

\section{Method}

A pretrained language model is equipped with a library of LoRA adapters. LoRAWeaver retrieves adapters at inference time and composes them based on the input query. Given an input query, the frozen base model produces hidden representations at the target layers. For each candidate adapter, an adapter-induced response is estimated by projecting the hidden representation through the LoRA factors. A stored prototype represents each adapter. A similarity score between the query-conditioned response and the prototype defines the retrieval score. The top-ranked adapters are selected and combined at inference time. The following formulation is applied independently at each target layer.

\subsection{Problem Setup}
\label{sec:problem-setup}

Let $f_{\theta}$ denote a frozen pretrained language model with $L$ LoRA target modules, indexed by $\ell \in \{1,\dots,L\}$. A target module is a single projection to which adapters are attached, so $L$ counts modules rather than transformer blocks: attaching adapters to \texttt{q\_proj} and \texttt{v\_proj} in each of the 32 blocks of LLaMA-2-7B gives $L=64$ (Table~\ref{tab:implementation-details}). The term \emph{layer-wise} is used throughout for routing performed independently at each of these $L$ target modules, and $\ell$ is referred to as a layer for brevity. Let
\[
\mathcal{E} = \{a_1, a_2, \dots, a_M\}
\]
be the adapter library, a set of $M$ LoRA adapters, where each adapter $a$ is associated with a representative text set
$\mathcal{D}_a = \{x^{(a)}_1, \dots, x^{(a)}_{N_a}\}$ of $N_a$ examples.

At layer $\ell$, adapter $a$ induces the update
\[
\Delta W_{a,\ell} = s_{a,\ell} B_{a,\ell} A_{a,\ell},
\]
where
$A_{a,\ell} \in \mathbb{R}^{r \times d_{\mathrm{in}}}$,
$B_{a,\ell} \in \mathbb{R}^{d_{\mathrm{out}} \times r}$, and
$s_{a,\ell} = \alpha_{a,\ell}/r$.

\subsection{Activation Prototypes}
\label{sec:activation-prototypes}

For each adapter $a$, the method constructs a prototype for each layer $\ell$ in the LoRA output space from support examples $\mathcal{D}_a$. Let $H(x)$ denote the hidden states entering the target layer for an input $x$. The hidden states are pooled with the attention mask to obtain a sequence representation $\bar{h}_{\ell}(x) \in \mathbb{R}^{d_{\mathrm{in}}}$ which is then averaged across all representative examples of adapter $a$ to obtain $\mu_{a,\ell}$. The prototype is obtained by applying the LoRA update of adapter $a$ to the mean hidden representation $\mu_{a,\ell}$ and is normalised:

\begin{equation}
\hat{p}_{a,\ell}
=
\frac{p_{a,\ell}}{\lVert p_{a,\ell} \rVert_2}
=
\frac{B_{a,\ell} A_{a,\ell} \mu_{a,\ell}}
{\left\lVert B_{a,\ell} A_{a,\ell} \mu_{a,\ell} \right\rVert_2}.
\end{equation}
The scaling $s_{a,\ell}>0$ is omitted here and in Eq.~\eqref{eq:retrieval} because it is a positive scalar that cancels under $\ell_2$ normalisation; it therefore has no effect on the scores, the selected set, or the mixture weights, and enters only through $\Delta W_{a,\ell}$ at composition time.
The vector $\hat{p}_{a,\ell}$ represents the direction of the adapter update on examples from the adapter training distribution. The prototype lies in the LoRA output space, with one stored vector per adapter and target layer.

\subsection{Retrieval Procedure}
\label{sec:layerwise-retrieval}

At target layer $\ell$, the frozen base model yields the masked mean-pooled query representation $\bar{h}_{\ell}(q) \in \mathbb{R}^{d_{\mathrm{in}}}$. For each adapter $a$, the method computes
\begin{equation}
\begin{aligned}
r_{a,\ell}(q) &= B_{a,\ell} A_{a,\ell} \bar{h}_{\ell}(q), \\
S_{a,\ell}(q) &= \hat{r}_{a,\ell}(q)^\top \hat{p}_{a,\ell}.
\end{aligned}
\label{eq:retrieval}
\end{equation}
where $\hat{r}_{a,\ell}(q) = r_{a,\ell}(q)/\lVert r_{a,\ell}(q)\rVert_2$. The selected adapters at layer $\ell$ are
\begin{equation}
\mathcal{T}_{\ell}(q) = \operatorname*{TopK}_{a \in \mathcal{E}} S_{a,\ell}(q),
\end{equation}
where $\operatorname*{TopK}$ returns the $K$ highest-scoring adapters and $K \ll M$, and the process is repeated for each target layer $\ell$.

\subsection{LoRA module composition}

Retrieval is independent of the composition rule; any composition method (e.g., TIES \cite{yadav2023ties}, DARE \cite{yu2024language}, KnOTS \cite{knots}, LoRA-LEGO \cite{zhao2024mergingloras}) can be used, but a linear composition is used throughout to isolate the effect of retrieval. The same similarity score that drives selection also sets the mixture weights, so no additional scoring is introduced for composition.

Given $\mathcal{T}_{\ell}(q)$ and scores $S_{a,\ell}(q)$, nonnegative mixture weights are defined by a softmax with inverse temperature:
\begin{equation}
\lambda_{a,\ell}(q)
=
\frac{\exp\!\left(\tau S_{a,\ell}(q)\right)}
{\sum_{b \in \mathcal{T}_{\ell}(q)} \exp\!\left(\tau S_{b,\ell}(q)\right)},
\qquad a \in \mathcal{T}_{\ell}(q),
\end{equation}
where $\tau > 0$ is an inverse temperature.

Each adapter is applied independently to the layer input, and the resulting adapter outputs are combined. Writing $\Delta h_{a,\ell}(x) = \Delta W_{a,\ell}\,x$ for the update that adapter $a$ induces on a token representation $x$, the mixed update is
\begin{equation}
\Delta h^{\mathrm{mix}}_{\ell}(x)
=
\sum_{a \in \mathcal{T}_{\ell}(q)} \lambda_{a,\ell}(q)\,\Delta h_{a,\ell}(x).
\end{equation}
The query enters only through the selected set $\mathcal{T}_{\ell}(q)$ and the weights $\lambda_{a,\ell}(q)$, which are computed once per query and then held fixed while the mixed update is applied to every token.
Because each LoRA update is linear in the layer input, this output-space mixture is exactly equivalent to linearly merging the selected adapter updates into a single update $\sum_{a \in \mathcal{T}_{\ell}(q)} \lambda_{a,\ell}(q)\,\Delta W_{a,\ell}$. The implementation keeps the LoRA factors separate and evaluates the mixture in output space, so no dense merged weight is ever formed (Table \ref{tab:implementation-details}).

The output of layer $\ell$ for token $x$ is then computed as
\begin{equation}
h^{*}_{\ell}(x)
=
W_{\ell}x + \Delta h^{\mathrm{mix}}_{\ell}(x).
\end{equation}
The pseudocode of the complete LoRAWeaver algorithm is provided in Appendix \ref{app:algorithm-implementation-details}.

\section{Experimental Setup}

\newcolumntype{Y}{>{\centering\arraybackslash}m{0.9cm}}

\begin{table*}[t]
\centering
\resizebox{\textwidth}{!}{%
\begin{tabular}{l|>{\columncolor{gray!20}}Y>{\columncolor{blue!12}}YYYYYYYY|>{\columncolor{gray!20}}Y>{\columncolor{blue!12}}YYYYYYYY}
\toprule
Domain-Metric & \multicolumn{9}{c}{\textbf{LLaMA-2-7B}} & \multicolumn{9}{c}{\textbf{LLaMA-2-13B}} \\
 & Oracle & LW & LR & Adapter Soup & LoRA Hub & LoGo (Ent) & LoGo (Norm) & qa-FLoRA & spectR & Oracle & LW & LR & Adapter Soup & LoRA Hub & LoGo (Ent) & LoGo (Norm) & qa-FLoRA & spectR \\
\midrule
\multicolumn{19}{l}{\textbf{NLG Tasks Clusters}} \\
struct to text-rouge-1 & 63.4 & 44.8 & \textbf{53.2} & \underline{51.8} & 51.0 & 36.2 & 40.9 & 49.8 & 47.3 & 63.8 & 47.6 & \textbf{56.6} & 45.8 & \underline{54.6} & 46.6 & 46.6 & 42.3 & 44.8 \\
struct to text-rouge-2 & 38.9 & 22.5 & \textbf{29.0} & 27.2 & \underline{27.7} & 17.1 & 20.0 & 25.3 & 23.1 & 38.9 & 24.8 & \textbf{33.5} & 23.2 & \underline{30.6} & 22.7 & 22.9 & 20.1 & 21.6 \\
struct to text-rouge-l & 56.5 & 39.8 & \textbf{47.4} & \underline{45.7} & 44.6 & 32.9 & 36.3 & 43.6 & 41.5 & 57.1 & 41.1 & \textbf{50.4} & 40.2 & \underline{48.5} & 40.8 & 40.8 & 36.7 & 40.4 \\
translation-bleu & 13.0 & \underline{12.6} & 11.4 & \textbf{12.8} & 10.8 & 12.0 & 10.7 & 12.3 & 11.5 & 12.9 & 13.6 & 14.0 & 11.2 & 13.1 & \underline{14.1} & 13.9 & 12.6 & \textbf{14.6} \\
\midrule
\multicolumn{19}{l}{\textbf{NLU Tasks Clusters}} \\
closed\_book QA-em & 44.5 & \underline{48.0} & 31.5 & 37.5 & 37.0 & 47.5 & \textbf{51.5} & 43.0 & \textbf{51.5} & 56.5 & 62.5 & 50.0 & 47.5 & 49.0 & \underline{63.0} & \textbf{63.5} & 50.0 & 55.5 \\
commonsense-em & 63.0 & 59.5 & \textbf{62.5} & 50.5 & 56.0 & 60.0 & 59.0 & 53.5 & \underline{61.0} & 66.0 & \textbf{70.0} & \underline{66.0} & 58.0 & 65.5 & 64.0 & 63.5 & 59.5 & 58.5 \\
coreference-em & 53.0 & \textbf{63.0} & 60.0 & 52.0 & 59.0 & 60.0 & \underline{62.0} & 52.0 & 61.0 & 76.0 & \textbf{69.0} & \underline{68.0} & 61.0 & 43.0 & 58.0 & 58.0 & 47.0 & 50.0 \\
nli-em & 72.5 & \underline{64.2} & \textbf{65.3} & 54.6 & 57.9 & 54.2 & 55.8 & 55.0 & 56.9 & 76.7 & \textbf{74.9} & 71.4 & 60.4 & 67.7 & 66.0 & 66.0 & \underline{71.6} & 52.8 \\
paraphrase-em & 65.0 & \underline{59.0} & \textbf{63.5} & 45.0 & 40.0 & 40.5 & 43.0 & 45.5 & 50.5 & 74.5 & \textbf{61.5} & 55.5 & 54.5 & 42.0 & 42.5 & 41.0 & \underline{60.5} & 45.0 \\
RC w/ CS-em & 71.0 & 47.0 & \underline{60.0} & 29.0 & \textbf{66.0} & 40.0 & 44.0 & 36.0 & 42.0 & 81.0 & \textbf{69.0} & 32.0 & 44.0 & 48.0 & 59.0 & 62.0 & \underline{63.0} & 52.0 \\
reading comp-em & 67.7 & \textbf{53.0} & 52.0 & 50.7 & \underline{52.3} & 50.0 & 49.7 & 49.0 & 48.0 & 58.0 & \underline{58.0} & \textbf{66.3} & 49.7 & 53.3 & 49.0 & 48.7 & 47.0 & 44.7 \\
sentiment-em & 90.0 & 89.0 & \textbf{90.0} & 81.0 & \underline{89.5} & 79.0 & 79.5 & 72.5 & 84.0 & 91.0 & \underline{91.0} & \textbf{91.5} & \underline{91.0} & \underline{91.0} & 84.5 & 84.0 & 86.5 & 80.0 \\
\midrule
Macro Average & 58.2 & \underline{50.2} & \textbf{52.1} & 44.8 & 49.3 & 44.1 & 46.0 & 44.8 & 48.2 & 62.7 & \textbf{56.9} & \underline{54.6} & 48.9 & 50.5 & 50.9 & 50.9 & 49.7 & 46.7 \\
\bottomrule
\end{tabular}%
}
\caption{Average performance for each task cluster under the IID setting. IID denotes the setting in which routing methods have access to every LoRA for each test sample, including the task-specific LoRA for that sample. LW denotes LoRAWeaver, the proposed method, and LR denotes LoRARetriever \citep{zhao2024loraretriever}. Per-task results are reported in Appendix \ref{sec:per-task-performace}. The highest result is shown in \textbf{bold}, the second-highest result is \underline{underlined}, and the result of the proposed method is highlighted with a blue background.}
\label{tab:iid}
\end{table*}

\begin{table*}[t]
\centering
\resizebox{\textwidth}{!}{%
\begin{tabular}{l|>{\columncolor{gray!20}}Y>{\columncolor{blue!12}}YYYYYYY|>{\columncolor{gray!20}}Y>{\columncolor{blue!12}}YYYYYYY}
\toprule
Domain-Metric & \multicolumn{8}{c}{\textbf{LLaMA-2-7B}} & \multicolumn{8}{c}{\textbf{LLaMA-2-13B}} \\
 & Oracle & LW & LR & Adapter Soup & LoGo (Ent) & LoGo (Norm) & qa-FLoRA & spectR & Oracle & LW & LR & Adapter Soup & LoGo (Ent) & LoGo (Norm) & qa-FLoRA & spectR \\
\midrule
\multicolumn{17}{l}{\textbf{NLG Tasks Clusters}} \\
struct to text-rouge-1 & 63.4 & 41.0 & 40.4 & \textbf{46.0} & 36.0 & 41.0 & \underline{45.6} & 44.7 & 63.8 & \textbf{46.4} & 42.0 & 44.9 & 43.3 & 44.8 & \underline{45.8} & 39.3 \\
struct to text-rouge-2 & 38.9 & 19.3 & 19.7 & \textbf{23.2} & 16.9 & 20.2 & \textbf{23.2} & \underline{20.9} & 38.9 & \textbf{23.8} & 20.6 & \underline{23.2} & 20.7 & 21.8 & 22.5 & 18.4 \\
struct to text-rouge-l & 56.5 & 36.0 & 35.2 & \textbf{40.7} & 32.7 & 36.5 & \underline{40.2} & 39.5 & 57.1 & \textbf{40.5} & 37.2 & 39.3 & 38.1 & 39.3 & \underline{40.0} & 35.1 \\
translation-bleu & 13.0 & 11.5 & 11.3 & \textbf{12.5} & 11.6 & 10.4 & \underline{12.4} & 11.4 & 12.9 & 13.6 & 11.3 & 13.4 & \underline{14.2} & 14.0 & 12.2 & \textbf{14.5} \\
\midrule
\multicolumn{17}{l}{\textbf{NLU Tasks Clusters}} \\
closed\_book QA-em & 44.5 & 47.0 & 44.0 & 36.5 & 45.5 & \underline{49.5} & 39.5 & \textbf{50.5} & 56.5 & \textbf{63.0} & 46.0 & 57.0 & 62.0 & \underline{62.5} & 40.0 & 53.5 \\
commonsense-em & 63.0 & 58.5 & 55.5 & 50.0 & \underline{59.5} & 58.0 & 51.5 & \textbf{60.5} & 66.0 & \textbf{70.0} & \underline{64.5} & 59.5 & 64.0 & 63.5 & 52.0 & 57.0 \\
coreference-em & 53.0 & \textbf{64.0} & \underline{63.0} & 46.0 & 60.0 & 62.0 & 56.0 & 59.0 & 76.0 & \textbf{60.0} & \underline{59.0} & 55.0 & 58.0 & 58.0 & 50.0 & 51.0 \\
nli-em & 72.5 & \textbf{61.8} & \underline{57.7} & 55.6 & 54.2 & 55.8 & 53.6 & 57.1 & 76.7 & \textbf{73.9} & \underline{66.1} & 59.4 & 65.8 & 65.6 & 55.0 & 51.6 \\
paraphrase-em & 65.0 & \textbf{55.0} & 50.0 & 44.0 & 40.5 & 43.0 & 42.5 & \underline{50.5} & 74.5 & \textbf{59.5} & \underline{55.0} & 41.5 & 42.5 & 41.0 & 41.5 & 44.5 \\
RC w/ CS-em & 71.0 & \underline{39.0} & 37.0 & 18.0 & \underline{39.0} & \textbf{42.0} & 21.0 & 38.0 & 81.0 & \textbf{57.0} & 25.0 & 43.0 & 54.0 & \underline{55.0} & 18.0 & 46.0 \\
reading comp-em & 67.7 & \textbf{51.0} & 48.7 & 43.3 & \underline{50.3} & 49.0 & 43.3 & 47.0 & 58.0 & \textbf{54.7} & \underline{49.3} & \underline{49.3} & 48.7 & 48.3 & 45.0 & 42.3 \\
sentiment-em & 90.0 & \underline{88.5} & \textbf{90.0} & 80.0 & 78.5 & 79.5 & 75.5 & 84.0 & 91.0 & \textbf{91.0} & \underline{87.5} & 81.0 & 84.0 & 83.5 & 76.0 & 77.5 \\
\midrule
Macro Average & 58.2 & \textbf{47.7} & 46.0 & 41.3 & 43.7 & 45.6 & 42.0 & \underline{46.9} & 62.7 & \textbf{54.5} & 47.0 & 47.2 & 49.6 & \underline{49.8} & 41.5 & 44.2 \\
\bottomrule
\end{tabular}%
} 
\caption{Cross-task setting: task-specific LoRA for each test sample is masked during retrieval. Thus, each sample is routed only among LoRAs trained on other tasks. This setting evaluates whether a routing method can select adapters that transfer across tasks. LW denotes LoRAWeaver, the proposed method, and LR denotes LoRARetriever \citep{zhao2024loraretriever}. Per-task results are reported in Appendix \ref{sec:per-task-performace}.
}
\label{tab:crosstask}
\end{table*}

\newlength{\firstcolwidth}
\setlength{\firstcolwidth}{4.6cm}

\begin{table*}[t]
\centering
\resizebox{1.0\textwidth}{!}{%
\begin{tabular}{>{\centering\arraybackslash}p{\firstcolwidth}|Y>{\columncolor{blue!15}}YYYYYYY|Y>{\columncolor{blue!15}}YYYYYYY}
\toprule
& \multicolumn{8}{c|}{\textbf{LLaMA-2-7B}} & \multicolumn{8}{c}{\textbf{LLaMA-2-13B}} \\
\midrule
Task & Base & LW & LR & LoGo (Ent) & LoGo (Norm) & Adapter Soup & qa-FLoRA & spectR & Base & LW & LR & LoGo (Ent) & LoGo (Norm) & Adapter Soup & qa-FLoRA & spectR \\
\midrule
BBH & 39.30 & \textbf{40.50} & 36.28 & \underline{39.35} & 38.10 & 36.20 & 39.10 & 38.60 & 41.09 & 43.09 & \underline{45.05} & \textbf{45.76} & 45.70 & 41.03 & 43.00 & 35.15 \\
GSM8K & 0.00 & 3.94 & \textbf{4.62} & 3.87 & 3.87 & 2.27 & 4.02 & \underline{4.09} & 0.00 & \textbf{5.31} & \underline{4.62} & 2.50 & 2.12 & 3.79 & 2.73 & 3.87 \\
MATH & 0.00 & \textbf{3.82} & \underline{3.64} & 3.40 & 3.38 & 1.42 & 2.60 & 3.60 & 0.00 & \textbf{5.60} & \underline{4.74} & 3.60 & 3.70 & 4.72 & 3.70 & 4.50 \\
MMLU & 30.15 & 47.90 & 42.86 & \textbf{49.30} & \underline{49.10} & 42.30 & 46.71 & 47.90 & 38.07 & \textbf{56.41} & 55.37 & \underline{55.50} & 54.47 & 53.75 & 52.01 & 53.52 \\
AQuA & 16.54 & 23.62 & \textbf{26.77} & 22.83 & 22.44 & 20.47 & 18.13 & \underline{24.80} & 18.13 & \underline{25.59} & 24.02 & 22.44 & 24.02 & 21.65 & \textbf{26.77} & 22.05 \\
\bottomrule
\end{tabular}%
}
\caption{Performance on OOD tasks. Base denotes the frozen backbone evaluated with no LoRA adapter applied, LW denotes LoRAWeaver, the proposed method, and LR denotes LoRARetriever \citep{zhao2024loraretriever}.}
\label{tab:additional}
\end{table*}

\begin{table*}[t]
\centering
\resizebox{1.0\textwidth}{!}{%
\begin{tabular}{>{\centering\arraybackslash}p{\firstcolwidth}|Y>{\columncolor{blue!15}}YYYYYYY|Y>{\columncolor{blue!15}}YYYYYYY}
\toprule
& \multicolumn{8}{c|}{\textbf{LLaMA-2-7B}} & \multicolumn{8}{c}{\textbf{LLaMA-2-13B}} \\
\midrule
Task & Base & LW & LR & LoGo (Ent) & LoGo (Norm) & Adapter Soup & qa-FLoRA & spectR & Base & LW & LR & LoGo (Ent) & LoGo (Norm) & Adapter Soup & qa-FLoRA & spectR \\
\midrule
Code Refinement & 28.4 & 52.1 & \underline{54.8} & 51.6 & 50.7 & 46.2 & \textbf{64.3} & 52.3 & 23.7 & 40.9 & \underline{55.6} & 46.1 & 46.2 & 47.3 & \textbf{67.6} & 33.5 \\
Code Translation (Java $\rightarrow$ C\#) & 2.7 & \textbf{6.0} & 5.2 & 3.3 & 1.9 & 5.3 & \underline{5.9} & 2.9 & 3.6 & 6.0 & 5.5 & 5.0 & 5.0 & 5.6 & \underline{6.6} & \textbf{7.3} \\
Code Translation (C\# $\rightarrow$ Java) & 3.0 & \underline{8.6} & 7.0 & 7.6 & 5.2 & 8.2 & \textbf{9.1} & 5.6 & 5.2 & \textbf{11.0} & \underline{10.5} & 9.4 & 9.7 & 10.7 & 7.5 & 7.5 \\
Code-to-Text (Java) & 0.1 & \underline{0.6} & 0.5 & \textbf{0.7} & \textbf{0.7} & 0.3 & 0.3 & \underline{0.6} & 0.3 & \textbf{1.0} & \underline{0.8} & \textbf{1.0} & \textbf{1.0} & 0.6 & 0.3 & 0.2 \\
Code-to-Text (Python) & 15.2 & \textbf{50.8} & 18.8 & \textbf{50.8} & \textbf{50.8} & 43.2 & 19.9 & \textbf{50.8} & 44.1 & \underline{54.8} & \textbf{55.8} & 53.4 & 53.3 & 46.3 & 28.8 & 20.4 \\
\bottomrule
\end{tabular}%
}
\caption{Performance on the CodeXGLUE dataset. Base denotes the frozen backbone evaluated with no LoRA adapter applied, LW denotes LoRAWeaver, the proposed method, and LR denotes LoRARetriever \citep{zhao2024loraretriever}.}
\label{tab:code}
\end{table*}

\paragraph{Base Models and LoRA Adapters.}

Experiments use LLaMA-2-7B and LLaMA-2-13B as backbone models. For each backbone, 48 LoRA adapters are obtained from publicly available Hugging Face checkpoints released by \citet{zhao2024loraretriever}, in which rank $r$ is set to 6, and the scaling hyperparameter $\alpha$ is set to 12. The adapters, which were trained on a subset of FLAN-v2 dataset tasks \cite{wei2022finetuned, chung2024scaling} covering Natural Language Understanding (NLU) and Natural Language Generation (NLG), remain fixed during evaluation. Following the categorisation of \citet{wei2022finetuned}, the tasks are grouped into 10 clusters. Each adapter is trained using the Alpaca format \cite{alpaca}. The proposed method and the baselines are evaluated with each backbone and its corresponding adapters. Appendix~\ref{sec:eval_dataset} provides further configuration details on the adapters, clusters, and training data.

\paragraph{Datasets and Tasks.}

The evaluation covers mixed-task retrieval, reasoning transfer, and code transfer with LLaMA-2-7B and LLaMA-2-13B backbones. The mixed-task benchmark follows \citet{zhao2024loraretriever} and contains 2,395 test instances sampled from the 48 LoRA trained FLAN-v2 \cite{wei2022finetuned, chung2024scaling, pmlr-v202-longpre23a} tasks and shuffled into one test set. 

Results are evaluated under two retrieval configurations. In the first, the \textbf{IID} setting, retrieval is carried out over the entire LoRA pool, including each instance’s own source-task adapter. In the second, the \textbf{cross-task} setting, the source-task LoRA is excluded from retrieval, allowing evaluation of how well knowledge transfers across different task-specific LoRAs in the pool.

To test LoRAWeaver (LW) on harder OOD benchmarks, the following additional benchmarks are used: BIG-Bench Hard (BBH)~\citep{suzgun2023challenging}, MMLU~\citep{hendrycks2021measuringmassivemultitasklanguage}, GSM8K, MATH, and AQuA. BBH evaluation uses seven subtasks: Boolean Expressions, Causal Judgement, Dyck Languages, Formal Fallacies, Navigate, Object Counting, and Web of Lies. The evaluation further includes five out-of-domain CodeXGLUE \cite{lu2021codexgluemachinelearningbenchmark} tasks. The OOD tasks have no in-pool adapter, so performance measures whether FLAN-trained adapters transfer to unseen task types. Across all benchmarks, both mixed-task and OOD, evaluation is zero-shot unless stated otherwise.

\paragraph{Baselines.}
The Base baseline is the frozen backbone evaluated with no LoRA adapter applied, and it is reported in Tables~\ref{tab:additional}, \ref{tab:code}, and~\ref{tab:llama3} to show the performance available without any adapter. LoRARetriever (LR) \cite{zhao2024loraretriever} trains an encoder that embeds queries and LoRA adapters into a shared space and retrieves the nearest adapter. Adapter Soup \cite{chronopoulou2023adaptersoup}, in a modified form, retrieves a candidate set of adapters using the INSTRUCTOR-XL \citep{su2023instructor} encoder and fuses the retrieved adapters per query rather than merging them once per domain. LoRAHub \cite{huang2024lorahub} fits per-task merging coefficients on labelled target-task examples, so it is reported only in the IID setting, the only setting in which the evaluation task has an adapter, and hence target-task supervision, available in the pool. LoGo (Ent) \cite{lee2025logo} estimates adapter relevance from the inverse entropy of adapter activations at the final transformer layer, and LoGo (Norm) estimates the same quantity from adapter activation magnitude. qa-FLoRA~\citep{shukla2025qaflora} selects adapters by measuring layer-wise KL divergence over vocabulary logits. spectR \cite{fleshman2025spectr} extends Arrow \cite{ostapenko2024arrow} and applies a magnitude-based metric equal to the adapter output norm for token-level and layer-level routing. The oracle baseline uses the adapter trained on the evaluation task in the IID and cross-task settings. Appendix \ref{app:baseline-implementation} reports the implementation details of each baseline.

\paragraph{Implementation details of LoRAWeaver.} For each layer, prototypes are constructed using 20 samples per adapter, following the sample size choice used in LoRARetriever \cite{zhao2024loraretriever}. The number of LoRA adapters selected for composition, $K$, is set to 3 based on ablation studies. Appendix~\ref{app:algorithm-implementation-details} provides additional implementation details of LoRAWeaver.

\paragraph{Metrics.}
Following~\citet{wei2022finetuned} and~\citet{zhao2024loraretriever}, task performance is measured with task-specific metrics: Exact Match (EM) accuracy for NLU tasks and for BBH, MMLU, GSM8K, MATH, and AQuA; BLEU for translation and CodeXGLUE; and ROUGE-1, ROUGE-2, and ROUGE-L for structure-to-text. Each metric is computed independently on each task's predictions, and every domain-metric cell in Tables~\ref{tab:iid} and~\ref{tab:crosstask} is the unweighted mean of the per-task scores in that cluster rather than a corpus-level score over pooled predictions. The Macro Average is the unweighted mean over the 12 domain-metric rows, so struct-to-text contributes three rows and every other cluster one. Appendix~\ref{sec:per-task-performace} reports per-task results.

\section{Results}

Tables~\ref{tab:iid} and~\ref{tab:crosstask} report the results on the mixed-task benchmark. LoRAWeaver achieves high aggregate scores across both backbones. In the IID setting, LoRAWeaver achieves averages of 50.2 and 56.9 on LLaMA-2-7B and LLaMA-2-13B, respectively, versus 52.1 and 54.6 for LoRARetriever, despite not training a retrieval encoder.

In the cross-task setting, the task-specific LoRA is masked during retrieval. Under this constraint, LoRAWeaver achieves 47.7 on LLaMA-2-7B and 54.5 on LLaMA-2-13B, and outperforms all routing baselines. These results indicate that LoRAWeaver retrieves and combines adapters from related tasks when the source-task adapter is unavailable.

Table~\ref{tab:additional} reports results on the OOD tasks. LoRAWeaver is competitive with the best training-free method across reasoning and knowledge benchmarks. Table~\ref{tab:code} reports results on CodeXGLUE. The FLAN-trained pool contains no code adapter. LoRAWeaver achieves the highest or joint-highest score among methods on Java-to-C\# translation on LLaMA-2-7B, on C\#-to-Java translation on LLaMA-2-13B, and on Java code documentation on LLaMA-2-13B. On Python code summarisation on LLaMA-2-7B, LoRAWeaver matches the highest score. The code result indicates that retrieval over the FLAN-trained pool transfers to unseen code tasks.

Across the benchmark suite, LoRAWeaver is competitive with LoRARetriever without using a trained retrieval encoder, and leads it in the cross-task setting on both backbones. The result supports the view that the LoRA adapters themselves provide an embedding mechanism for retrieval. All per-task results are provided in Appendix \ref{sec:per-task-performace}.

\section{Analysis}

The following analyses isolate the components of LoRAWeaver's routing signal, test its transfer to a second backbone family, and quantify its cost.

\begin{figure}[t]
    \centering
    \includegraphics[height=0.6\columnwidth]{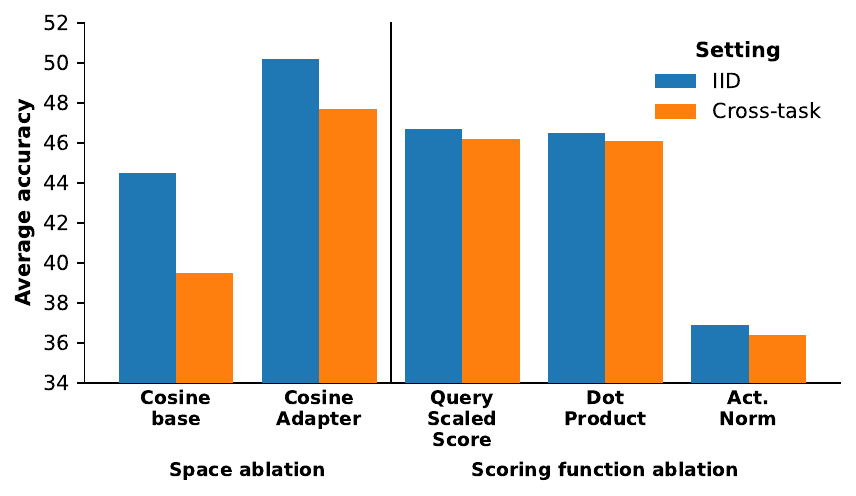}
    \caption{Ablation of retrieval representation space and scoring function on LLaMA-2-7B, evaluated on the mixed-task FLAN-v2 benchmark under both the IID and cross-task settings. Cosine similarity in the adapter-output space outperforms scoring in the base hidden-state space and alternative magnitude-based scores, indicating that adapter-output directions provide a more reliable signal for selecting the relevant LoRA adapter in both settings.}
    \label{fig:variants}
\end{figure}

\subsection{Retrieval scoring function and representation space}
\label{sec:retrieval-scoring}

LoRAWeaver (Cosine adapter) scores each LoRA adapter by cosine similarity between the query representation and the support-set prototype in the adapter-output space. Cosine base scores each LoRA adapter by cosine similarity in the base model hidden-state space. Cosine adapter outperforms Cosine base in both settings. On LLaMA-2-7B, Cosine adapter obtains macro averages of 47.7 on the cross-task benchmark and 50.2 on the IID benchmark, compared with 39.5 and 44.5 for Cosine base in Figure \ref{fig:variants}. These results indicate that adapter-output space contains more task-discriminative directional structure than base hidden-state space.

LoRAWeaver (Cosine adapter) is also compared with three adapter-output scoring alternatives: Query Scaled score, Dot product, and Act. Norm. Query Scaled score uses the dot product of the prototype and the unnormalised query vector. Dot product uses unnormalised query and prototype vectors. Act. Norm uses the adapter activation norm. Cosine adapter obtains the highest score in Figure \ref{fig:variants}. On the mixed-task FLAN-v2 benchmark in the IID setting with LLaMA-2-7B, it exceeds Query Scaled score by 3.5 points and Act. Norm by 13.3 points.

\begin{figure}[t]
    \centering
    \includegraphics[width=\columnwidth]{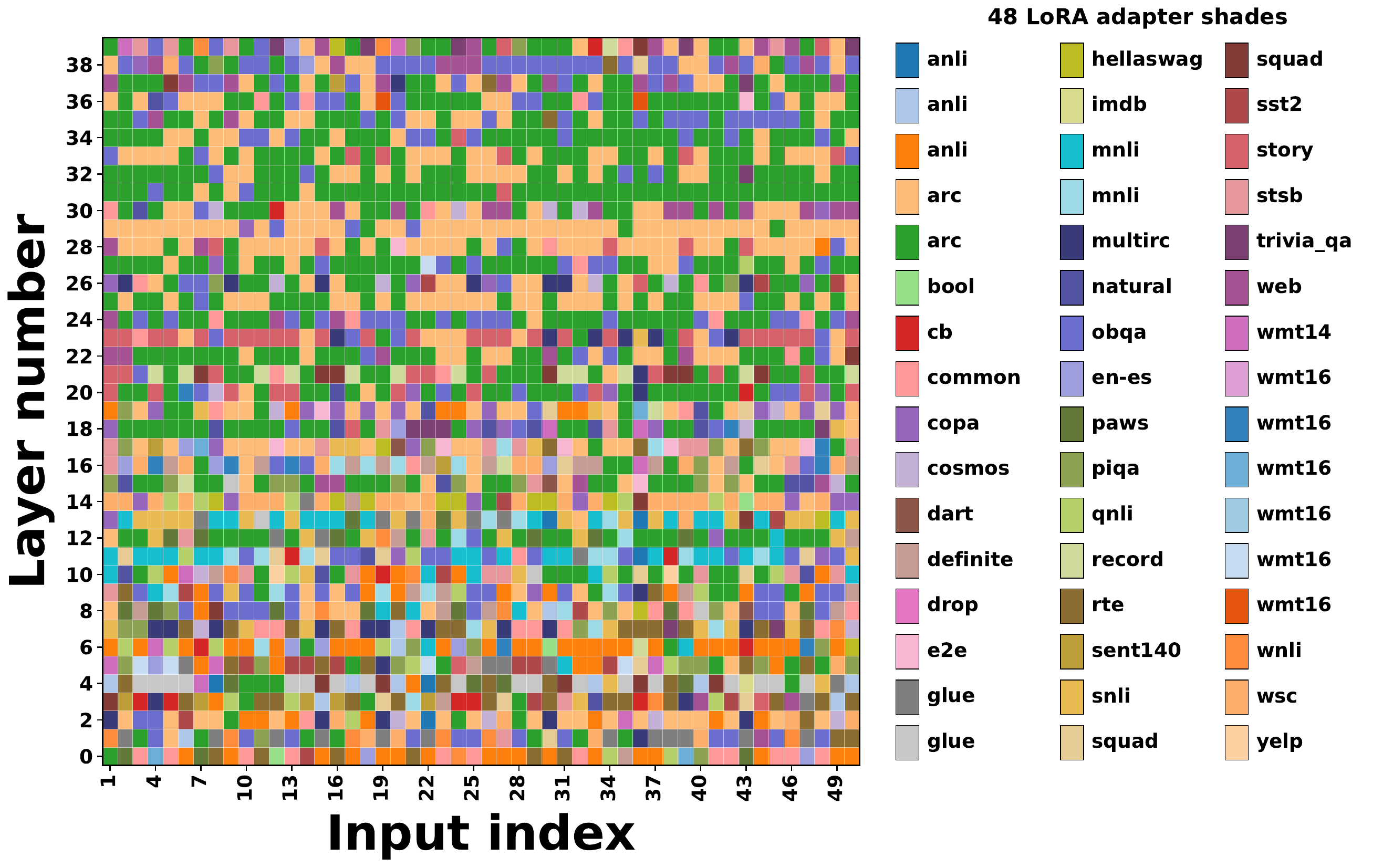}
    
\caption{Layer-wise adapter selection for ARC-Challenge on LLaMA-2-13B in the IID setting. Each column is a test input and each row is one transformer block, showing the top-1 adapter retrieved at that block's \texttt{q\_proj} target module; colours index the 48 candidate adapters. Rows are zero-indexed blocks $b \in \{0,\dots,39\}$, not the target-module index $\ell \in \{1,\dots,80\}$ of Section 3.1; \texttt{v\_proj} selections are omitted for legibility and follow the same pattern. Although the oracle adapter (green) is top-1 at many blocks, other blocks retrieve different adapters, and this diversity accompanies the gain over the oracle adapter alone.}
    \label{fig:layerwise}
\end{figure}

\begin{figure}[t]
    \centering
    \includegraphics[width=0.9\columnwidth]{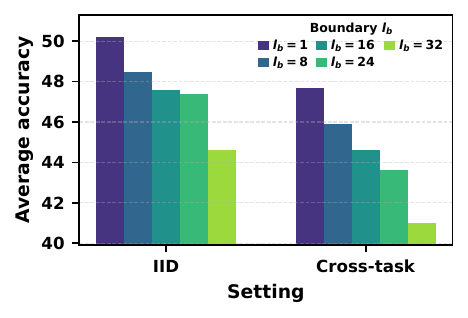}
    \caption{
Boundary-layer ablation on LLaMA-2-7B (32 layers), evaluated on the mixed-task FLAN-v2 benchmark under both the IID and cross-task settings. For a boundary layer $\ell_b$, layers before $\ell_b$ reuse the adapters selected at $\ell_b$, while later layers perform independent retrieval. Performance improves as the boundary moves earlier, showing that independent layer-wise retrieval contributes to final performance.
}
\label{fig:boundarylayer}
\end{figure}

\subsection{Role of independent layer-wise retrieval}
\label{sec:layerwise-retrieval-analysis}

LoRAWeaver retrieves LoRA adapters independently at each layer rather than selecting one global adapter set for the full network. An ablation evaluates the contribution of this layer-wise retrieval mechanism. Given a boundary layer $\ell_b$, all layers before $\ell_b$ use the adapters selected at $\ell_b$, and all layers after $\ell_b$ retain independent retrieval. When $\ell_b$ is the final layer, one adapter set is shared across all layers. This configuration gives the lowest performance. Moving $\ell_b$ towards earlier layers restores independent retrieval for more layers, which increases performance, as shown in Figure \ref{fig:boundarylayer}. The ablation indicates that a single shared adapter decision cannot preserve the information contained in the layer-wise LoRA output geometry.

For each layer $\ell$, a task-by-task similarity matrix is built from the sample-based LoRA prototypes defined in Section~\ref{sec:activation-prototypes}. Appendix \ref{sec:task-similarity-heatmaps} reports the resulting heatmaps in the LoRA output space for LLaMA-2-7B and LLaMA-2-13B. The heatmaps do not show stable task-level clusters across layers. Adapter prototypes are therefore not organised solely by high-level task identity.

The heatmaps indicate that LoRA adapter similarities vary across layers. Prior work \cite{belinkov2019bert,geva2021ffn,olah2021transformer} has shown that transformer layers encode different textual patterns, with earlier layers primarily representing lexical and local syntactic information and deeper layers representing semantic features and task-specific abstractions. This layer dependence provides a possible explanation for the benefit of independent layer-wise retrieval. LoRAWeaver matches each query to adapter prototypes separately in each layer's LoRA output space rather than using one global adapter decision. The boundary-layer ablation is consistent with this explanation: performance decreases when early layers reuse adapters selected at later layers. Adapter retrieval, therefore, depends on the geometry of the LoRA output space at each depth.

This routing mechanism may also explain the performance of LoRAWeaver over a single oracle adapter on specific tasks. For example, LoRAWeaver achieves a 4-point IID performance improvement over the oracle adapter on LLaMA-2-13B for the ARC-Challenge, from 66.0 to 70.0, a 6\% relative gain (Table \ref{tab:iidtask}). A colourmap analysis of the top-1 selected LoRA adapter illustrates this behaviour (Figure \ref{fig:layerwise}). The oracle adapter (green) activates in multiple layers for most inputs. Alternative layers retrieve different adapters. This result suggests that independent layer-wise retrieval contributes to LoRA routing effectiveness.

\subsection{Generalisation to a LLaMA-3.1-8B Backbone}
\label{sec:llama3}

The proposed routing mechanism is evaluated on LLaMA-3.1-8B to assess backbone generalisation. Constructing a second large adapter pool is computationally expensive. The evaluation therefore adopts a composite-task protocol inspired by qa-FLoRA \citep{shukla2025qaflora}. Unlike qa-FLoRA, which retains only the two task-specific adapters required for each evaluation instance and fuses both during inference, the present setting maintains a four adapter pool and applies routing to select two adapters used for composition. Four LoRA experts corresponding to Chinese (zh), Spanish (es), Russian (ru), and Math are trained on both LLaMA-2-7B and LLaMA-3.1-8B. Performance is evaluated on Multilingual Grade School Math (MGSM) \citep{shi2023language} using an 8-shot chain-of-thought prompting setup. Table~\ref{tab:llama3} reports the results. On LLaMA-3.1-8B, LoRAWeaver obtains the best training-free accuracy on MGSM-zh (64.00) and MGSM-es (71.60) and ties LoGo\,(Ent) on MGSM-ru (66.40). On LLaMA-2-7B it ties for best on MGSM-zh and MGSM-es and is second to spectR on MGSM-ru (17.60 versus 20.80). These observations indicate that the directional-similarity routing criterion transfers to a second backbone family. Additional details of the evaluation setup, training procedure, and implementation settings are reported in Appendix~\ref{app:llama3}.

\begin{table*}[t]
\centering
\resizebox{1.0\textwidth}{!}{%
\begin{tabular}{>{\centering\arraybackslash}p{\firstcolwidth}|Y>{\columncolor{blue!15}}YYYYYY|Y>{\columncolor{blue!15}}YYYYYY}
\toprule
& \multicolumn{7}{c|}{\textbf{LLaMA-3.1-8B}} & \multicolumn{7}{c}{\textbf{LLaMA-2-7B}} \\
\midrule
Task & Base & LW & Adapter Soup & LoGo (Ent) & LoGo (Norm) & qa-FLoRA & spectR & Base & LW & Adapter Soup & LoGo (Ent) & LoGo (Norm) & qa-FLoRA & spectR \\
\midrule
MGSM-zh & 32.40 & \textbf{64.00} & 53.20 & \underline{58.80} & 42.00 & 43.60 & 43.20 & 0.00 & \textbf{20.80} & \underline{18.80} & \textbf{20.80} & \underline{18.80} & 15.60 & 14.80 \\
MGSM-es & 44.00 & \textbf{71.60} & 58.80 & 67.60 & \underline{68.40} & 51.60 & 45.60 & 0.00 & \textbf{25.60} & \underline{24.40} & 23.20 & \textbf{25.60} & 23.60 & 14.40 \\
MGSM-ru & 15.60 & \textbf{66.40} & \underline{56.80} & \textbf{66.40} & 44.80 & 55.20 & 48.00 & 0.00 & \underline{17.60} & 15.20 & 16.80 & 15.20 & 16.40 & \textbf{20.80} \\
\bottomrule
\end{tabular}%
}
\caption{Multilingual grade-school math (MGSM) accuracy on a \emph{composite} four-adapter pool (zh, es, ru, Math), evaluated on LLaMA-2-7B and LLaMA-3.1-8B. This benchmark is distinct from the FLAN-v2 pool used in Tables~\ref{tab:iid}--\ref{tab:crosstask}. Base denotes the frozen backbone evaluated with no LoRA adapter applied, and LW denotes LoRAWeaver, the proposed method. LoRARetriever is omitted because it requires a trained retrieval encoder, which is unavailable for this backbone and pool.
}
\label{tab:llama3}
\end{table*}

\begin{figure}[t]
    \centering
    \includegraphics[width=\columnwidth]{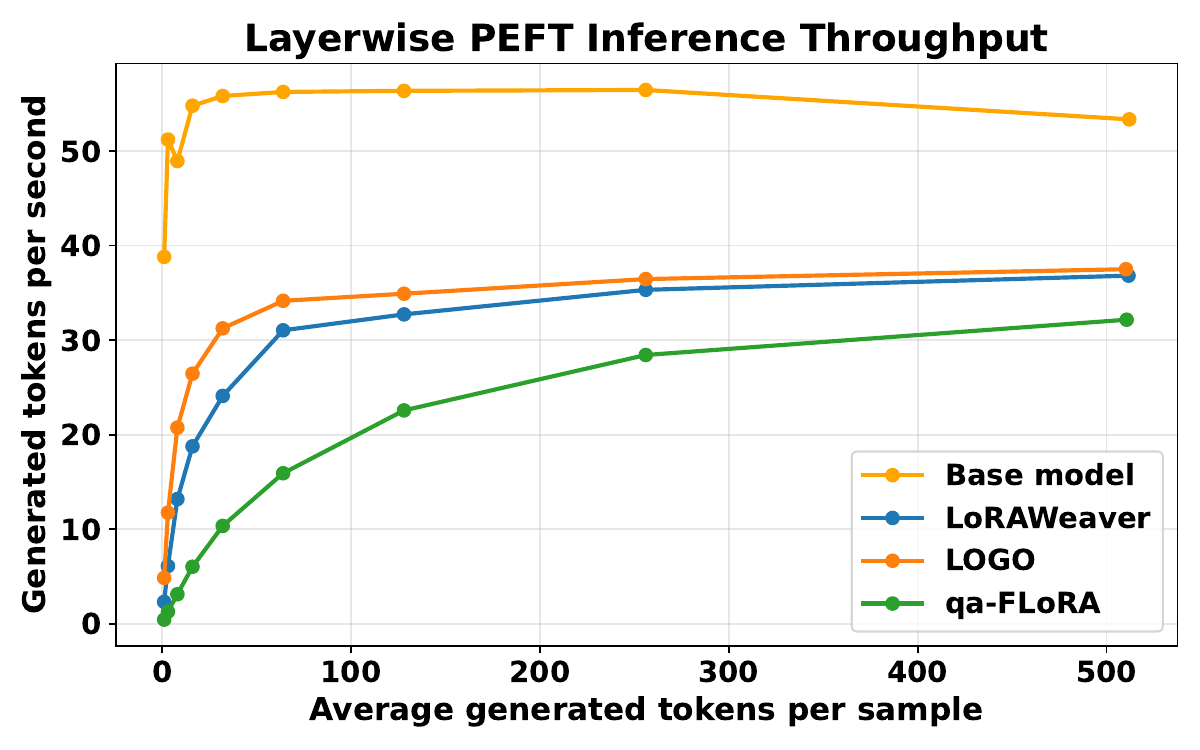}
    \caption{
Throughput comparison across routing methods. Query-level routing amortises the retrieval cost over the generated sequence.
}
    \label{fig:longGeneration}
\end{figure}

\subsection{Computational Efficiency}
\label{sec:inference-efficiency}

LoRAWeaver extracts its routing signal once per query, during a single probe pass, and reuses the selected adapters for the rest of decoding. Its routing cost is therefore fixed with respect to generation length and is amortised over the generated sequence. 
Figure \ref{fig:longGeneration} reports throughput on long-form generation (CNN-DailyMail \cite{nallapati-etal-2016-abstractive}, LLaMA-2-7B, single NVIDIA L40S). For short outputs, LoRAWeaver has lower throughput than the base model, because the fixed probe cost is spread over few tokens. As generation lengthens the per-token cost falls, throughput stabilises after roughly 120 generated tokens.

\subsection{Scalability Analysis}
\label{sec:scalability}

This section analyses the computational and memory costs of layer-wise routing with respect to the total adapter pool size \(M\). For each of the \(L\) probed LoRA target modules, LoRAWeaver ranks the candidate adapters and retains the top-\(K\) adapters, where \(K \ll M\). After selection, each target module uses its own retained adapter set. The post-selection composition and generation cost is therefore controlled by \(K\) per target module, rather than by the full adapter pool size.

The additional memory overhead comes from adapter parameters and probe-pass activations. Adapter-parameter memory grows linearly with the number of loaded adapters. Each LoRA target module of width \(d\) stores factors \(A \in \mathbb{R}^{r \times d}\) and \(B \in \mathbb{R}^{d \times r}\), that is \(2rd\) parameters, so an adapter covering all \(L\) target modules holds \(2rdL\) parameters. For the released LLaMA-2-7B LoRA adapters (\(r=6\), \(d=4096\), \(L=64\)), this gives \(2 \times 6 \times 4096 \times 64 \approx 3.15\)M parameters per adapter, equal to about 12.6 MB in fp32 or 6.3 MB in fp16/bf16. Loading all 48 adapters adds about 0.60 GB in fp32, excluding framework and allocator overhead. Probe activation memory depends on the number of probed target modules, sequence length, batch size, hidden size, and dtype. The activation term does not depend on \(M\) unless activations are stored separately for each adapter. During probing, KV caching is disabled to avoid unnecessary memory allocation.

Layer-wise selection computes a routing score for each candidate adapter at each probed target module. The ranking cost therefore scales linearly with \(M\) for each target module, and as \(O(LM)\) across the \(L\) probed target modules, where \(L\) is the number of LoRA target modules defined in Section~\ref{sec:problem-setup} (\(L=64\) for LLaMA-2-7B, i.e.\ \texttt{q\_proj} and \texttt{v\_proj} in each of the 32 blocks). The overhead is independent of generation length because it occurs before decoding. Subsequent generation uses only the selected top-\(K\) adapters at each target module. At an observed cost of 0.006s per adapter, probing 48 adapters adds approximately 0.29s per selection step on NVIDIA L40S GPU.

\section{Conclusion}
\label{sec:conclusion}

This paper introduced LoRAWeaver, a training-free method for LoRA adapter retrieval and composition. Each adapter is summarised by a single prototype in the adapter-output space, and a query is routed to the adapters whose transformed representation aligns most closely with that prototype by cosine similarity. Across mixed-task and out-of-distribution benchmarks on two backbones, LoRAWeaver leads training-free baselines in the cross-task setting and approaches a trained retrieval encoder without learning a router or merging weights. An ablation isolating the scoring function shows that direction in the adapter-output space, rather than response magnitude, drives this performance. A layer-wise analysis further shows that discriminative geometry differs across depth, supporting independent per-layer retrieval. Because selection requires no training, new adapters can be added to the pool and used online in settings with growing adapter libraries.

\section{Limitations}
\label{sec:limitations}

The proposed retrieval mechanism does not require sharing raw calibration text when only prototype vectors are shared, but the shared layer-wise prototypes remain data-dependent statistics. Each prototype encodes the direction of the LoRA-induced response to an average hidden representation, and therefore may preserve information about the calibration distribution. This dependency creates a potential attack surface for semantic inference, candidate matching, or other reconstruction-style analyses, particularly when prototypes are computed from few examples. Accordingly, the method is a sample-free sharing strategy, not a privacy-preserving one in the formal sense. In privacy-sensitive deployments, prototype release can be combined with safeguards such as minimum cohort sizes, per-example clipping, noisy aggregation, or differential privacy.

\section{Ethical Considerations}

The privacy implications of releasing adapter prototypes are a property of the method rather than a separate ethical concern, and are treated in Section~\ref{sec:limitations}. Two further considerations apply to deployment.

Reused adapters may carry biases or errors from the base model or adapter pool. Incorrect adapter selection can amplify these issues. Evaluation across relevant user groups, domains, and failure modes is necessary, particularly in high-stakes applications.

Adapter composition can support low-resource adaptation, but it may also facilitate misuse. Adapter libraries and released checkpoints require documentation covering training data, intended use, known limitations, and access restrictions where appropriate.

\section{Impact Statement}

This paper studies model adaptation through adapter retrieval instead of repeated retraining. The method restricts inference-time computation to a selected subset of LoRA adapters, which reduces compute use and model-management overhead in adapted models. These properties support deployment in resource-constrained and large-scale settings. The approach does not report energy or carbon savings, but it contributes to Digital Decarbonisation by suggesting an effective way to optimise computation while maintaining or improving performance in scenarios with dynamic pools of adapters. 

\section{LLM Usage}
Large language model (LLM) assistance was limited to linguistic refinement, grammar correction, readability revision, and exposition editing. LLMs did not generate, verify, or shape any scientific content, research ideas, methods, theoretical derivations, algorithms, experiments, or conclusions. All research contributions were conceived, developed, and executed by the authors. The authors retain full responsibility for the manuscript.

\section{Reproducibility Statement}

All datasets and LoRA adapters used in the experiments are publicly available. The code is available at:
\url{https://github.com/Inishpy/LoRAWeaver}

\bibliography{custom}

\begin{thebibliography}{47}
\providecommand{\natexlab}[1]{#1}

\bibitem[{Bell et~al.(2025)Bell, Quarantiello, Coleman, Li, Li, Madeddu,
  Piccoli, and Lomonaco}]{bell2025future}
Jack Bell, Luigi Quarantiello, Eric~Nuertey Coleman, Lanpei Li, Malio Li, Mauro
  Madeddu, Elia Piccoli, and Vincenzo Lomonaco. 2025.
\newblock \href {https://arxiv.org/abs/2506.03320} {The future of continual
  learning in the era of foundation models: Three key directions}.
\newblock In \emph{Trustworthy and Collaborative Artificial Intelligence
  Workshop, HHAI 2025}.

\bibitem[{Chronopoulou et~al.(2023)Chronopoulou, Peters, Fraser, and
  Dodge}]{chronopoulou2023adaptersoup}
Alexandra Chronopoulou, Matthew Peters, Alexander Fraser, and Jesse Dodge.
  2023.
\newblock \href {https://doi.org/10.18653/v1/2023.findings-eacl.153}
  {{A}dapter{S}oup: Weight averaging to improve generalization of pretrained
  language models}.
\newblock In \emph{Findings of the Association for Computational Linguistics:
  EACL 2023}, pages 2054--2063, Dubrovnik, Croatia. Association for
  Computational Linguistics.

\bibitem[{Chung et~al.(2024)Chung, Hou, Longpre, Zoph, Tay, Fedus, Li, Wang,
  Dehghani, Brahma, Webson, Gu, Dai, Suzgun, Chen, Chowdhery, Castro-Ros,
  Pellat, Robinson, Valter, Narang, Mishra, Yu, Zhao, Huang, Dai, Yu, Petrov,
  Chi, Dean, Devlin, Roberts, Zhou, Le, and Wei}]{chung2024scaling}
Hyung~Won Chung, Le~Hou, Shayne Longpre, Barret Zoph, Yi~Tay, William Fedus,
  Yunxuan Li, Xuezhi Wang, Mostafa Dehghani, Siddhartha Brahma, Albert Webson,
  Shixiang~Shane Gu, Zhuyun Dai, Mirac Suzgun, Xinyun Chen, Aakanksha
  Chowdhery, Alex Castro-Ros, Marie Pellat, Kevin Robinson, and 16 others.
  2024.
\newblock Scaling instruction-finetuned language models.
\newblock \emph{J. Mach. Learn. Res.}, 25(1).

\bibitem[{Dou et~al.(2024)Dou, Zhou, Liu, Gao, Shen, Xiong, Zhou, Wang, Xi,
  Fan, Pu, Zhu, Zheng, Gui, Zhang, and Huang}]{dou2024loramoe}
Shihan Dou, Enyu Zhou, Yan Liu, Songyang Gao, Wei Shen, Limao Xiong, Yuhao
  Zhou, Xiao Wang, Zhiheng Xi, Xiaoran Fan, Shiliang Pu, Jiang Zhu, Rui Zheng,
  Tao Gui, Qi~Zhang, and Xuanjing Huang. 2024.
\newblock \href {https://doi.org/10.18653/v1/2024.acl-long.106} {{L}o{RAM}o{E}:
  Alleviating world knowledge forgetting in large language models via
  {M}o{E}-style plugin}.
\newblock In \emph{Proceedings of the 62nd Annual Meeting of the Association
  for Computational Linguistics (Volume 1: Long Papers)}, pages 1932--1945,
  Bangkok, Thailand. Association for Computational Linguistics.

\bibitem[{Elhage et~al.(2021)Elhage, Nanda, Olsson, Henighan, Joseph, Mann,
  Askell, Bai, Chen, Conerly, DasSarma, Drain, Ganguli, Hatfield-Dodds,
  Hernandez, Jones, Kernion, Lovitt, Ndousse, Amodei, Brown, Clark, Kaplan,
  McCandlish, and Olah}]{olah2021transformer}
Nelson Elhage, Neel Nanda, Catherine Olsson, Tom Henighan, Nicholas Joseph, Ben
  Mann, Amanda Askell, Yuntao Bai, Anna Chen, Tom Conerly, Nova DasSarma, Dawn
  Drain, Deep Ganguli, Zac Hatfield-Dodds, Danny Hernandez, Andy Jones, Jackson
  Kernion, Liane Lovitt, Kamal Ndousse, and 6 others. 2021.
\newblock A mathematical framework for transformer circuits.
\newblock \emph{Transformer Circuits Thread}.
\newblock Https://transformer-circuits.pub/2021/framework/index.html.

\bibitem[{Fleshman and Van~Durme(2025)}]{fleshman2025spectr}
William Fleshman and Benjamin Van~Durme. 2025.
\newblock \href {https://doi.org/10.48550/ARXIV.2504.03454} {Spectr:
  Dynamically composing {LM} experts with spectral routing}.
\newblock \emph{CoRR}, abs/2504.03454.

\bibitem[{Gao et~al.(2025)Gao, Chen, Rao, Liu, Sun, Zhang, Peng, Guo, and
  Subrahmanian}]{gao2024mola}
Chongyang Gao, Kezhen Chen, Jinmeng Rao, Ruibo Liu, Baochen Sun, Yawen Zhang,
  Daiyi Peng, Xiaoyuan Guo, and Vs~Subrahmanian. 2025.
\newblock \href {https://doi.org/10.18653/v1/2025.findings-naacl.284}
  {{M}o{LA}: {M}o{E} {L}o{RA} with layer-wise expert allocation}.
\newblock In \emph{Findings of the Association for Computational Linguistics:
  NAACL 2025}, pages 5112--5127, Albuquerque, New Mexico. Association for
  Computational Linguistics.

\bibitem[{Geva et~al.(2021)Geva, Schuster, Berant, and Levy}]{geva2021ffn}
Mor Geva, Roei Schuster, Jonathan Berant, and Omer Levy. 2021.
\newblock \href {https://doi.org/10.18653/v1/2021.emnlp-main.446} {Transformer
  feed-forward layers are key-value memories}.
\newblock In \emph{EMNLP 2021 - 2021 Conference on Empirical Methods in Natural
  Language Processing, Proceedings}, EMNLP 2021 - 2021 Conference on Empirical
  Methods in Natural Language Processing, Proceedings, Online. Association for
  Computational Linguistics (ACL).

\bibitem[{Hendrycks et~al.(2021)Hendrycks, Burns, Basart, Zou, Mazeika, Song,
  and Steinhardt}]{hendrycks2021measuringmassivemultitasklanguage}
Dan Hendrycks, Collin Burns, Steven Basart, Andy Zou, Mantas Mazeika, Dawn
  Song, and Jacob Steinhardt. 2021.
\newblock \href {https://openreview.net/forum?id=d7KBjmI3GmQ} {Measuring
  massive multitask language understanding}.
\newblock In \emph{9th International Conference on Learning Representations,
  {ICLR} 2021, Virtual Event, Austria, May 3-7, 2021}. OpenReview.net.

\bibitem[{Hu et~al.(2022)Hu, Shen, Wallis, Allen{-}Zhu, Li, Wang, Wang, and
  Chen}]{hu2022lora}
Edward~J. Hu, Yelong Shen, Phillip Wallis, Zeyuan Allen{-}Zhu, Yuanzhi Li,
  Shean Wang, Lu~Wang, and Weizhu Chen. 2022.
\newblock \href {https://openreview.net/forum?id=nZeVKeeFYf9} {Lora: Low-rank
  adaptation of large language models}.
\newblock In \emph{The Tenth International Conference on Learning
  Representations, {ICLR} 2022, Virtual Event, April 25-29, 2022}.
  OpenReview.net.

\bibitem[{Huang et~al.(2024)Huang, Liu, Lin, Pang, Du, and
  Lin}]{huang2024lorahub}
Chengsong Huang, Qian Liu, Bill~Yuchen Lin, Tianyu Pang, Chao Du, and Min Lin.
  2024.
\newblock Lorahub: Efficient cross-task generalization via dynamic lora
  composition.
\newblock In \emph{Conference on Language Modeling (COLM)}.

\bibitem[{{Hugging Face}(2026)}]{huggingface2025}
{Hugging Face}. 2026.
\newblock Hugging face hub.
\newblock Accessed 2026-05-24.

\bibitem[{Ilharco et~al.(2023)Ilharco, Ribeiro, Wortsman, Schmidt, Hajishirzi,
  and Farhadi}]{ilharco2023iclrediting}
Gabriel Ilharco, Marco~Tulio Ribeiro, Mitchell Wortsman, Ludwig Schmidt,
  Hannaneh Hajishirzi, and Ali Farhadi. 2023.
\newblock \href {https://mlanthology.org/iclr/2023/ilharco2023iclr-editing/}
  {{Editing Models with Task Arithmetic}}.
\newblock In \emph{International Conference on Learning Representations}.

\bibitem[{Jawahar et~al.(2019)Jawahar, Sagot, and Seddah}]{belinkov2019bert}
Ganesh Jawahar, Beno{\^i}t Sagot, and Djam{\'e} Seddah. 2019.
\newblock \href {https://doi.org/10.18653/v1/P19-1356} {What does {BERT} learn
  about the structure of language?}
\newblock In \emph{Proceedings of the 57th Annual Meeting of the Association
  for Computational Linguistics}, pages 3651--3657, Florence, Italy.
  Association for Computational Linguistics.

\bibitem[{Lai et~al.(2023)Lai, Nguyen, Ngo, Nguyen, Dernoncourt, Rossi, and
  Nguyen}]{lai2023}
Viet Lai, Chien Nguyen, Nghia Ngo, Thuat Nguyen, Franck Dernoncourt, Ryan
  Rossi, and Thien Nguyen. 2023.
\newblock \href {https://doi.org/10.18653/v1/2023.emnlp-demo.28} {Okapi:
  Instruction-tuned large language models in multiple languages with
  reinforcement learning from human feedback}.
\newblock In \emph{Proceedings of the 2023 Conference on Empirical Methods in
  Natural Language Processing: System Demonstrations}, pages 318--327,
  Singapore. Association for Computational Linguistics.

\bibitem[{Lee et~al.(2026)Lee, Das, Gupta, and Gummadi}]{lee2025logo}
Seungeon Lee, Soumi Das, Manish Gupta, and Krishna~P. Gummadi. 2026.
\newblock \href {https://doi.org/10.18653/v1/2026.acl-long.1837} {{L}o{RA} on
  the go: Instance-level dynamic {L}o{RA} selection and merging}.
\newblock In \emph{Proceedings of the 64th Annual Meeting of the {A}ssociation
  for {C}omputational {L}inguistics (Volume 1: Long Papers)}, San Diego,
  California, United States. Association for Computational Linguistics.

\bibitem[{Li et~al.(2024)Li, Ma, Wang, Cheng, Duan, Zuo, Yang, and
  Tang}]{li2024mixlora}
Dengchun Li, Yingzi Ma, Naizheng Wang, Zhiyuan Cheng, Lei Duan, Jie Zuo, Cal
  Yang, and Mingjie Tang. 2024.
\newblock \href {https://doi.org/10.48550/ARXIV.2404.15159} {Mixlora: Enhancing
  large language models fine-tuning with lora based mixture of experts}.
\newblock \emph{CoRR}, abs/2404.15159.

\bibitem[{Liu et~al.(2026)Liu, Je, Ciccone, Xu, YSS, and
  Raffel}]{liu2026appealrealityrecyclingloras}
Haokun Liu, Gyung~Hyun Je, Marco Ciccone, Zhenlin Xu, Prasanth YSS, and Colin
  Raffel. 2026.
\newblock \href {https://arxiv.org/abs/2602.12323} {The appeal and reality of
  recycling loras with adaptive merging}.
\newblock \emph{Preprint}, arXiv:2602.12323.

\bibitem[{Longpre et~al.(2023)Longpre, Hou, Vu, Webson, Chung, Tay, Zhou, Le,
  Zoph, Wei, and Roberts}]{pmlr-v202-longpre23a}
Shayne Longpre, Le~Hou, Tu~Vu, Albert Webson, Hyung~Won Chung, Yi~Tay, Denny
  Zhou, Quoc~V. Le, Barret Zoph, Jason Wei, and Adam Roberts. 2023.
\newblock The flan collection: designing data and methods for effective
  instruction tuning.
\newblock In \emph{Proceedings of the 40th International Conference on Machine
  Learning}, ICML'23. JMLR.org.

\bibitem[{Lu et~al.(2021)Lu, Guo, Ren, Huang, Svyatkovskiy, Blanco, Clement,
  Drain, Jiang, Tang, Li, Zhou, Shou, Zhou, Tufano, Gong, Zhou, Duan,
  Sundaresan, Deng, Fu, and Liu}]{lu2021codexgluemachinelearningbenchmark}
Shuai Lu, Daya Guo, Shuo Ren, Junjie Huang, Alexey Svyatkovskiy, Ambrosio
  Blanco, Colin~B. Clement, Dawn Drain, Daxin Jiang, Duyu Tang, Ge~Li, Lidong
  Zhou, Linjun Shou, Long Zhou, Michele Tufano, Ming Gong, Ming Zhou, Nan Duan,
  Neel Sundaresan, and 3 others. 2021.
\newblock \href
  {https://datasets-benchmarks-proceedings.neurips.cc/paper/2021/hash/c16a5320fa475530d9583c34fd356ef5-Abstract-round1.html}
  {Codexglue: {A} machine learning benchmark dataset for code understanding and
  generation}.
\newblock In \emph{Proceedings of the Neural Information Processing Systems
  Track on Datasets and Benchmarks 1, NeurIPS Datasets and Benchmarks 2021,
  December 2021, virtual}.

\bibitem[{{ModelScope}(2026)}]{modelscope2025}
{ModelScope}. 2026.
\newblock \href {https://modelscope.cn} {Modelscope}.
\newblock Accessed: 2026-05-24.

\bibitem[{Muqeeth et~al.(2024)Muqeeth, Liu, Liu, and
  Raffel}]{pmlr-v235-muqeeth24a}
Mohammed Muqeeth, Haokun Liu, Yufan Liu, and Colin Raffel. 2024.
\newblock \href {https://proceedings.mlr.press/v235/muqeeth24a.html} {Learning
  to route among specialized experts for zero-shot generalization}.
\newblock In \emph{Proceedings of the 41st International Conference on Machine
  Learning}, volume 235 of \emph{Proceedings of Machine Learning Research}.
  PMLR.

\bibitem[{Nallapati et~al.(2016)Nallapati, Zhou, dos Santos,
  G{\"u}l{\c{c}}ehre, and Xiang}]{nallapati-etal-2016-abstractive}
Ramesh Nallapati, Bowen Zhou, Cicero dos Santos, {\c{C}}a{\u{g}}lar
  G{\"u}l{\c{c}}ehre, and Bing Xiang. 2016.
\newblock \href {https://doi.org/10.18653/v1/K16-1028} {Abstractive text
  summarization using sequence-to-sequence rnns and beyond}.
\newblock In \emph{Proceedings of the 20th SIGNLL Conference on Computational
  Natural Language Learning}, pages 280--290, Berlin, Germany. Association for
  Computational Linguistics.

\bibitem[{Nath et~al.(2026)Nath, Peridis, Benjamin, Liu, Kolouri, Kinnell, Li,
  Liu, Dora, and Soltoggio}]{nath2026policy}
Saptarshi Nath, Christos Peridis, Eseoghene Benjamin, Xinran Liu, Soheil
  Kolouri, Peter Kinnell, Zexin Li, Cong Liu, Shirin Dora, and Andrea
  Soltoggio. 2026.
\newblock \href {https://doi.org/10.1609/aaai.v40i29.39633} {Policy search,
  retrieval, and composition via task similarity in collaborative agentic
  systems}.
\newblock In \emph{Proceedings of the AAAI Conference on Artificial
  Intelligence}, volume~40, pages 24504--24512.

\bibitem[{Ostapenko et~al.(2024)Ostapenko, Su, Ponti, Charlin, Le~Roux, Caccia,
  and Sordoni}]{ostapenko2024arrow}
Oleksiy Ostapenko, Zhan Su, Edoardo Ponti, Laurent Charlin, Nicolas Le~Roux,
  Lucas Caccia, and Alessandro Sordoni. 2024.
\newblock \href {https://proceedings.mlr.press/v235/ostapenko24a.html} {Towards
  modular {LLM}s by building and reusing a library of {L}o{RA}s}.
\newblock In \emph{Proceedings of the 41st International Conference on Machine
  Learning}, volume 235 of \emph{Proceedings of Machine Learning Research},
  pages 38885--38904. PMLR.

\bibitem[{Shi et~al.(2023)Shi, Suzgun, Freitag, Wang, Srivats, Vosoughi, Chung,
  Tay, Ruder, Zhou, Das, and Wei}]{shi2023language}
Freda Shi, Mirac Suzgun, Markus Freitag, Xuezhi Wang, Suraj Srivats, Soroush
  Vosoughi, Hyung~Won Chung, Yi~Tay, Sebastian Ruder, Denny Zhou, Dipanjan Das,
  and Jason Wei. 2023.
\newblock \href {https://openreview.net/forum?id=fR3wGCk-IXp} {Language models
  are multilingual chain-of-thought reasoners}.
\newblock In \emph{The Eleventh International Conference on Learning
  Representations}.

\bibitem[{Shukla et~al.(2025)Shukla, Sriram, Narayanaswamy, and
  Jain}]{shukla2025qaflora}
Shreya Shukla, Aditya Sriram, Milinda~Kuppur Narayanaswamy, and Hiteshi Jain.
  2025.
\newblock \href {https://doi.org/10.48550/ARXIV.2512.11366} {qa-flora:
  Data-free query-adaptive fusion of loras for llms}.
\newblock \emph{CoRR}, abs/2512.11366.

\bibitem[{Soltoggio et~al.(2024)Soltoggio, Ben-Iwhiwhu, Braverman, Eaton,
  Epstein, Ge, Halperin, How, Itti, Jacobs, Kantharaju, Le, Lee, Liu, Monteiro,
  Musliner, Nath, Panda, Peridis, Pirsiavash, Parekh, Roy, Shperberg,
  Siegelmann, Stone, Vedder, Wu, Yang, Zheng, and
  Kolouri}]{soltoggio2024collective}
Andrea Soltoggio, Eseoghene Ben-Iwhiwhu, Vladimir Braverman, Eric Eaton,
  Benjamin Epstein, Yunhao Ge, Lucy Halperin, Jonathan How, Laurent Itti,
  Michael~A. Jacobs, Pavan Kantharaju, Long Le, Steven Lee, Xinran Liu,
  Sildomar~T. Monteiro, David Musliner, Saptarshi Nath, Priyadarshini Panda,
  Christos Peridis, and 11 others. 2024.
\newblock \href {https://doi.org/10.1038/s42256-024-00800-2} {A collective {AI}
  via lifelong learning and sharing at the edge}.
\newblock \emph{Nature Machine Intelligence}, 6(3):251--264.

\bibitem[{Stoica et~al.(2025)Stoica, Ramesh, Ecsedi, Choshen, and
  Hoffman}]{knots}
George Stoica, Pratik Ramesh, Boglarka Ecsedi, Leshem Choshen, and Judy
  Hoffman. 2025.
\newblock Model merging with svd to tie the knots.
\newblock \emph{ICLR}.

\bibitem[{Su et~al.(2023)Su, Shi, Kasai, Wang, Hu, Ostendorf, Yih, Smith,
  Zettlemoyer, and Yu}]{su2023instructor}
Hongjin Su, Weijia Shi, Jungo Kasai, Yizhong Wang, Yushi Hu, Mari Ostendorf,
  Wen-tau Yih, Noah~A. Smith, Luke Zettlemoyer, and Tao Yu. 2023.
\newblock \href {https://doi.org/10.18653/v1/2023.findings-acl.71} {One
  embedder, any task: Instruction-finetuned text embeddings}.
\newblock In \emph{Findings of the Association for Computational Linguistics:
  ACL 2023}, pages 1102--1121, Toronto, Canada. Association for Computational
  Linguistics.

\bibitem[{Suzgun et~al.(2023)Suzgun, Scales, Sch{\"a}rli, Gehrmann, Tay, Chung,
  Chowdhery, Le, Chi, Zhou, and Wei}]{suzgun2023challenging}
Mirac Suzgun, Nathan Scales, Nathanael Sch{\"a}rli, Sebastian Gehrmann, Yi~Tay,
  Hyung~Won Chung, Aakanksha Chowdhery, Quoc~V. Le, Ed~H. Chi, Denny Zhou, and
  Jason Wei. 2023.
\newblock \href {https://doi.org/10.18653/v1/2023.findings-acl.824}
  {Challenging {BIG-Bench} tasks and whether chain-of-thought can solve them}.
\newblock In \emph{Findings of the Association for Computational Linguistics:
  ACL 2023}, pages 13003--13051. Association for Computational Linguistics.

\bibitem[{Taori et~al.(2023)Taori, Gulrajani, Zhang, Dubois, Li, Guestrin,
  Liang, and Hashimoto}]{alpaca}
Rohan Taori, Ishaan Gulrajani, Tianyi Zhang, Yann Dubois, Xuechen Li, Carlos
  Guestrin, Percy Liang, and Tatsunori~B. Hashimoto. 2023.
\newblock Stanford alpaca: An instruction-following llama model.
\newblock \url{https://github.com/tatsu-lab/stanford_alpaca}.

\bibitem[{Tian et~al.(2024)Tian, Shi, Guo, Li, and Xu}]{tian2024hydralora}
Chunlin Tian, Zhan Shi, Zhijiang Guo, Li~Li, and Chengzhong Xu. 2024.
\newblock Hydralora: an asymmetric lora architecture for efficient fine-tuning.
\newblock In \emph{Proceedings of the 38th International Conference on Neural
  Information Processing Systems}, NIPS '24, Red Hook, NY, USA. Curran
  Associates Inc.

\bibitem[{Wang et~al.(2024)Wang, Ping, Wang, Han, Chen, Liu, and
  Sun}]{wang2024loraflow}
Hanqing Wang, Bowen Ping, Shuo Wang, Xu~Han, Yun Chen, Zhiyuan Liu, and Maosong
  Sun. 2024.
\newblock \href {https://doi.org/10.18653/v1/2024.acl-long.695} {{L}o{RA}-flow:
  Dynamic {L}o{RA} fusion for large language models in generative tasks}.
\newblock In \emph{Proceedings of the 62nd Annual Meeting of the Association
  for Computational Linguistics (Volume 1: Long Papers)}, pages 12871--12882,
  Bangkok, Thailand. Association for Computational Linguistics.

\bibitem[{Wei et~al.(2022)Wei, Bosma, Zhao, Guu, Yu, Lester, Du, Dai, and
  Le}]{wei2022finetuned}
Jason Wei, Maarten Bosma, Vincent~Y. Zhao, Kelvin Guu, Adams~Wei Yu, Brian
  Lester, Nan Du, Andrew~M. Dai, and Quoc~V. Le. 2022.
\newblock \href {https://openreview.net/forum?id=gEZrGCozdqR} {Finetuned
  language models are zero-shot learners}.
\newblock In \emph{The Tenth International Conference on Learning
  Representations, {ICLR} 2022, Virtual Event, April 25-29, 2022}.
  OpenReview.net.

\bibitem[{Wortsman et~al.(2022)Wortsman, Ilharco, Gadre, Roelofs,
  Gontijo-Lopes, Morcos, Namkoong, Farhadi, Carmon, Kornblith, and
  Schmidt}]{modelsoups}
Mitchell Wortsman, Gabriel Ilharco, Samir~Ya Gadre, Rebecca Roelofs, Raphael
  Gontijo-Lopes, Ari~S Morcos, Hongseok Namkoong, Ali Farhadi, Yair Carmon,
  Simon Kornblith, and Ludwig Schmidt. 2022.
\newblock \href {https://proceedings.mlr.press/v162/wortsman22a.html} {Model
  soups: averaging weights of multiple fine-tuned models improves accuracy
  without increasing inference time}.
\newblock In \emph{Proceedings of the 39th International Conference on Machine
  Learning}, volume 162 of \emph{Proceedings of Machine Learning Research},
  pages 23965--23998. PMLR.

\bibitem[{Wu et~al.(2024)Wu, Huang, and Wei}]{wu2024mole}
Xun Wu, Shaohan Huang, and Furu Wei. 2024.
\newblock \href {https://openreview.net/forum?id=uWvKBCYh4S} {Mixture of lora
  experts}.
\newblock In \emph{The Twelfth International Conference on Learning
  Representations, {ICLR} 2024, Vienna, Austria, May 7-11, 2024}.
  OpenReview.net.

\bibitem[{Xu et~al.(2024)Xu, Lai, and Huang}]{xu2024meteora}
Jingwei Xu, Junyu Lai, and Yunpeng Huang. 2024.
\newblock \href {https://arxiv.org/abs/2405.13053} {Meteora: Multiple-tasks
  embedded lora for large language models}.
\newblock \emph{Preprint}, arXiv:2405.13053.

\bibitem[{Yadav et~al.(2025)Yadav, Raffel, Muqeeth, Caccia, Liu, Chen, Bansal,
  Choshen, and Sordoni}]{yadav2025tmlr-survey}
Prateek Yadav, Colin Raffel, Mohammed Muqeeth, Lucas Caccia, Haokun Liu,
  Tianlong Chen, Mohit Bansal, Leshem Choshen, and Alessandro Sordoni. 2025.
\newblock \href {https://mlanthology.org/tmlr/2025/yadav2025tmlr-survey/} {{A
  Survey on Model MoErging: Recycling and Routing Among Specialized Experts for
  Collaborative Learning}}.
\newblock \emph{Transactions on Machine Learning Research}.

\bibitem[{Yadav et~al.(2023)Yadav, Tam, Choshen, Raffel, and
  Bansal}]{yadav2023ties}
Prateek Yadav, Derek Tam, Leshem Choshen, Colin Raffel, and Mohit Bansal. 2023.
\newblock Ties-merging: resolving interference when merging models.
\newblock In \emph{Proceedings of the 37th International Conference on Neural
  Information Processing Systems}, NIPS '23, Red Hook, NY, USA. Curran
  Associates Inc.

\bibitem[{Yang et~al.(2025)Yang, Zhou, Ding, Huai, Liu, Chen, Xie, and
  He}]{10.1145/3705725}
Yutao Yang, Jie Zhou, Xuanwen Ding, Tianyu Huai, Shunyu Liu, Qin Chen, Yuan
  Xie, and Liang He. 2025.
\newblock \href {https://doi.org/10.1145/3705725} {Recent advances of
  foundation language models-based continual learning: A survey}.
\newblock \emph{ACM Comput. Surv.}, 57(5).

\bibitem[{Yu et~al.(2024{\natexlab{a}})Yu, Yu, Yu, Huang, and
  Li}]{yu2024language}
Le~Yu, Bowen Yu, Haiyang Yu, Fei Huang, and Yongbin Li. 2024{\natexlab{a}}.
\newblock \href {https://proceedings.mlr.press/v235/yu24p.html} {Language
  models are super mario: Absorbing abilities from homologous models as a free
  lunch}.
\newblock In \emph{Proceedings of the 41st International Conference on Machine
  Learning}, volume 235 of \emph{Proceedings of Machine Learning Research},
  pages 57755--57775. PMLR.

\bibitem[{Yu et~al.(2024{\natexlab{b}})Yu, JIANG, Shi, YU, Liu, Zhang, Kwok,
  Li, Weller, and Liu}]{yu2023metamath}
Longhui Yu, Weisen JIANG, Han Shi, Jincheng YU, Zhengying Liu, Yu~Zhang, James
  Kwok, Zhenguo Li, Adrian Weller, and Weiyang Liu. 2024{\natexlab{b}}.
\newblock \href
  {https://proceedings.iclr.cc/paper_files/paper/2024/file/c400474e8a36d0812fdee52739288b12-Paper-Conference.pdf}
  {Metamath: Bootstrap your own mathematical questions for large language
  models}.
\newblock In \emph{International Conference on Learning Representations}.

\bibitem[{Zhang and Li(2024)}]{zhang2024dlplora}
Yuxuan Zhang and Ruizhe Li. 2024.
\newblock \href {https://doi.org/10.48550/ARXIV.2410.01497} {Dlp-lora:
  Efficient task-specific lora fusion with a dynamic, lightweight plugin for
  large language models}.
\newblock \emph{CoRR}, abs/2410.01497.

\bibitem[{Zhao et~al.(2024{\natexlab{a}})Zhao, Gan, Wang, Hu, Shen, Yang,
  Kuang, and Wu}]{zhao2024ramole}
Ziyu Zhao, Leilei Gan, Guoyin Wang, Yuwei Hu, Tao Shen, Hongxia Yang, Kun
  Kuang, and Fei Wu. 2024{\natexlab{a}}.
\newblock \href {https://arxiv.org/abs/2406.16989} {Retrieval-augmented mixture
  of lora experts for uploadable machine learning}.
\newblock \emph{Preprint}, arXiv:2406.16989.

\bibitem[{Zhao et~al.(2024{\natexlab{b}})Zhao, Gan, Wang, Zhou, Yang, Kuang,
  and Wu}]{zhao2024loraretriever}
Ziyu Zhao, Leilei Gan, Guoyin Wang, Wangchunshu Zhou, Hongxia Yang, Kun Kuang,
  and Fei Wu. 2024{\natexlab{b}}.
\newblock \href {https://doi.org/10.18653/v1/2024.findings-acl.263}
  {{L}ora{R}etriever: Input-aware {L}o{RA} retrieval and composition for mixed
  tasks in the wild}.
\newblock In \emph{Findings of the Association for Computational Linguistics:
  ACL 2024}, pages 4447--4462, Bangkok, Thailand. Association for Computational
  Linguistics.

\bibitem[{Zhao et~al.(2025)Zhao, shen, Zhu, Li, Su, Wang, and
  Wu}]{zhao2024mergingloras}
Ziyu Zhao, tao shen, Didi Zhu, Zexi Li, Jing Su, Xuwu Wang, and Fei Wu. 2025.
\newblock \href
  {https://proceedings.iclr.cc/paper_files/paper/2025/file/b54e0146a82945f01e69c2e3309ba925-Paper-Conference.pdf}
  {Merging loras like playing lego: Pushing the modularity of lora to extremes
  through rank-wise clustering}.
\newblock In \emph{International Conference on Learning Representations}.

\end{thebibliography}

\appendix

\section{Details of Training and Evaluation Datasets}
\label{sec:eval_dataset}

A subset of FLAN-v2 \citep{wei2022finetuned} is used for LoRA adapter training and for the mixed-task evaluation set. The adapter pool contains 48 task-specific LoRA adapters for each backbone. Each adapter is associated with 20 representative support examples for prototype construction. The mixed-task evaluation set contains 2,395 examples sampled from the same 48 tasks. The adapter pool, support examples, and evaluation set are taken from the public release by \citet{zhao2024loraretriever}. Table~\ref{tab:appendix-flan-adapters} lists the adapter tasks, task clusters, metrics, and number of mixed-task evaluation examples.

\begin{table*}[t]
\centering
\small
\setlength{\tabcolsep}{3pt}
\renewcommand{\arraystretch}{1.15}
\resizebox{\textwidth}{!}{%
\begin{tabular}{p{0.16\textwidth} p{0.50\textwidth} c c c p{0.20\textwidth}}
\toprule
\textbf{Cluster} & \textbf{Datasets / adapter tasks} & \textbf{Metric} & \textbf{Adapters} & \textbf{Eval examples} & \textbf{Notes} \\
\midrule
Closed-book QA
& natural\_questions; arc\_easy; arc\_challenge; trivia\_qa
& EM & 4 & 200
& General-knowledge and science QA without external retrieval. \\

Commonsense reasoning
& story\_cloze; piqa; copa; hellaswag
& EM & 4 & 200
& Commonsense, physical, and abductive reasoning tasks. \\

Coreference resolution
& definite\_pronoun\_resolution; wsc
& EM & 2 & 100
& Pronoun and entity-reference resolution tasks. \\

Natural language inference
& cb; wnli; mnli\_matched; mnli\_mismatched; anli\_r1; anli\_r2; anli\_r3; snli; qnli; rte
& EM & 10 & 495
& Entailment-style classification. The CB split contributes 45 examples; the other NLI tasks contribute 50 each. \\

Paraphrase detection
& paws\_wiki; glue\_qqp; glue\_mrpc; stsb
& EM & 4 & 200
& Sentence-pair semantic equivalence and similarity tasks. \\

Reading comprehension with commonsense
& cosmos\_qa; record
& EM & 2 & 100
& Passage understanding requiring commonsense or entity reasoning. \\

Reading comprehension
& multirc; squad\_v1; squad\_v2; openbookqa; bool\_q; drop
& EM & 6 & 300
& Passage-based QA, open-book QA, boolean QA, and discrete reasoning. \\

Sentiment analysis
& sst2; yelp\_polarity\_reviews; imdb\_reviews; sentiment140
& EM & 4 & 200
& Binary sentiment classification and sentiment-conditioned generation. \\

Struct-to-text
& web\_nlg\_en; dart; e2e\_nlg; common\_gen
& ROUGE & 4 & 200
& Natural-language generation from structured or concept input. \\

Translation
& wmt16\_translate\_tren; wmt16\_translate\_deen; wmt16\_translate\_ruen; wmt16\_translate\_fien; wmt16\_translate\_roen; wmt14\_enfr; wmt16\_translate\_csen; para\_crawl\_enes
& BLEU & 8 & 400
& Machine translation tasks from WMT and ParaCrawl. \\
\midrule
\textbf{Total}
& \textbf{48 FLAN-v2 adapter tasks}
& -- & \textbf{48} & \textbf{2,395}
& Each adapter has 20 support examples for prototype construction. \\
\bottomrule
\end{tabular}
}
\caption{FLAN-v2 adapter pool and mixed-task evaluation data. The same adapter-task set is used for both LLaMA-2-7B and LLaMA-2-13B.}
\label{tab:appendix-flan-adapters}
\end{table*}

\paragraph{IID and cross-task retrieval settings.}
The same mixed-task evaluation set is used in both retrieval settings. In the IID setting, the retrieval pool contains all 48 adapters, including the adapter trained on the task from which the current test example was drawn. In the cross-task setting, the source-task adapter is masked before retrieval. For example, if the current input comes from \texttt{arc\_challenge}, the \texttt{arc\_challenge} LoRA is removed from the candidate pool for that input, and routing is performed over the remaining 47 adapters. This setting evaluates whether the retrieval method can identify transferable adapters when the task-specific adapter is unavailable.

\paragraph{Prompt-level shot setting.}
All reported benchmark evaluations are zero-shot at the prompt level. The prompt contains only the current evaluation instance formatted with the Alpaca instruction template; no answered demonstrations are prepended for FLAN-v2 mixed-task evaluation, BBH, MMLU, GSM8K, MATH, AQuA, or CodeXGLUE. The 20 support examples associated with each adapter are used only offline to construct adapter prototypes for routing. They are not inserted into the model context during evaluation and therefore do not constitute one-shot or few-shot in-context examples.

\begin{table*}[t]
\centering
\small
\setlength{\tabcolsep}{3pt}
\renewcommand{\arraystretch}{1.15}
\resizebox{\textwidth}{!}{%
\begin{tabular}{p{0.12\textwidth} p{0.39\textwidth} c c c c p{0.20\textwidth}}
\toprule
\textbf{Benchmark} & \textbf{Subtasks / subjects} & \textbf{Metric} & \textbf{Adapters} & \textbf{Eval examples} & \textbf{Shot setting} & \textbf{Notes} \\
\midrule
BBH
& Boolean\_Expressions; Causal\_Judgement; Dyck\_Languages; Formal\_Fallacies; Navigate; Object\_Counting; Web of Lies
& EM & 0 & Full test split for each of the 7 subtasks
& Zero-shot
& Evaluated on the test split. These tasks are not used to train any adapter in the pool. \\

MMLU
& 57 subjects, including STEM, humanities, social science, and professional exams
& EM & 0 & Full test split for 57 subjects
& Zero-shot
& Evaluated subject-wise on the test split and averaged across subjects. \\

GSM8K
& main
& EM & 0 & Full test split
& Zero-shot
& Grade-school math word problems; evaluated on the test split. \\

MATH
& all
& EM & 0 & Full test split
& Zero-shot
& Competition-style math problems; evaluated on all test subsets. \\

AQuA
& test
& EM & 0 & Full test split
& Zero-shot
& Multiple-choice algebraic word problems from AQuA-RAT. \\

CodeXGLUE
& code-refinement-small; code-translation-java-cs; code-translation-cs-java; code-to-text-java; code-to-text-python
& BLEU & 0 & Full test split for each subtask
& Zero-shot
& Out-of-domain code tasks. CodeXGLUE uses a longer generation budget than classification-style tasks. \\
\bottomrule
\end{tabular}
}
\caption{Out-of-distribution evaluation benchmarks. No OOD benchmark has a matching task-specific adapter in the FLAN-v2 adapter pool, so all OOD experiments measure transfer from the 48 FLAN-trained adapters.}
\label{tab:appendix-ood-benchmarks}
\end{table*}

\paragraph{Prompt format.}
All mixed-task and OOD examples are wrapped with the Alpaca prompt template before generation. In the implementation, the example text is passed as the instruction field and the input field is left empty. The resulting prompt is:

\begin{quote}
\small\ttfamily
Below is an instruction that describes a task. Write a response that appropriately completes the request.\textbackslash n\textbackslash n\\
\#\#\# Instruction:\textbackslash n\\
\{instruction\}\textbackslash n\textbackslash n\\
\#\#\# Response:\textbackslash n
\end{quote}

When an example contains answer choices, the choices are already included inside the \texttt{\{instruction\}} text under an \texttt{OPTIONS:} block. The model output is parsed after the \texttt{\#\#\# Response:} delimiter and compared with the reference using the task-specific metric in Tables~\ref{tab:appendix-flan-adapters} and~\ref{tab:appendix-ood-benchmarks}.

\subsection{Implementation of Baselines}
\label{app:baseline-implementation}

All baselines are evaluated with the same backbone, adapter pool, prompt template, decoding settings, metric computation, and composition method as LoRAWeaver. Each method routes over the same 48 FLAN-v2 LoRA adapters and uses the same IID and cross-task candidate masks. The main implementation difference between baselines is the scoring function used to rank candidate adapters. For a method-specific score $S^m_{a,\ell}(q)$, adapters are selected as
\begin{equation}
\mathcal{T}^m_\ell(q)
=
\operatorname*{TopK}_{a\in\mathcal{C}(q)} S^m_{a,\ell}(q),
\end{equation}
where $\mathcal{C}(q)$ is the candidate pool. The selected adapters are then normalised with a softmax:
\begin{equation}
\lambda^m_{a,\ell}(q)
=
\frac{\exp(\tau S^m_{a,\ell}(q))}
{\sum\limits_{b\in\mathcal{T}^m_\ell(q)} \exp(\tau S^m_{b,\ell}(q))},
\quad a \in \mathcal{T}^m_\ell(q).
\end{equation}
The selected adapters are then composed with
\begin{equation}
\Delta h^{m,\mathrm{mix}}_\ell(x)
=
\sum_{a\in\mathcal{T}^m_\ell(q)}
\lambda^m_{a,\ell}(q)\,\Delta h_{a,\ell}(x).
\end{equation}
For methods that output one sequence-level score $S^m_a(q)$ rather than layer-specific scores, the same score is reused for all target modules, i.e., $S^m_{a,\ell}(q)=S^m_a(q)$. In the cross-task setting, the adapter trained on the current instance's source task is removed before ranking for every retrieval method.

\paragraph{LoRARetriever (LR).}
The LoRARetriever baseline follows the retrieve-then-compose setting of \citet{zhao2024loraretriever}. The trained retrieval encoder checkpoint \texttt{Styxxxx/lora\_retriever} is used. Let $g(\cdot)$ denote this encoder. Each adapter is represented by the mean embedding of its support examples,
\begin{equation}
\begin{aligned}
e_a
&=
\frac{1}{|\mathcal{D}_a|}
\sum_{x\in\mathcal{D}_a} g(x),\\
e_q&=g(q).
\end{aligned}
\end{equation}
Adapters are ranked by cosine similarity in the learned retrieval space:
\begin{equation}
S^{\mathrm{LR}}_a(q)
=
\cos(e_q,e_a).
\end{equation}
LoRARetriever therefore produces a single query-level adapter ranking. The resulting score is reused across target modules before applying the same top-$K$ and decoding settings as the other baselines.

\paragraph{Adapter Soup}
The Adapter Soup baseline follows the retrieval-and-fusion idea of \citet{chronopoulou2023adaptersoup}, modified to operate per query instead of once per domain. The general-purpose sentence encoder $\phi(\cdot)$ is INSTRUCTOR-XL \citep{su2023instructor}, using the public checkpoint \texttt{hkunlp/instructor-xl} (1.5B parameters, 768-dimensional embeddings) off the shelf, with no fine-tuning on the adapter pool or on any retrieval supervision. INSTRUCTOR encodes an instruction together with the text, and every input, whether a support example or a query, is embedded with the same fixed instruction prefix \emph{``Represent the sentence for similar task retrieval:''}. Each adapter is represented by the mean embedding of its support examples,
\begin{equation}
\begin{aligned}
u_a
&=
\frac{1}{|\mathcal{D}_a|}
\sum_{x\in\mathcal{D}_a} \phi(x),\\
u_q&=\phi(q).
\end{aligned}
\end{equation}
where $\mathcal{D}_a$ is the same set of 20 support examples used by every other method (Appendix~\ref{sec:eval_dataset}, Table~\ref{tab:appendix-flan-adapters}). Adapters are ranked by cosine similarity in the sentence-embedding space,
\begin{equation}
S^{\mathrm{AS}}_a(q)
=
\cos(u_q,u_a).
\end{equation}
In the implementation, $u_a$ is $L_2$-normalised once at index construction and retrieval is an inner-product search over the normalised adapter embeddings, which is equivalent to the cosine ranking above.
Adapter Soup produces one query-level score. The difference from LoRARetriever is the encoder: Adapter Soup uses the fixed general-purpose INSTRUCTOR-XL encoder, and LoRARetriever uses the \texttt{Styxxxx/lora\_retriever} encoder, which is INSTRUCTOR-XL further trained for adapter retrieval on the same adapter pool. The two baselines are therefore identical except for the encoder weights, which isolates the contribution of retrieval-specific training. The score is reused across target modules before applying the shared top-$K$ and composition steps.

\paragraph{LoRAHub.}
The LoRAHub baseline follows the gradient-free composition procedure of \citet{huang2024lorahub}. LoRAHub fits per-task coefficients on the target adapter's 20 labelled support examples. The cross-task setting masks that adapter and its support set, and the OOD benchmarks have no matching adapter at all, so in both settings LoRAHub's fitting data is unavailable by construction. It is reported only in the IID setting, where the evaluation task is represented in the pool. For a target task, scalar coefficients $\{w_a\}_{a\in\mathcal{C}}$ over the candidate adapters are obtained by minimising a cross-entropy objective on a small set of target-task examples $\mathcal{D}_{\mathrm{few}}$ ($|\mathcal{D}_{\mathrm{few}}| = 20$ in the experiments),
\begin{equation}
\begin{aligned}
\{w_a\}
={} & \operatorname*{arg\,min}_{\{w_a\}}
\sum_{(x,y)\in\mathcal{D}_{\mathrm{few}}}
\mathcal{L}_{\mathrm{CE}}\!\left(y,\, x \,;\, \Delta \hat{W}\right) \\
& + \gamma \sum_{a} |w_a|, \\[4pt]
\Delta \hat{W}
={} & \hat{B}\hat{A}, \\[2pt]
\hat{B} ={} & \textstyle\sum_{a} w_a B_a,
\quad
\hat{A} = \textstyle\sum_{a} w_a A_a,
\end{aligned}
\end{equation}
where each candidate adapter factorises as $\Delta W_a = B_a A_a$ with shared rank $r$, following the convention of Section~\ref{sec:problem-setup}, so that $A_a \in \mathbb{R}^{r \times d_{\mathrm{in}}}$, $B_a \in \mathbb{R}^{d_{\mathrm{out}} \times r}$, and the composed factors $\hat{A} \in \mathbb{R}^{r \times d_{\mathrm{in}}}$, $\hat{B} \in \mathbb{R}^{d_{\mathrm{out}} \times r}$ multiply to an update $\Delta \hat{W} \in \mathbb{R}^{d_{\mathrm{out}} \times d_{\mathrm{in}}}$ of rank at most $r$. Here $\mathcal{L}_{\mathrm{CE}}$ is the loss of the backbone conditioned on the composed update $\Delta \hat{W}$, and $\gamma$ controls the $L_1$ penalty. Because the coefficients are applied to the low-rank factors before multiplication, the composed update is quadratic in $\{w_a\}$ and includes cross terms $B_a A_b$ for $a\neq b$; it is therefore not a linear combination $\sum_a w_a \Delta W_a$ of the individual updates. A derivative-free optimiser produces the coefficients without backpropagation through the backbone. The coefficients are computed once per task and reused for every query in that task, and the composed update $\Delta \hat{W}$ is applied at all target modules. LoRAHub therefore composes adapters through learned scalar coefficients rather than through a per-query ranking score.

\paragraph{LoGo.}
LoGo is implemented as a training-free, instance-level router following \citet{lee2025logo}. A single probe forward pass is run with the candidate adapters active, and adapter relevance is estimated only from the final transformer block. Let $\ell^\star$ denote the final-layer \texttt{q\_proj} LoRA module and $t^\star$ the last valid input token. For adapter $a$, the probe activation is
\begin{equation}
\begin{aligned}
z_a(q)
&=
\Delta W_{a,\ell^\star}
h_{\ell^\star,t^\star}(q),\\
\Delta W_{a,\ell^\star}
&=
s_{a,\ell^\star}B_{a,\ell^\star}A_{a,\ell^\star}.
\end{aligned}
\end{equation}
LoGo defines two scores from the probe activation $z_a(q)$, and both are evaluated as separate baselines. The entropy score is
\begin{equation}
\begin{aligned}
p_a(q)&=\operatorname{softmax}(z_a(q)),\\
S^{\mathrm{LoGo\text{-}Ent}}_a(q)
&=
-H(p_a(q))\\
&=
\sum_i p_{a,i}(q)\log p_{a,i}(q).
\end{aligned}
\end{equation}
Lower entropy gives a larger routing score. The magnitude score is the norm of the same activation,
\begin{equation}
S^{\mathrm{LoGo\text{-}Norm}}_a(q)
=
\|z_a(q)\|_2.
\end{equation}
A larger norm gives a larger routing score. LoGo (Ent) uses $S^{\mathrm{LoGo\text{-}Ent}}_a(q)$, and LoGo (Norm) uses $S^{\mathrm{LoGo\text{-}Norm}}_a(q)$. Both variants produce one adapter score per query from $(\ell^\star,t^\star)$, and that score is reused for all composed target modules.

\paragraph{qa-FLoRA.} \cite{shukla2025qaflora}. For each query, the candidate adapter set is formed under the same candidate-mask and adapter-budget rules used by the other baselines. The base model's layer-wise vocabulary distribution is computed at the last input token $t^\star$; then one candidate adapter is activated at a time, and the corresponding adapter-conditioned vocabulary distribution is computed:
\begin{equation}
\begin{aligned}
p^0_\ell(q)
&=
\operatorname{softmax}(o^0_{\ell,t^\star}(q)),
\\
p^a_\ell(q)
&=
\operatorname{softmax}(o^a_{\ell,t^\star}(q)).
\end{aligned}
\end{equation}
Adapter weights are obtained from the KL divergence between the base distribution and the adapter-conditioned distribution,
\begin{equation}
\begin{aligned}
S^{\mathrm{qa}}_{a,\ell}(q)
&=
D_{\mathrm{KL}}\!\left(
p^0_\ell(q)
\,\|\,p^a_\ell(q)
\right).
\end{aligned}
\end{equation}
These scores are normalised over the candidate adapters at each layer. The resulting KL-derived layer scores are used for adapter composition during generation.

\paragraph{spectR.}
spectR follows the spectral routing procedure of \citet{fleshman2025spectr}. For every adapter and target module, the LoRA update is decomposed with an SVD,
\begin{equation}
\begin{aligned}
\Delta W_{a,\ell}
&=
s_{a,\ell}B_{a,\ell}A_{a,\ell}\\
&=
U_{a,\ell}\Sigma_{a,\ell}V_{a,\ell}^{\top},
\end{aligned}
\end{equation}
and the aligned spectral factors are cached. During inference, each token activation is projected into each adapter's spectral subspace:
\begin{equation}
\begin{aligned}
r_{a,\ell,t}(q)
&=
\Sigma_{a,\ell}V_{a,\ell}^{\top}
h_{\ell,t}(q),\\
S^{\mathrm{spectR}}_{a,\ell,t}(q)
&=
\|r_{a,\ell,t}(q)\|_2.
\end{aligned}
\end{equation}
Adapters are ranked by the norm of the projected activation, and the top-$K$ rank components are composed at token and layer level. Because $U_{a,\ell}$ has orthonormal columns, $\|U_{a,\ell}r_{a,\ell,t}(q)\|_2 = \|r_{a,\ell,t}(q)\|_2$, so at full rank this score is exactly the magnitude of the adapter output $\|\Delta W_{a,\ell}h_{\ell,t}(q)\|_2$; under rank truncation it is a lower bound on that quantity. The score therefore measures how strongly the token activation lies in the adapter's learned update subspace. spectR performs token-level routing as well as layer-level routing, using weight-space structure and activation-conditioned signals without training an additional router.

\paragraph{Oracle.}
The oracle baseline uses the task-specific adapter associated with the evaluation instance. In the IID mixed-task setting, this is the adapter trained for the source task. For OOD benchmarks where no matching adapter exists in the FLAN-v2 pool, the oracle is undefined and is therefore not reported. The oracle is therefore an upper reference for task-identifiable in-pool examples, not a learned routing method.

\section{Algorithm and Implementation Details}
\label{app:algorithm-implementation-details}

\begin{algorithm}[t]
\caption{LoRAWeaver: Layer-wise Prototype Routing and Composition}
\label{alg:lorweaver}
\small
\begin{algorithmic}[1]
\Require Frozen LM $f_\theta$; adapter library $\mathcal{E}$; support sets $\{\mathcal{D}_a\}_{a\in\mathcal{E}}$; LoRA factors $\{A_{a,\ell},B_{a,\ell},s_{a,\ell}\}$; top-$K$ value $K$; inverse temperature $\tau$
\Ensure Layer-wise adapter mixtures for query $q$

\Statex \textbf{Offline prototype construction}
\ForAll{$a \in \mathcal{E}$}
    \ForAll{$x \in \mathcal{D}_a$}
        \State Run $x$ through $f_\theta$ once and capture hidden inputs $H_\ell(x)$ at all target modules
        \For{$\ell = 1$ to $L$}
            \State $\bar h_\ell(x) \gets \mathrm{MaskedMeanPool}(H_\ell(x))$
        \EndFor
    \EndFor
    \For{$\ell = 1$ to $L$}
        \State $\mu_{a,\ell} \gets |\mathcal{D}_a|^{-1}\sum_{x\in\mathcal{D}_a}\bar h_\ell(x)$
        \State $p_{a,\ell} \gets B_{a,\ell}A_{a,\ell}\mu_{a,\ell}$
        \State $\hat p_{a,\ell} \gets p_{a,\ell}/\|p_{a,\ell}\|_2$
    \EndFor
\EndFor

\Statex \textbf{Online routing and composition}
\State Set candidate pool $\mathcal{C}(q)\gets\mathcal{E}$, optionally removing the source-task adapter in the cross-task setting
\State Run $q$ through $f_\theta$ and capture hidden inputs $H_\ell(q)$ at all target modules
\For{$\ell = 1$ to $L$}
    \State $\bar h_\ell(q) \gets \mathrm{MaskedMeanPool}(H_\ell(q))$
    \ForAll{$a \in \mathcal{C}(q)$}
        \State $r_{a,\ell}(q) \gets B_{a,\ell}A_{a,\ell}\bar h_\ell(q)$
        \State $\hat r_{a,\ell}(q) \gets r_{a,\ell}(q)/\|r_{a,\ell}(q)\|_2$
        \State $S_{a,\ell}(q) \gets \hat r_{a,\ell}(q)^\top \hat p_{a,\ell}$
    \EndFor
    \State $\mathcal{T}_\ell(q) \gets \operatorname*{TopK}_{a\in\mathcal{C}(q)} S_{a,\ell}(q)$
    \State $Z_\ell(q) \gets \sum_{b\in\mathcal{T}_\ell(q)}\exp\!\left(\tau S_{b,\ell}(q)\right)$
    \ForAll{$a \in \mathcal{T}_\ell(q)$}
        \State $\lambda_{a,\ell}(q) \gets \exp\!\left(\tau S_{a,\ell}(q)\right)/Z_\ell(q)$
    \EndFor
\EndFor
\State Generate with the layer-wise mixed LoRA updates $\Delta h^{\mathrm{mix}}_\ell(x)=\sum_{a\in\mathcal{T}_\ell(q)}\lambda_{a,\ell}(q)\Delta h_{a,\ell}(x)$, applied to every token representation $x$ at layer $\ell$
\end{algorithmic}
\end{algorithm}

The masked mean pooling operation used in Algorithm \ref{alg:lorweaver} is
\begin{equation}
\label{eq:masked-mean-pool}
\begin{aligned}
\bar h_\ell(x) &= M_x^{-1}\sum_{t=1}^{T_x} m_t H_{\ell,t}(x),\\
M_x &= \max\!\left\{1,\sum_{t=1}^{T_x} m_t\right\}.
\end{aligned}
\end{equation}
where $m_t$ is the attention mask. During prototype construction and online routing, hidden states are captured from the inputs to the LoRA target modules. A single pass yields the hidden inputs at all target modules simultaneously, so prototype construction costs one forward pass per support example, $\sum_{a\in\mathcal{E}}|\mathcal{D}_a|$ in total, and routing costs one probe pass per query. KV caching is disabled during these probe passes because routing only requires one forward pass over the input sequence. Implementation details of LoRAWeaver are reported in Table \ref{tab:implementation-details}.

\begin{table*}[t]
\centering
\small
\setlength{\tabcolsep}{4pt}
\renewcommand{\arraystretch}{1.15}
\resizebox{\textwidth}{!}{%
\begin{tabular}{p{0.24\textwidth} p{0.68\textwidth}}
\toprule
\textbf{Item} & \textbf{Setting} \\
\midrule
Backbone models
& \texttt{meta-llama/Llama-2-7b-hf} and \texttt{meta-llama/Llama-2-13b-hf}. \\

Adapter pool
& 48 publicly released FLAN-v2 LoRA adapters for each backbone. Adapter names follow \texttt{Styxxxx/llama2\_7b\_lora-\{task\}} and \texttt{Styxxxx/llama2\_13b\_lora-\{task\}}. \\

LoRA configuration
& Rank $r=6$ and scaling hyperparameter $\alpha=12$, giving $s=\alpha/r=2$. \\

LoRA target modules
& The LoRA adapters are attached to the self-attention query and value projections, \texttt{q\_proj} and \texttt{v\_proj}, in every transformer block. The configuration gives 64 target modules for LLaMA-2-7B and 80 target modules for LLaMA-2-13B. \\

Prototype support set
& 20 representative input examples per adapter are used to construct the layer-wise prototypes. Inputs are tokenised with maximum length 512 and pooled with the attention-mask-weighted mean in Eq.~\ref{eq:masked-mean-pool}. \\

Retrieval setting
& Unless otherwise stated, layer-wise routing uses cosine-prototype scoring and selects top-$K=3$ adapters at each target module. In the cross-task setting, the source-task adapter is removed from the candidate pool before ranking. \\

Composition weights
& Selected adapters are weighted with a softmax over their retrieval scores, $\lambda_{a,\ell}\propto\exp(\tau S_{a,\ell})$. The softmax inverse temperature is fixed to $\tau=10$ in all experiments. \\

Runtime composition
& Top-$K$ composition is implemented as a direct output-space mixture in PEFT. For each target module $m$, the routing step produces a sparse vector over the active adapters, with nonzero entries only for the selected top-$K$ adapters. During generation, PEFT keeps the LoRA factors separate and computes
$W_m u_{m,t}+\sum_{a\in\mathcal{T}_m(q)}\lambda_{a,m}s_{a,m}B_{a,m}A_{a,m}u_{m,t}$
inside the module forward pass. Thus, the method does not form a dense merged update $\sum_a\lambda_{a,m}\Delta W_{a,m}$ or reattach merged weights at each layer. \\

Decoding
& Greedy decoding is used with \texttt{do\_sample=False} and \texttt{num\_beams=1}; no sampling temperature is applied. The default generation budget is 50 new tokens; CodeXGLUE, GSM8K and MATH evaluations use 256 new tokens. Unless specified otherwise, reported benchmark runs use batch size 1. \\

Precision and implementation
& Models are loaded through Hugging Face Transformers and PEFT. Base models and adapters are held in \texttt{bfloat16} at evaluation time. LoRA factors used for routing are extracted from PEFT adapter weights. \\

Hardware
& Efficiency experiments are run on a single NVIDIA L40S GPU. The main evaluation scripts launch one GPU process per evaluation job. \\

Probe-pass cache
& KV caching is disabled during prototype and query probe passes. During probing, only the hidden inputs needed for LoRA target modules are captured and mean-pooled. \\
\bottomrule
\end{tabular}
}
\caption{Implementation details for LoRAWeaver}
\label{tab:implementation-details}
\end{table*}

\begin{figure}[t]
    \centering
    \includegraphics[width=\columnwidth]{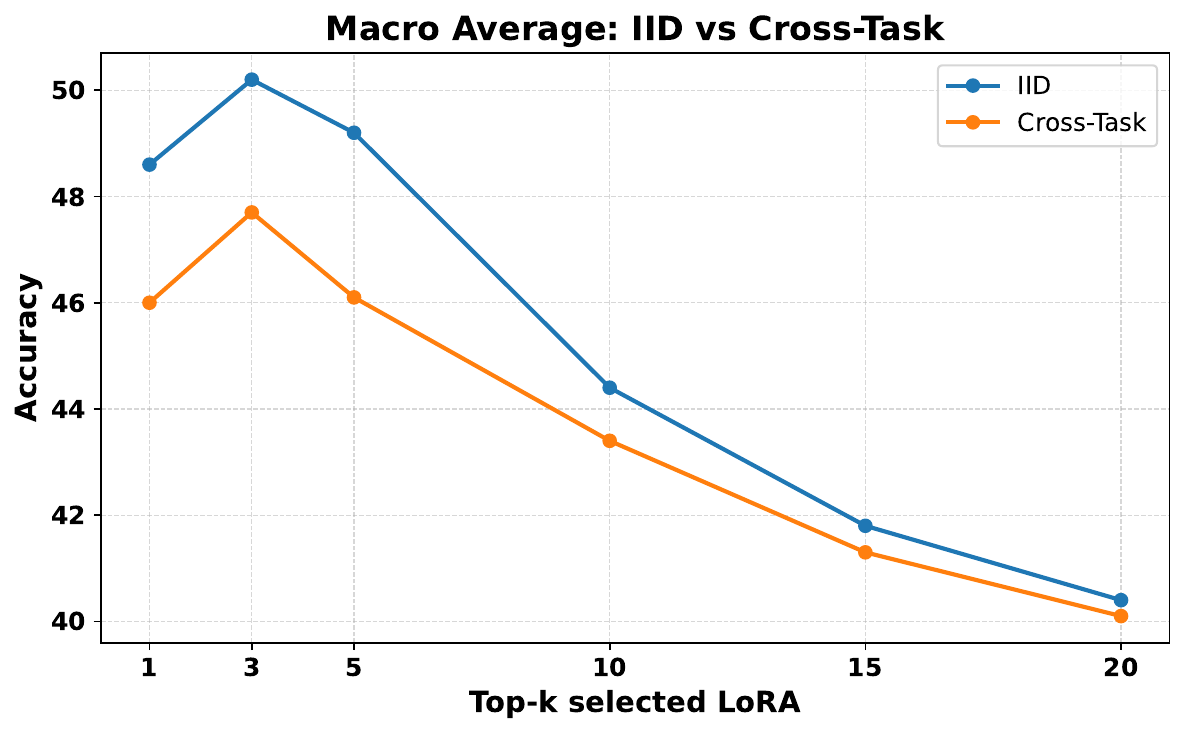}
    \caption{Ablation of the number of selected LoRA modules on LLaMA-2-7B, evaluated on the mixed-task FLAN-v2 benchmark. Macro-average performance is shown for the IID and cross-task settings as $K$ varies. The peak at $K=3$ indicates the selected $K$ for LoRAWeaver.}
    \label{fig:numberofloras}
\end{figure}

\subsection{Effect of Top-$K$ Retrieval}
\label{sec:selected-modules}

An ablation evaluates the number of selected modules $K$ in LoRAWeaver across $K \in \{1, 3, 5, 10, 15, 20\}$ on the mixed-task FLAN-v2 benchmark with LLaMA-2-7B. Figure~\ref{fig:numberofloras} shows that performance increases up to $K=3$ and declines for higher values of $K$. The result indicates that selecting three modules provides the best trade-off between compositional gain and module interference.

\section{LLaMA-3.1-8B Backbone Generalisation: Experimental Details}
\label{app:llama3}

This section describes the experimental setup for the backbone generalisation study in Section~\ref{sec:llama3}. The experiment adopts a composite-task evaluation setting inspired by qa-FLoRA \citep{shukla2025qaflora}, with one key modification. The qa-FLoRA setup retains only the two task-specific adapters required for each evaluation instance, and both adapters are fused during inference. The present study uses a 4 adapter pool that contains multiple domain experts. The routing method selects 2 adapters used for composition from this pool. The objective is to examine whether the adapter-output directional-similarity signal transfers beyond the LLaMA-2 backbone family under a constrained compute budget that does not permit training a complete FLAN-v2 adapter pool or a retrieval encoder for an additional backbone.

\paragraph{Backbones.}
The experiments use LLaMA-2-7B and LLaMA-3.1-8B as frozen base models. Model adaptation is performed with LoRA modules. Inference with LLaMA-3.1-8B uses bfloat16 precision.

\paragraph{LoRA experts.}
Each backbone contains four LoRA experts. Three experts target Chinese (zh), Russian (ru), and Spanish (es). One expert targets mathematical reasoning. The language experts are trained on the corresponding 52K conversational datasets from \citet{lai2023}. The Math expert is trained on MetaMathQA \citep{yu2023metamath}. All adapters use LoRA rank $r=64$ and scaling factor $\alpha=16$. Training runs for three epochs with a cosine learning-rate schedule, a peak learning rate of $1\times10^{-4}$, and a warmup ratio of $0.04$.

\paragraph{Evaluation.}
Routing performance is evaluated on the multilingual grade-school mathematics benchmark MGSM \citep{shi2023language}. The evaluation uses the Chinese, Russian, and Spanish subsets. Each subset contains 250 test samples. All methods use 8-shot chain-of-thought prompting. Performance is measured with exact-match accuracy on the final numerical answer.

\paragraph{Baselines and routing configuration.}
A retrieval encoder is not trained for the selected backbone and adapter pool. As a result, LoRARetriever is excluded from the comparison. The evaluated methods are LoRAWeaver, Adapter Soup, LoGo (Ent), LoGo (Norm), qa-FLoRA, and spectR. All methods operate without additional training. Implementation details appear in Appendix~\ref{app:baseline-implementation}. For LoRAWeaver, each adapter prototype is constructed from 20 support samples. Adapter composition uses the top $K=2$ selected adapters.

\section{Per-Task Performance}
\label{sec:per-task-performace}

Tables~\ref{tab:iidtask} and~\ref{tab:crosstasks} report the full per-task results for the mixed-task benchmark. Each domain-metric cell in Tables~\ref{tab:iid} and~\ref{tab:crosstask} is the column mean of the corresponding task rows in these tables. For struct-to-text this means ROUGE is computed separately on \texttt{web\_nlg\_en}, \texttt{dart}, \texttt{e2e\_nlg}, and \texttt{common\_gen} and the four scores averaged.

\begin{table*}[t] 
\begingroup
\scriptsize
\setlength{\tabcolsep}{2pt}
\renewcommand{\arraystretch}{0.82}
\resizebox*{\textwidth}{\textheight}{%
\newcolumntype{Y}{>{\centering\arraybackslash}m{0.7cm}}
\begin{tabular}{l|>{\columncolor{gray!20}\centering\arraybackslash}m{0.7cm}>{\columncolor{blue!12}\centering\arraybackslash}m{0.7cm}YYYYYYY|>{\columncolor{gray!20}\centering\arraybackslash}m{0.7cm}>{\columncolor{blue!12}\centering\arraybackslash}m{0.7cm}YYYYYYY}
\toprule
\toprule
 & \multicolumn{9}{|c|}{\textbf{Llama-2-7b}} & \multicolumn{9}{c}{\textbf{Llama-2-13b}} \\
Task & Oracle & LW & LR & Adapter Soup & LoRA Hub & LoGo (Ent) & LoGo (Norm) & qa-FLoRA & spectR & Oracle & LW & LR & Adapter Soup & LoRA Hub & LoGo (Ent) & LoGo (Norm) & qa-FLoRA & spectR \\
\midrule
\midrule
\multicolumn{19}{l}{\textbf{Struct to Text}} \\
web\_nlg\_en Rouge-1 & 69.5 & 33.5 & 50.8 & \underline{51.2} & \textbf{57.1} & 27.2 & 37.9 & 50.7 & 48.4 & 72.3 & 45.1 & \textbf{56.8} & 40.6 & \underline{49.2} & 40.4 & 40.3 & 29.1 & 39.8 \\
web\_nlg\_en Rouge-2 & 48.8 & 15.7 & 27.3 & \underline{28.0} & \textbf{35.2} & 11.7 & 20.1 & 27.5 & 24.5 & 52.6 & 24.7 & \textbf{33.3} & 19.4 & \underline{29.1} & 18.8 & 19.6 & 13.4 & 18.2 \\
web\_nlg\_en Rouge-l & 62.6 & 31.0 & 46.5 & 46.5 & \textbf{51.5} & 25.4 & 33.9 & \underline{46.6} & 43.6 & 65.9 & 40.5 & \textbf{52.1} & 36.9 & \underline{45.0} & 36.7 & 36.6 & 25.9 & 38.1 \\
dart Rouge-1 & 71.4 & 56.7 & \textbf{67.1} & 58.4 & \underline{58.6} & 47.8 & 48.4 & 58.4 & 55.1 & 68.0 & 57.1 & \textbf{72.9} & 55.5 & \underline{62.6} & 56.1 & 57.0 & 29.7 & 53.2 \\
dart Rouge-2 & 48.8 & 33.1 & \textbf{45.0} & 33.0 & \underline{35.6} & 26.3 & 25.1 & 32.4 & 29.7 & 44.9 & 32.5 & \textbf{53.5} & 32.4 & \underline{39.4} & 31.1 & 31.5 & 12.7 & 29.1 \\
dart Rouge-l & 64.4 & 49.8 & \textbf{59.4} & \underline{52.6} & 52.0 & 43.8 & 43.9 & 51.9 & 48.5 & 61.7 & 50.1 & \textbf{67.3} & 49.6 & \underline{56.7} & 50.7 & 51.6 & 25.8 & 48.3 \\
e2e\_nlg Rouge-1 & 65.5 & 58.9 & 48.4 & \textbf{64.7} & 59.2 & 45.5 & 50.1 & \underline{60.2} & 57.7 & 66.4 & 58.9 & \underline{66.0} & 58.8 & \textbf{66.5} & 56.4 & 56.4 & 61.9 & 58.0\\
e2e\_nlg Rouge-2 & 39.3 & 32.5 & 24.4 & \textbf{36.9} & 32.5 & 23.8 & 26.0 & \underline{32.8} & 29.9 & 38.7 & 31.2 & \underline{38.8} & 32.0 & \textbf{39.3} & 28.8 & 28.9 & 35.1 & 31.1\\
e2e\_nlg Rouge-l & 55.8 & 50.9 & 40.9 & \textbf{54.7} & 49.9 & 39.6 & 43.0 & \underline{51.4} & 49.4 & 56.5 & 47.1 & \underline{55.6} & 49.8 & \textbf{55.9} & 46.2 & 46.0 & 50.9 & 50.9\\
common\_gen Rouge-1 & 47.2 & 30.1 & \textbf{46.7} & \underline{32.9} & 29.2 & 24.5 & 27.1 & 29.7 & 28.1 & 48.4 & 29.5 & 30.7 & 28.3 & \underline{40.1} & 33.6 & 32.8 & \textbf{48.5} & 28.2 \\
common\_gen Rouge-2 & 18.6 & 8.6 & \textbf{19.4} & \underline{10.8} & 7.6 & 6.4 & 8.7 & 8.5 & 8.2 & 19.4 & 10.6 & 8.6 & 8.9 & \underline{14.6} & 12.0 & 11.8 & \textbf{19.4} & 8.1 \\
common\_gen Rouge-l & 43.3 & 27.3 & \textbf{42.9} & \underline{28.8} & 25.2 & 23.0 & 24.4 & 24.5 & 24.4 & 44.2 & 26.7 & 26.4 & 24.7 & \underline{36.6} & 29.7 & 29.1 & \textbf{44.1} & 24.4 \\
\midrule
\multicolumn{19}{l}{\textbf{Translation}} \\
wmt16\_translate\_tren BLEU & 3.4 & \textbf{3.8} & 1.7 & \textbf{3.8} & 2.7 & 3.1 & 1.5 & \underline{3.4} & 3.1 & 2.4 & \textbf{4.3} & 2.7 & 2.5 & \underline{3.6} & 2.4 & 2.4 & 0.8 & 3.0 \\
wmt16\_translate\_deen BLEU & 18.4 & \textbf{19.7} & 19.3 & \underline{19.6} & 17.1 & 18.1 & 16.6 & 17.1 & 19.2 & 20.6 & 18.6 & 20.5 & 19.1 & 19.0 & 21.5 & \underline{21.8} & 21.3 & \textbf{22.2} \\
wmt16\_translate\_ruen BLEU & 10.5 & \textbf{10.2} & \underline{9.9} & 9.1 & 8.9 & 8.8 & 7.2 & 9.7 & 8.1 & 11.6 & 11.0 & \textbf{11.8} & 10.0 & \underline{11.2} & 9.9 & 10.3 & 10.1 & 10.5 \\
wmt16\_translate\_fien BLEU & 6.9 & 7.1 & 7.3 & 7.1 & \textbf{8.0} & \underline{7.5} & 5.4 & 6.5 & 6.6 & 6.7 & 8.0 & 6.8 & 6.4 & 6.9 & \textbf{10.3} & 9.4 & 6.7 & \underline{9.7} \\
wmt16\_translate\_roen BLEU & 14.0 & 13.0 & 9.6 & \textbf{13.5} & 9.9 & 10.7 & 9.8 & \underline{13.1} & 10.8 & 14.0 & 11.8 & \underline{14.2} & 9.6 & 11.3 & 11.1 & 10.7 & 12.0 & \textbf{16.5} \\
wmt14\_enfr BLEU & 16.3 & \underline{17.6} & 16.7 & 17.3 & 13.9 & \textbf{18.1} & 16.4 & 17.5 & 15.2 & 16.4 & 18.6 & 19.6 & 17.2 & 19.6 & \textbf{20.5} & \underline{20.4} & 19.2 & 19.8\\
wmt16\_translate\_csen BLEU & 9.9 & 7.6 & \textbf{8.0} & \underline{7.7} & 6.9 & \underline{7.7} & 6.0 & 6.9 & 7.1 & 6.9 & \textbf{12.1} & \underline{11.4} & 7.4 & 9.3 & 10.4 & 9.7 & 6.3 & 8.6 \\
para\_crawl\_enes BLEU & 24.8 & 21.4 & 19.1 & \textbf{24.1} & 19.3 & 21.8 & \underline{22.4} & \textbf{24.1} & 22.2 & 24.6 & 24.5 & 24.9 & 17.7 & 24.0 & \textbf{27.0} & \underline{26.3} & 24.2 & 26.1 \\
\midrule
\multicolumn{19}{l}{\textbf{COMMONSENSE}} \\
story\_cloze EM & 72.0 & 80.0 & 80.0 & 70.0 & 76.0 & \underline{82.0} & 80.0 & 74.0 & \textbf{86.0} & 96.0 & \textbf{100.0} & 96.0 & 80.0 & \underline{98.0} & 94.0 & 94.0 & 88.0 & 90.0\\
piqa EM & 44.0 & \textbf{42.0} & 36.0 & 36.0 & \textbf{42.0} & 38.0 & \underline{40.0} & 38.0 & \textbf{42.0} & 48.0 & \textbf{46.0} & 36.0 & 40.0 & 42.0 & 42.0 & \underline{44.0} & 36.0 & 38.0 \\
copa EM & 86.0 & 78.0 & \textbf{86.0} & 78.0 & \underline{80.0} & 78.0 & 76.0 & \underline{80.0} & 78.0 & 78.0 & \textbf{80.0} & \underline{78.0} & \textbf{80.0} & 70.0 & \underline{78.0} & \underline{78.0} & 64.0 & 68.0 \\
hellaswag EM & 50.0 & 38.0 & \textbf{48.0} & 18.0 & 26.0 & \underline{42.0} & 40.0 & 22.0 & 38.0 & 42.0 & \textbf{54.0} & \textbf{54.0} & 32.0 & \underline{52.0} & 42.0 & 38.0 & 50.0 &38.0 \\
\midrule
\multicolumn{19}{l}{\textbf{sentiment}} \\
sst2 EM & 98.0 & \textbf{96.0} & \textbf{96.0} & 82.0 & \underline{94.0} & 84.0 & 90.0 & 68.0 & \underline{94.0} & 98.0 & \textbf{100.0} & \underline{98.0} & 96.0 & 94.0 & 90.0 & 90.0 & \underline{98.0} & 70.0 \\
yelp\_polarity\_reviews EM & 98.0 & \underline{96.0} & \textbf{98.0} & \textbf{98.0} & \underline{96.0} & 92.0 & 92.0 & 94.0 & 94.0 & 98.0 & \underline{98.0} & \textbf{100.0} & 96.0 & \underline{98.0} & 94.0 & 94.0 & 96.0 & 94.0 \\
imdb\_reviews EM & 96.0 & \textbf{98.0} & 94.0 & 90.0 & \underline{96.0} & 74.0 & 76.0 & 72.0 & 80.0 & 98.0 & \underline{98.0} & \underline{98.0} & \textbf{100.0} & \underline{98.0} & 80.0 & 80.0 & 82.0 & 78.0 \\
sentiment140 EM & 68.0 & 66.0 & \textbf{72.0} & 54.0 & \textbf{72.0} & 66.0 & 60.0 & 56.0 & \underline{68.0} & 70.0 & 68.0 & 70.0 & 72.0 & \underline{74.0} & \underline{74.0} & 72.0 & 70.0 & \textbf{78.0} \\
\midrule
\multicolumn{19}{l}{\textbf{READING Comp.}} \\
multirc EM & 68.0 & \underline{46.0} & \textbf{60.0} & \underline{46.0} & \textbf{60.0} & 34.0 & 34.0 & 38.0 & 38.0 & 48.0 & \underline{58.0} & \textbf{84.0} & 50.0 & \underline{58.0} & 26.0 & 24.0 & 30.0 & 24.0 \\
squad\_v2 EM & 62.0 & 28.0 & \textbf{62.0} & 30.0 & \underline{58.0} & 30.0 & 26.0 & 24.0 & 20.0 & 44.0 & \underline{32.0} & \textbf{60.0} & 28.0 & 24.0 & 20.0 & 20.0 & 12.0 & 20.0 \\
squad\_v1 EM & 70.0 & 68.0 & 54.0 & 64.0 & 68.0 & \underline{76.0} & \textbf{78.0} & 68.0 & 68.0 & 68.0 & \underline{66.0} & 62.0 & \textbf{68.0} & 60.0 & 62.0 & 62.0 & 60.0 & 58.0 \\
openbookqa EM & 82.0 & \textbf{74.0} & 56.0 & \underline{72.0} & 22.0 & 66.0 & 68.0 & 68.0 & 70.0 & 84.0 & \textbf{78.0} & \underline{76.0} & 66.0 & \underline{76.0} & 74.0 & 74.0 & \textbf{78.0} & 66.0 \\
bool\_q EM & 84.0 & 80.0 & 74.0 & 80.0 & \underline{82.0} & 78.0 & 76.0 & \textbf{84.0} & 80.0 & 76.0 & \textbf{88.0} & \underline{84.0} & 72.0 & 74.0 & \textbf{88.0} & \textbf{88.0} &82.0 & \underline{84.0} \\
drop EM & 40.0 & \underline{22.0} & 6.0 & 12.0 & \textbf{24.0} & 16.0 & 16.0 & 12.0 & 12.0 & 28.0 & 26.0 & \textbf{32.0} & 14.0 & \underline{28.0} & 24.0 & 24.0 & 20.0 & 16.0 \\
\midrule
\multicolumn{19}{l}{\textbf{CLOSE-BOOK QA}} \\
natural\_questions EM & 18.0 & \textbf{18.0} & 6.0 & \underline{16.0} & 10.0 & \textbf{18.0} & \underline{16.0} & 14.0 & \textbf{18.0} & 20.0 & \underline{20.0} & 14.0 & 18.0 & \textbf{22.0} & 18.0 & \underline{20.0} & 14.0 & 12.0 \\
arc\_easy EM & 50.0 & 70.0 & 52.0 & 52.0 & 60.0 & 68.0 & \textbf{80.0} & 64.0 & \underline{78.0} & 92.0 & \underline{92.0} & 84.0 & 66.0 & 76.0 & \textbf{94.0} & \textbf{94.0} & 86.0 &90.0 \\
arc\_challenge EM & 46.0 & 48.0 & 44.0 & 32.0 & 46.0 & 44.0 & \underline{52.0} & 40.0 & \textbf{54.0} & 66.0 & \textbf{70.0} & \underline{66.0} & 52.0 & 58.0 & \textbf{70.0} & \textbf{70.0} & \underline{66.0} & 64.0 \\
trivia\_qa EM & 64.0 & 56.0 & 24.0 & 50.0 & 32.0 & \textbf{60.0} & \underline{58.0} & 54.0 & 56.0 & 48.0 & \underline{68.0} & 36.0 & 54.0 & 40.0 & \textbf{70.0} & \textbf{70.0} & 34.0 & 56.0 \\
\midrule
\multicolumn{19}{l}{\textbf{COREFERENCE}} \\
definite\_pronoun\_resolution EM & 56.0 & \textbf{72.0} & \underline{70.0} & 56.0 & 60.0 & 68.0 & \textbf{72.0} & 60.0 & 68.0 & 88.0 & \underline{82.0} & \textbf{90.0} & 72.0 & 30.0 & 70.0 & 70.0 & 44.0 & 62.0 \\
wsc EM & 50.0 & \underline{54.0} & 50.0 & 48.0 & \textbf{58.0} & 52.0 & 52.0 & 44.0 & \underline{54.0} & 64.0 & \textbf{56.0} & 46.0 & \underline{50.0} & \textbf{56.0} & 46.0 & 46.0 & \underline{50.0} & 38.0 \\
\midrule
\multicolumn{19}{l}{\textbf{READ. COMP. W/ COMMONSENSE}} \\
cosmos\_qa EM & 72.0 & 56.0 & \underline{68.0} & 40.0 & \textbf{80.0} & 46.0 & 56.0 & 40.0 & 54.0 & 86.0 & \textbf{84.0} & 26.0 & 48.0 & 66.0 & 80.0 & \textbf{84.0} & \underline{82.0} & 56.0 \\
record EM & 70.0 & \underline{38.0} & \textbf{52.0} & 18.0 & \textbf{52.0} & 34.0 & 32.0 & 32.0 & 30.0 & 76.0 & \textbf{54.0} & 38.0 & 40.0 & 30.0 & 38.0 & 40.0 & 44.0 & \underline{48.0} \\
\midrule
\multicolumn{19}{l}{\textbf{PARAPHRASE}} \\
paws\_wiki EM & 90.0 & 62.0 & \textbf{90.0} & 42.0 & 54.0 & 46.0 & 58.0 & 48.0 & \underline{64.0} & 88.0 & \textbf{74.0} & 50.0 & \underline{60.0} & 58.0 & 52.0 & 52.0 & 52.0 & 46.0 \\
glue\_qqp EM & 74.0 & \textbf{74.0} & \underline{66.0} & 64.0 & 58.0 & 50.0 & 54.0 & 60.0 & 58.0 & 88.0 & \underline{80.0} & \textbf{90.0} & 60.0 & 26.0 & 46.0 & 44.0 & 78.0 & 52.0 \\
glue\_mrpc EM & 60.0 & \textbf{64.0} & 58.0 & \underline{60.0} & 38.0 & 50.0 & 44.0 & \textbf{64.0} & 56.0 & 82.0 & 64.0 & \underline{66.0} & 62.0 & 56.0 & 62.0 & 62.0 & \textbf{84.0} & 62.0 \\
stsb EM & 36.0 & \underline{36.0} & \textbf{40.0} & 14.0 & 10.0 & 16.0 & 16.0 & 10.0 & 24.0 & 40.0 & \underline{28.0} & 16.0 & \textbf{36.0} & \underline{28.0} & 10.0 & 6.0 & \underline{28.0} & 20.0 \\
\midrule
\multicolumn{19}{l}{\textbf{NLI}} \\
cb EM & 86.7 & \underline{82.2} & 71.1 & \underline{82.2} & \textbf{88.9} & 75.6 & 75.6 & \underline{82.2} & 73.3 & 93.3 & \textbf{88.9} & 80.0 & 75.6 & \underline{86.7} & 80.0 & 80.0 & 82.2 & 62.2 \\
wnli EM & 72.0 & \underline{64.0} & 54.0 & 52.0 & \textbf{70.0} & 48.0 & 52.0 & 54.0 & 54.0 & 72.0 & \textbf{66.0} & 60.0 & \underline{62.0} & 58.0 & 50.0 & 46.0 & \underline{62.0} & 40.0 \\
mnli\_matched EM & 88.0 & \textbf{84.0} & \textbf{84.0} & 58.0 & 46.0 & 58.0 & 62.0 & 56.0 & \underline{70.0} & 86.0 & \textbf{94.0} & 88.0 & 82.0 & 88.0 & 88.0 & \underline{90.0} & 86.0 & 68.0 \\
anli\_r3 EM & 46.0 & 40.0 & \textbf{46.0} & 40.0 & 42.0 & 40.0 & 38.0 & \underline{44.0} & 42.0 & 46.0 & 48.0 & \textbf{58.0} & 42.0 & 26.0 & \underline{52.0} & \underline{52.0} & \textbf{58.0} & 44.0 \\
anli\_r2 EM & 46.0 & \underline{40.0} & \textbf{44.0} & 28.0 & 34.0 & 38.0 & 36.0 & 26.0 & 38.0 & 56.0 & \underline{50.0} & \textbf{56.0} & 46.0 & 40.0 & 40.0 & 40.0 & \textbf{56.0} & 40.0 \\
anli\_r1 EM & 50.0 & 34.0 & \textbf{50.0} & 34.0 & 36.0 & 32.0 & \underline{38.0} & 36.0 & 36.0 & 70.0 & \underline{56.0} & \textbf{70.0} & 40.0 & 48.0 & 48.0 & 48.0 & \textbf{70.0} & 34.0 \\
mnli\_mismatched EM & 92.0 & \underline{90.0} & \textbf{94.0} & 68.0 & 58.0 & 80.0 & 78.0 & 64.0 & 88.0 & 92.0 & \textbf{100.0} & \underline{92.0} & 82.0 & 80.0 & \underline{92.0} & 90.0 & \underline{92.0} & 78.0 \\
snli EM & 96.0 & \textbf{90.0} & \underline{84.0} & 58.0 & 68.0 & 72.0 & 64.0 & 56.0 & 60.0 & 92.0 & \textbf{96.0} & 88.0 & 62.0 & \underline{94.0} & 74.0 & 78.0 & 88.0 & 64.0 \\
qnli EM & 94.0 & 58.0 & 62.0 & \underline{64.0} & \textbf{74.0} & 46.0 & 56.0 & \underline{64.0} & 48.0 & 78.0 & \underline{72.0} & 54.0 & 54.0 & \textbf{74.0} & 68.0 & 68.0 & 56.0 & 48.0 \\
rte EM & 54.0 & 60.0 & \underline{64.0} & 62.0 & 62.0 & 52.0 & 58.0 & \textbf{68.0} & 60.0 & 82.0 & \underline{78.0} & 68.0 & 58.0 & \textbf{82.0} & 68.0 & 68.0 & 66.0 & 50.0 \\
\bottomrule
\end{tabular}
}
\endgroup

\caption{Per-task IID performance on the 48-task FLAN-v2 mixed-task benchmark.}
\label{tab:iidtask}
\end{table*}

\begin{table*}[t] 

\begingroup
\scriptsize
\setlength{\tabcolsep}{2pt}
\renewcommand{\arraystretch}{0.82}
\resizebox*{\textwidth}{\textheight}{%
\newcolumntype{Y}{>{\centering\arraybackslash}m{0.7cm}}
\begin{tabular}{l|>{\columncolor{gray!20}\centering\arraybackslash}m{0.7cm}>{\columncolor{blue!12}\centering\arraybackslash}m{0.7cm}YYYYYY|>{\columncolor{gray!20}\centering\arraybackslash}m{0.7cm}>{\columncolor{blue!12}\centering\arraybackslash}m{0.7cm}YYYYYY}
\toprule
 & \multicolumn{8}{|c|}{\textbf{Llama-2-7b}} & \multicolumn{8}{c}{\textbf{Llama-2-13b}} \\
Task & Oracle & LW & LR & Adapter Soup & LoGo (Ent) & LoGo (Norm) & qa-FLoRA & spectR & Oracle & LW & LR & Adapter Soup & LoGo (Ent) & LoGo (Norm) & qa-FLoRA & spectR \\
\midrule
\midrule
\multicolumn{17}{l}{\textbf{Struct to Text}} \\
web\_nlg\_en Rouge-1 & 69.5 & 30.1 & 28.3 & \underline{43.5} & 26.0 & 38.4 & \textbf{45.2} & 42.0 & 72.3 & 42.9 & 40.3 & \textbf{44.5} & 36.3 & 36.3 & \underline{43.5} & 31.8 \\
web\_nlg\_en Rouge-2 & 48.8 & 13.8 & 12.6 & \underline{23.0} & 10.9 & 20.8 & \textbf{23.9} & 20.5 & 52.6 & \textbf{23.1} & 21.1 & \underline{22.7} & 16.4 & 17.1 & 22.5 & 14.0 \\
web\_nlg\_en Rouge-l & 62.6 & 28.4 & 25.9 & \underline{39.6} & 24.3 & 34.6 & \textbf{41.3} & 38.1 & 65.9 & 39.0 & 37.0 & \textbf{40.0} & 33.1 & 33.0 & \underline{39.5} & 30.6 \\
dart Rouge-1 & 71.4 & 51.4 & 51.8 & \textbf{56.9} & 47.8 & 48.4 & \underline{54.9} & 52.3 & 68.0 & \textbf{55.6} & 53.0 & 49.7 & 51.1 & \underline{54.4} & \underline{54.4} & 47.5 \\
dart Rouge-2 & 48.8 & 27.4 & 29.0 & \underline{30.8} & 26.3 & 25.1 & \textbf{31.3} & 26.4 & 44.9 & \textbf{31.5} & \underline{29.9} & 29.5 & 26.8 & 29.4 & 29.2 & 24.7 \\
dart Rouge-l & 64.4 & 45.6 & 46.5 & \textbf{50.7} & 43.8 & 43.9 & \underline{49.2} & 47.2 & 61.7 & \underline{48.7} & 46.9 & 44.2 & 46.1 & \textbf{48.9} & 48.6 & 42.9 \\
e2e\_nlg Rouge-1 & 65.5 & 55.2 & \textbf{56.2} & \underline{55.7} & 45.5 & 50.1 & 54.2 & \underline{55.7} & 66.4 & 56.0 & 49.8 & \textbf{57.3} & 55.8 & \underline{56.4} & 56.0 & 54.7 \\
e2e\_nlg Rouge-2 & 39.3 & 29.1 & \underline{30.0} & \textbf{30.4} & 23.8 & 26.0 & 29.0 & 28.3 & 38.7 & 29.7 & 23.6 & \underline{30.0} & 28.5 & 29.1 & \textbf{30.3} & 28.3 \\
e2e\_nlg Rouge-l & 55.8 & 45.9 & 46.8 & \textbf{47.8} & 39.6 & 43.0 & 46.8 & \underline{47.5} & 56.5 & 45.6 & 42.1 & 46.6 & 46.3 & 46.5 & \textbf{47.2} & \underline{46.7} \\
common\_gen Rouge-1 & 47.2 & 27.2 & 25.1 & 28.0 & 24.5 & 27.1 & \underline{28.1} & \textbf{28.8} & 48.4 & \underline{31.1} & 24.8 & 28.0 & 29.9 & \textbf{32.0} & 29.3 & 23.3 \\
common\_gen Rouge-2 & 18.6 & 7.0 & 7.1 & \textbf{8.7} & 6.4 & \textbf{8.7} & 8.4 & \underline{8.6} & 19.4 & \underline{11.1} & 8.0 & 10.8 & 11.0 & \textbf{11.8} & 8.1 & 6.6 \\
common\_gen Rouge-l & 43.3 & 24.2 & 21.7 & \underline{24.5} & 23.0 & 24.4 & 23.4 & \textbf{25.1} & 44.2 & \underline{28.7} & 22.9 & 26.3 & 26.8 & \textbf{28.8} & 24.6 & 20.2 \\
\midrule
\multicolumn{17}{l}{\textbf{Translation}} \\
wmt16\_translate\_tren BLEU & 3.4 & \textbf{2.6} & 2.3 & 2.4 & \underline{2.5} & 1.5 & \underline{2.5} & 1.9 & 2.4 & \textbf{3.3} & 2.0 & \textbf{3.3} & 2.4 & 2.4 & 2.5 & \underline{3.1} \\
wmt16\_translate\_deen BLEU & 18.4 & \underline{19.2} & \textbf{19.5} & 18.6 & 18.0 & 16.6 & 19.0 & 18.6 & 20.6 & 19.3 & 19.2 & 18.3 & 20.1 & \underline{21.5} & 18.9 & \textbf{22.6} \\
wmt16\_translate\_ruen BLEU & 10.5 & 8.8 & 7.6 & \textbf{11.0} & 8.8 & 7.5 & \underline{9.4} & 8.7 & 11.6 & \underline{11.7} & 8.4 & 10.5 & \textbf{13.0} & \underline{11.7} & 9.4 & 10.2 \\
wmt16\_translate\_fien BLEU & 6.9 & 7.3 & \textbf{8.0} & \underline{7.5} & \underline{7.5} & 5.4 & 7.4 & 6.6 & 6.7 & 8.5 & 3.3 & \underline{9.7} & \textbf{10.3} & 9.4 & 7.0 & 9.5 \\
wmt16\_translate\_roen BLEU & 14.0 & 10.1 & 10.8 & \textbf{13.6} & 9.7 & 9.5 & \underline{12.9} & 10.6 & 14.0 & 11.6 & \underline{13.9} & 10.6 & 11.1 & 10.7 & 12.0 & \textbf{15.4} \\
wmt14\_enfr BLEU & 16.3 & 16.9 & 16.1 & \textbf{18.2} & \underline{17.6} & 16.3 & \underline{17.6} & 15.4 & 16.4 & 19.3 & 18.1 & 19.2 & \underline{20.6} & \textbf{20.7} & 17.4 & 20.4 \\
wmt16\_translate\_csen BLEU & 9.9 & 7.2 & \underline{7.5} & 6.2 & \textbf{7.7} & 6.0 & 6.0 & 7.4 & 6.9 & \underline{10.9} & 7.1 & \textbf{11.1} & 10.2 & 10.5 & 5.7 & 8.3 \\
para\_crawl\_enes BLEU & 24.8 & 20.0 & 18.4 & \underline{22.3} & 20.9 & 20.3 & \textbf{24.4} & 22.2 & 24.6 & 24.4 & 18.5 & 24.3 & \underline{25.9} & 24.9 & 24.5 & \textbf{26.3} \\
\midrule
\multicolumn{17}{l}{\textbf{COMMONSENSE}} \\
story\_cloze EM & 72.0 & \underline{82.0} & 80.0 & 78.0 & \underline{82.0} & 80.0 & 70.0 & \textbf{86.0} & 96.0 & \textbf{100.0} & 92.0 & 90.0 & \underline{94.0} & \underline{94.0} & 72.0 & 86.0 \\
piqa EM & 44.0 & \textbf{42.0} & \underline{40.0} & 36.0 & \underline{40.0} & \underline{40.0} & 36.0 & \textbf{42.0} & 48.0 & \textbf{48.0} & 36.0 & \underline{44.0} & 42.0 & \underline{44.0} & 38.0 & 36.0 \\
copa EM & 86.0 & 74.0 & 72.0 & 72.0 & 74.0 & 72.0 & \textbf{78.0} & \underline{76.0} & 78.0 & \textbf{84.0} & \underline{78.0} & 70.0 & \underline{78.0} & \underline{78.0} & 76.0 & 70.0 \\
hellaswag EM & 50.0 & 36.0 & 30.0 & 14.0 & \textbf{42.0} & \underline{40.0} & 22.0 & 38.0 & 42.0 & \underline{48.0} & \textbf{52.0} & 34.0 & 42.0 & 38.0 & 22.0 & 36.0 \\
\midrule
\multicolumn{17}{l}{\textbf{sentiment}} \\
sst2 EM & 98.0 & \underline{92.0} & \underline{92.0} & 80.0 & 84.0 & 90.0 & 84.0 & \textbf{94.0} & 98.0 & \textbf{98.0} & \underline{94.0} & 84.0 & 90.0 & 90.0 & 84.0 & 70.0 \\
yelp\_polarity\_reviews EM & 98.0 & \underline{96.0} & \textbf{98.0} & \textbf{98.0} & 92.0 & 92.0 & 94.0 & 94.0 & 98.0 & \textbf{98.0} & \underline{94.0} & 90.0 & \underline{94.0} & \underline{94.0} & \underline{94.0} & 86.0 \\
imdb\_reviews EM & 96.0 & \underline{98.0} & \textbf{100.0} & 76.0 & 74.0 & 76.0 & 62.0 & 82.0 & 98.0 & \textbf{100.0} & \underline{98.0} & 74.0 & 80.0 & 80.0 & 62.0 & 76.0 \\
sentiment140 EM & 68.0 & \underline{68.0} & \textbf{70.0} & 66.0 & 64.0 & 60.0 & 62.0 & 66.0 & 70.0 & 68.0 & 64.0 & \underline{76.0} & 72.0 & 70.0 & 64.0 & \textbf{78.0} \\
\midrule
\multicolumn{17}{l}{\textbf{READING Comp.}} \\
multirc EM & 68.0 & \underline{46.0} & \textbf{50.0} & 42.0 & 34.0 & 34.0 & 32.0 & 38.0 & 48.0 & \textbf{52.0} & 36.0 & \underline{40.0} & 24.0 & 24.0 & 32.0 & 24.0 \\
squad\_v2 EM & 62.0 & 24.0 & 22.0 & 18.0 & \textbf{28.0} & \underline{26.0} & 16.0 & 18.0 & 44.0 & \underline{24.0} & 18.0 & \textbf{30.0} & 20.0 & 20.0 & 22.0 & 18.0 \\
squad\_v1 EM & 70.0 & \underline{74.0} & 70.0 & 56.0 & \textbf{76.0} & \underline{74.0} & 60.0 & 66.0 & 68.0 & \textbf{68.0} & 60.0 & 58.0 & \underline{64.0} & 62.0 & 62.0 & 56.0 \\
openbookqa EM & 82.0 & 66.0 & 64.0 & 66.0 & \textbf{70.0} & \underline{68.0} & \underline{68.0} & \textbf{70.0} & 84.0 & \underline{76.0} & \textbf{78.0} & 70.0 & 74.0 & 74.0 & 68.0 & 66.0 \\
bool\_q EM & 84.0 & \textbf{82.0} & 72.0 & 68.0 & 78.0 & 76.0 & 72.0 & \underline{80.0} & 76.0 & \underline{84.0} & 82.0 & 82.0 & \textbf{88.0} & \textbf{88.0} & 74.0 & 80.0 \\
drop EM & 40.0 & \underline{14.0} & \underline{14.0} & 10.0 & \textbf{16.0} & \textbf{16.0} & 12.0 & 10.0 & 28.0 & \textbf{24.0} & \underline{22.0} & 16.0 & \underline{22.0} & \underline{22.0} & 12.0 & 10.0 \\
\midrule
\multicolumn{17}{l}{\textbf{CLOSE-BOOK QA}} \\
natural\_questions EM & 18.0 & 14.0 & \underline{16.0} & 10.0 & \textbf{18.0} & \underline{16.0} & 12.0 & \textbf{18.0} & 20.0 & \textbf{22.0} & 8.0 & 16.0 & 18.0 & \underline{20.0} & 16.0 & 12.0 \\
arc\_easy EM & 50.0 & 72.0 & 66.0 & 56.0 & 68.0 & \textbf{80.0} & 60.0 & \underline{78.0} & 92.0 & \underline{92.0} & 72.0 & 88.0 & \textbf{94.0} & \textbf{94.0} & 56.0 & \underline{92.0} \\
arc\_challenge EM & 46.0 & \underline{44.0} & \underline{44.0} & 34.0 & 38.0 & \underline{44.0} & 38.0 & \textbf{50.0} & 66.0 & \textbf{72.0} & 60.0 & 60.0 & \underline{66.0} & \underline{66.0} & 42.0 & 56.0 \\
trivia\_qa EM & 64.0 & \textbf{58.0} & 50.0 & 46.0 & \textbf{58.0} & \textbf{58.0} & 48.0 & \underline{56.0} & 48.0 & \underline{66.0} & 44.0 & 64.0 & \textbf{70.0} & \textbf{70.0} & 46.0 & 54.0 \\
\midrule
\multicolumn{17}{l}{\textbf{COREFERENCE}} \\
definite\_pronoun\_resolution EM & 56.0 & \textbf{72.0} & \textbf{72.0} & 56.0 & \underline{68.0} & \textbf{72.0} & 62.0 & \underline{68.0} & 88.0 & \textbf{70.0} & \textbf{70.0} & \underline{68.0} & \textbf{70.0} & \textbf{70.0} & 54.0 & 62.0 \\
wsc EM & 50.0 & \textbf{56.0} & \underline{54.0} & 36.0 & 52.0 & 52.0 & 50.0 & 50.0 & 64.0 & \textbf{50.0} & \underline{48.0} & 42.0 & 46.0 & 46.0 & 46.0 & 40.0 \\
\midrule
\multicolumn{17}{l}{\textbf{READ. COMP. W/ COMMONSENSE}} \\
cosmos\_qa EM & 72.0 & 46.0 & 48.0 & 28.0 & 46.0 & \textbf{56.0} & 28.0 & \underline{52.0} & 86.0 & \underline{80.0} & 24.0 & 72.0 & \underline{80.0} & \textbf{84.0} & 26.0 & 58.0 \\
record EM & 70.0 & \textbf{32.0} & 26.0 & 8.0 & \textbf{32.0} & \underline{28.0} & 14.0 & 24.0 & 76.0 & \textbf{34.0} & 26.0 & 14.0 & \underline{28.0} & 26.0 & 10.0 & \textbf{34.0} \\
\midrule
\multicolumn{17}{l}{\textbf{PARAPHRASE}} \\
paws\_wiki EM & 90.0 & \textbf{64.0} & 56.0 & 44.0 & 46.0 & \underline{58.0} & 46.0 & \textbf{64.0} & 88.0 & \textbf{70.0} & \textbf{70.0} & 46.0 & \underline{52.0} & \underline{52.0} & 42.0 & 44.0 \\
glue\_qqp EM & 74.0 & \textbf{66.0} & \underline{60.0} & \underline{60.0} & 50.0 & 54.0 & 52.0 & 56.0 & 88.0 & \textbf{76.0} & \textbf{76.0} & 44.0 & 46.0 & 44.0 & \underline{52.0} & \underline{52.0} \\
glue\_mrpc EM & 60.0 & \textbf{64.0} & \underline{60.0} & \underline{60.0} & 50.0 & 44.0 & 58.0 & 56.0 & 82.0 & \textbf{64.0} & 58.0 & 60.0 & \underline{62.0} & \underline{62.0} & 60.0 & \underline{62.0} \\
stsb EM & 36.0 & \textbf{26.0} & \underline{24.0} & 12.0 & 16.0 & 16.0 & 14.0 & \textbf{26.0} & 40.0 & \textbf{28.0} & 16.0 & 16.0 & 10.0 & 6.0 & 12.0 & \underline{20.0} \\
\midrule
\multicolumn{17}{l}{\textbf{NLI}} \\
cb EM & 86.7 & \textbf{77.8} & 66.7 & 64.4 & \underline{75.6} & \underline{75.6} & 55.6 & 73.3 & 93.3 & \textbf{88.9} & \underline{86.7} & 75.6 & 80.0 & 80.0 & 57.8 & 60.0 \\
wnli EM & 72.0 & \textbf{58.0} & \underline{56.0} & 44.0 & 48.0 & 52.0 & 50.0 & \underline{56.0} & 72.0 & \textbf{62.0} & \underline{58.0} & 44.0 & 50.0 & 46.0 & 46.0 & 40.0 \\
mnli\_matched EM & 88.0 & \textbf{80.0} & 70.0 & \underline{72.0} & 58.0 & 62.0 & 58.0 & \underline{72.0} & 86.0 & \textbf{90.0} & \underline{88.0} & 78.0 & \underline{88.0} & \textbf{90.0} & 68.0 & 68.0 \\
anli\_r3 EM & 46.0 & 40.0 & \underline{42.0} & \underline{42.0} & 40.0 & 38.0 & \textbf{44.0} & \underline{42.0} & 46.0 & \underline{50.0} & 36.0 & 32.0 & \textbf{52.0} & \textbf{52.0} & 44.0 & 44.0 \\
anli\_r2 EM & 46.0 & \underline{40.0} & \textbf{46.0} & 36.0 & 38.0 & 36.0 & 36.0 & 38.0 & 56.0 & \textbf{50.0} & \underline{44.0} & 36.0 & 40.0 & 40.0 & 30.0 & 40.0 \\
anli\_r1 EM & 50.0 & 34.0 & 34.0 & \textbf{42.0} & 32.0 & \underline{38.0} & 36.0 & 36.0 & 70.0 & \textbf{54.0} & 36.0 & 44.0 & \underline{48.0} & \underline{48.0} & 38.0 & 30.0 \\
mnli\_mismatched EM & 92.0 & \underline{86.0} & 82.0 & 66.0 & 80.0 & 78.0 & 70.0 & \textbf{88.0} & 92.0 & \textbf{96.0} & 90.0 & 78.0 & \underline{92.0} & 90.0 & 78.0 & 78.0 \\
snli EM & 96.0 & \textbf{86.0} & 66.0 & 62.0 & \underline{72.0} & 64.0 & 58.0 & 58.0 & 92.0 & \textbf{98.0} & \underline{88.0} & 60.0 & 74.0 & 78.0 & 64.0 & 62.0 \\
qnli EM & 94.0 & 58.0 & 56.0 & \underline{60.0} & 46.0 & 56.0 & \textbf{64.0} & 48.0 & 78.0 & \textbf{72.0} & 66.0 & 64.0 & \underline{68.0} & \underline{68.0} & 62.0 & 48.0 \\
rte EM & 54.0 & 58.0 & 58.0 & \textbf{68.0} & 52.0 & 58.0 & \underline{64.0} & 60.0 & 82.0 & \underline{78.0} & 68.0 & \textbf{82.0} & 66.0 & 64.0 & 62.0 & 46.0 \\
\bottomrule
\end{tabular}
}
\endgroup
\caption{Per-task cross-task performance on the 48-task FLAN-v2 mixed-task benchmark.}
\label{tab:crosstasks}
\end{table*}

\section{Layer-wise Task Similarity Heatmaps}
\label{sec:task-similarity-heatmaps}

\begin{figure*}[t]
\centering

\begin{subfigure}[t]{0.48\textwidth}
    \centering
    \includegraphics[width=\linewidth]{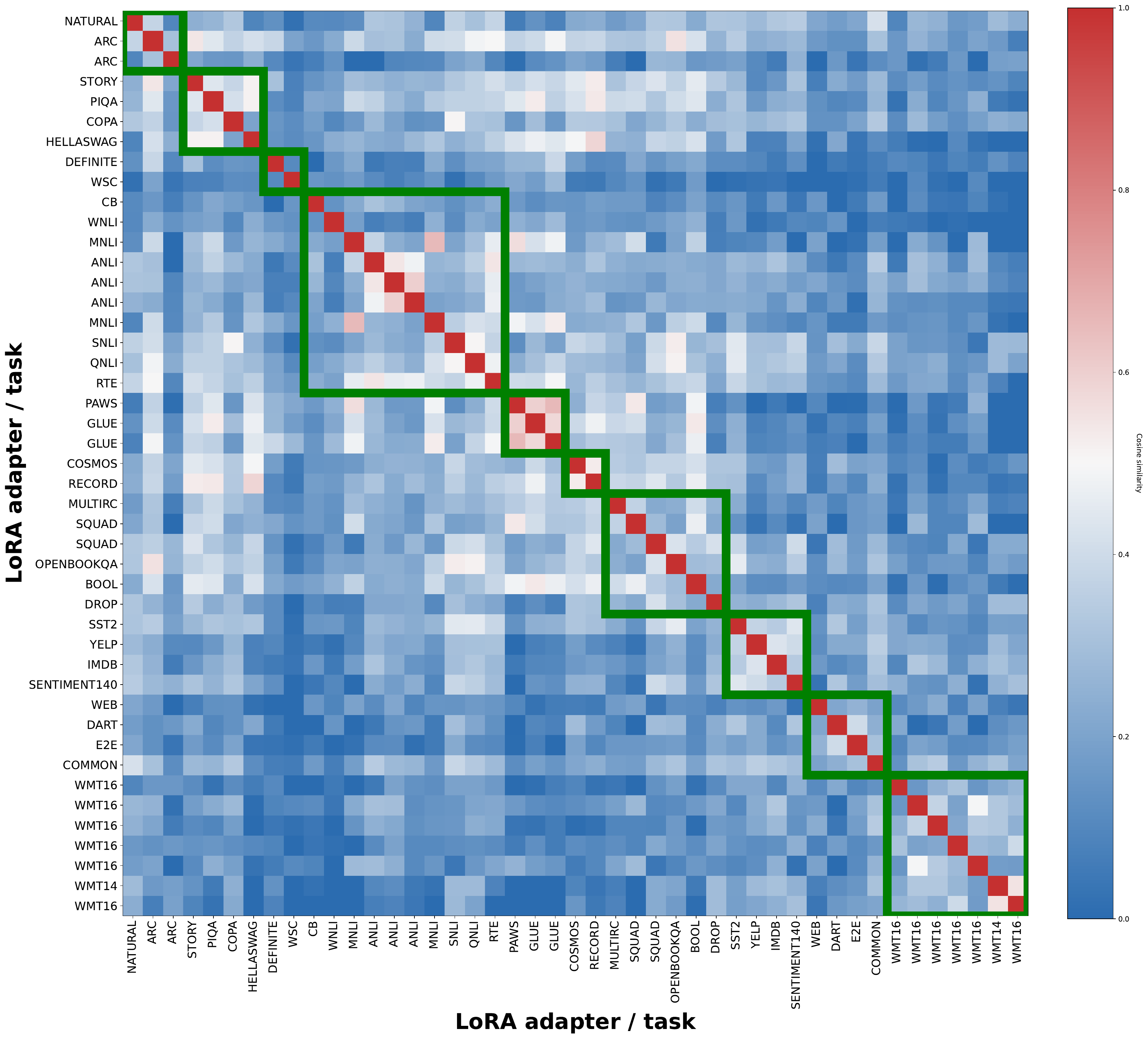}
    \caption{Layer 1}
\end{subfigure}
\hfill
\begin{subfigure}[t]{0.48\textwidth}
    \centering
    \includegraphics[width=\linewidth]{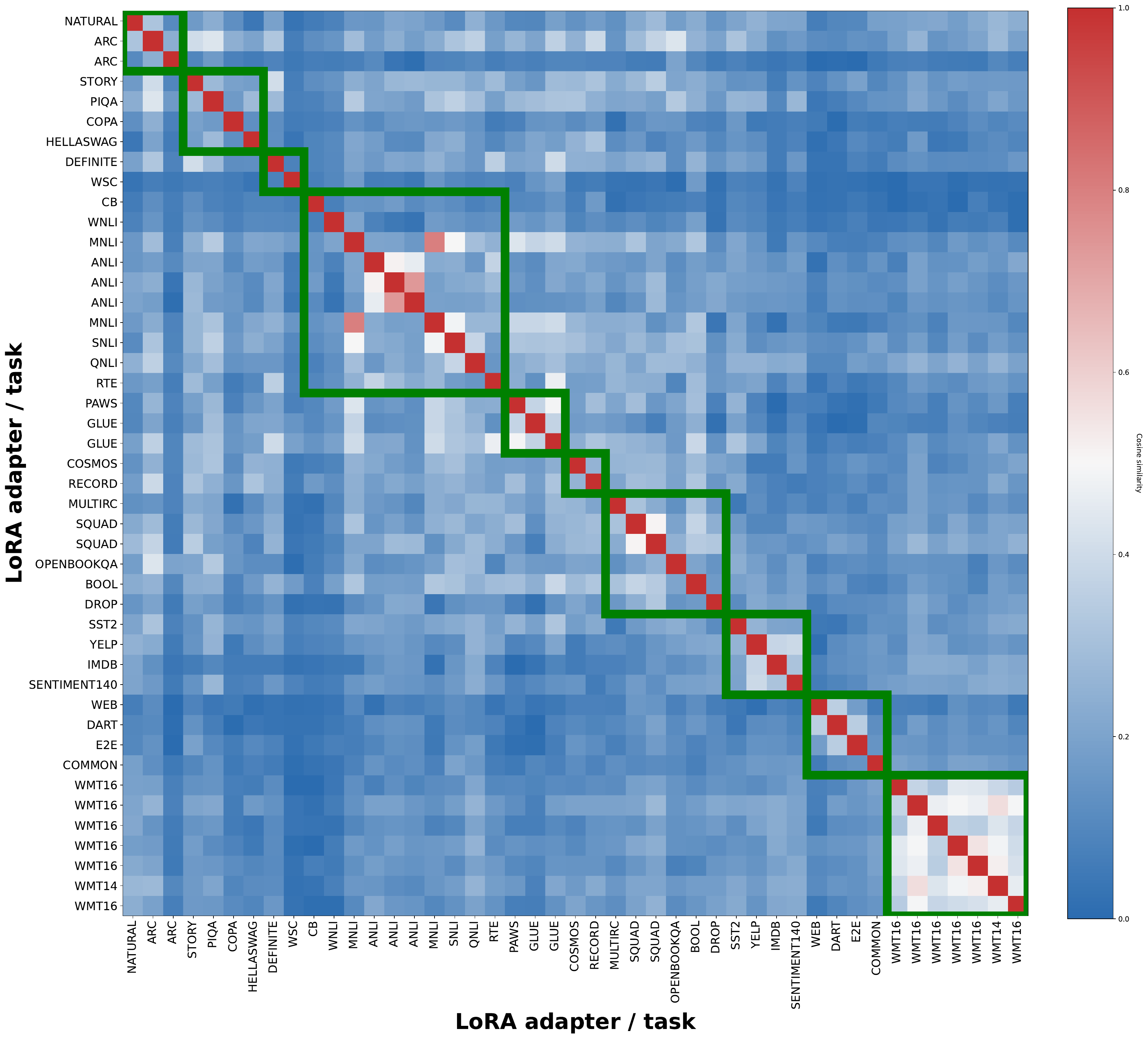}
    \caption{Layer 6}
\end{subfigure}

\vspace{0.5em}

\begin{subfigure}[t]{0.48\textwidth}
    \centering
    \includegraphics[width=\linewidth]{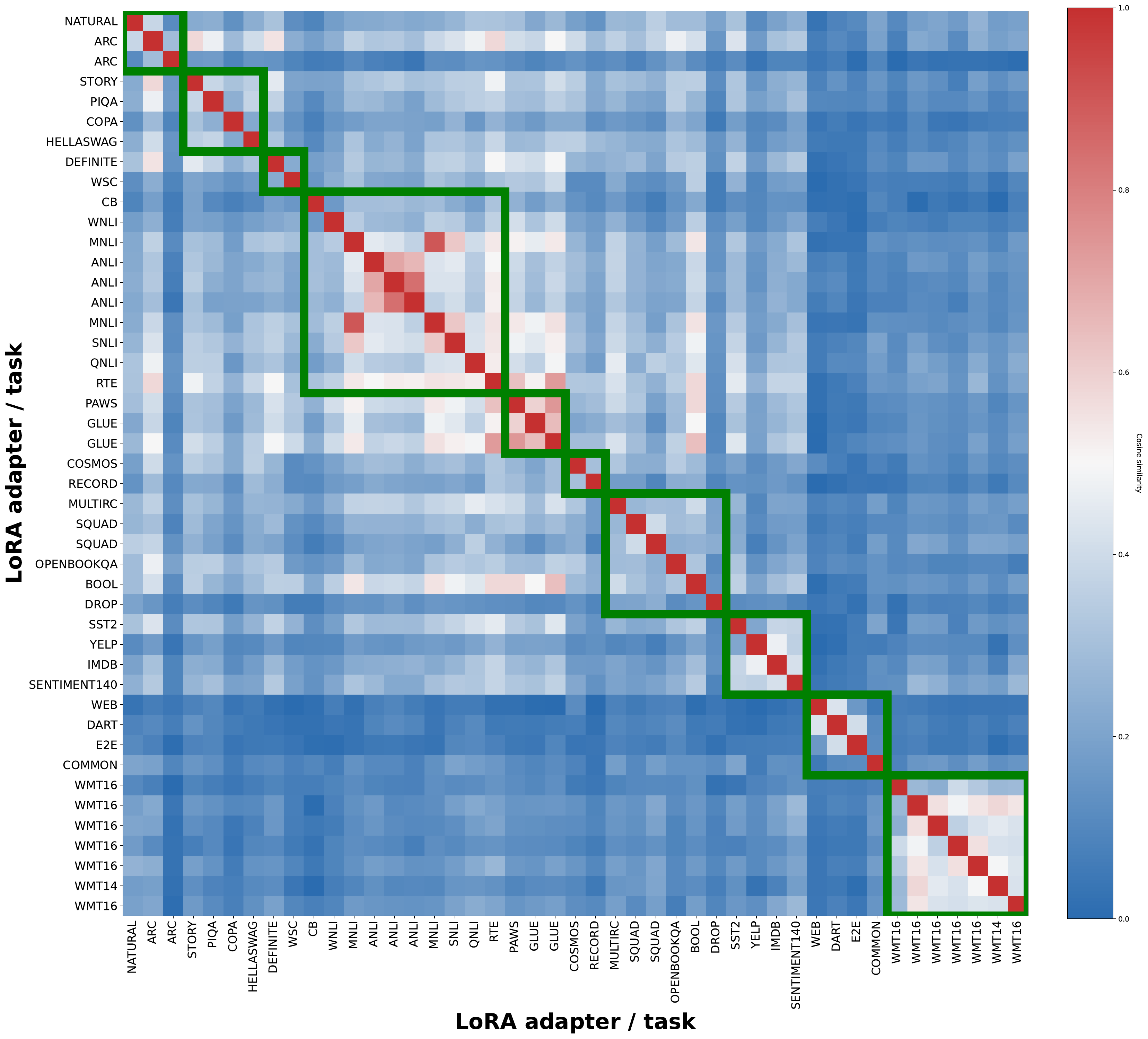}
    \caption{Layer 11}
\end{subfigure}
\hfill
\begin{subfigure}[t]{0.48\textwidth}
    \centering
    \includegraphics[width=\linewidth]{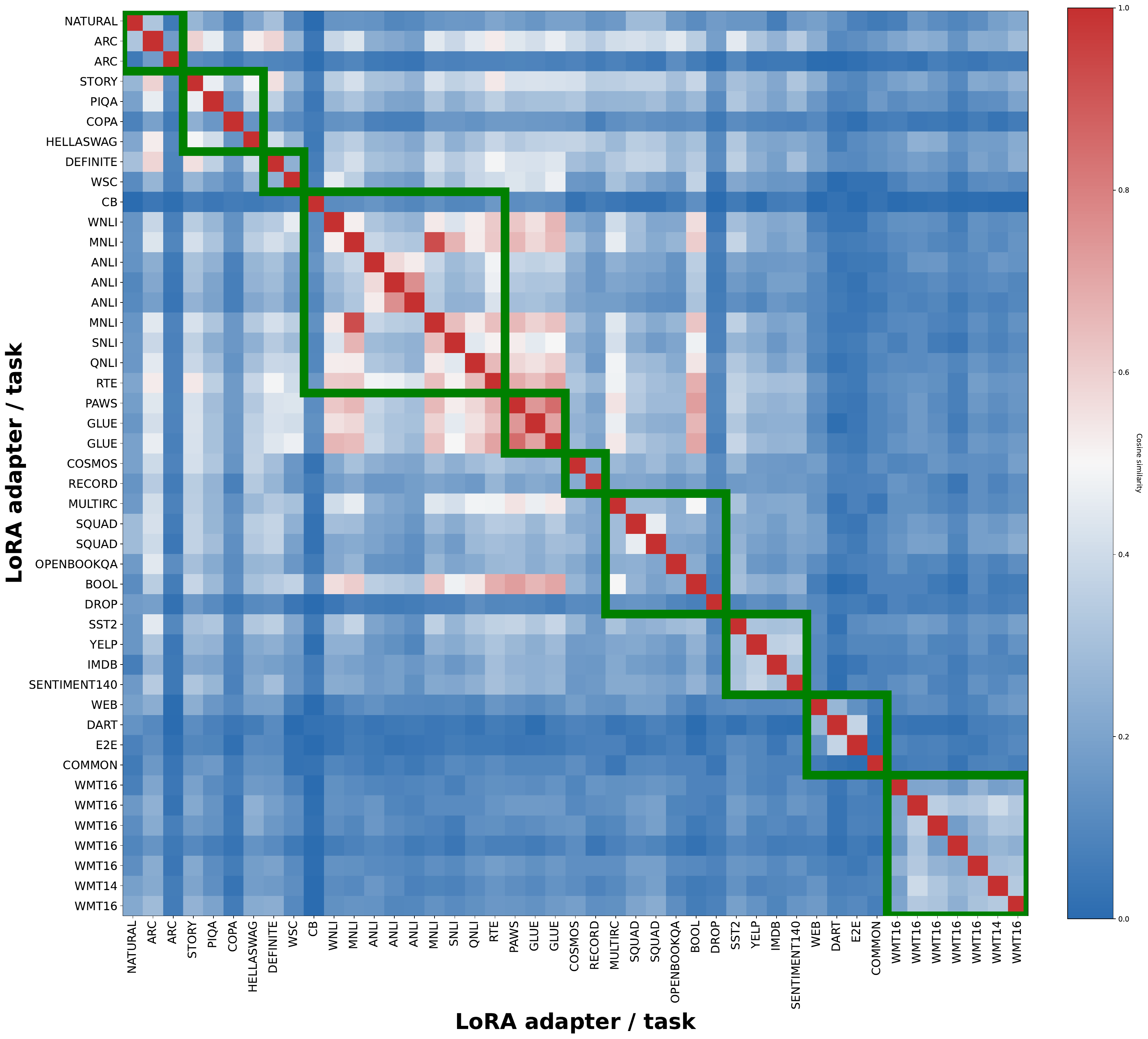}
    \caption{Layer 19}
\end{subfigure}

\vspace{0.5em}

\begin{subfigure}[t]{0.48\textwidth}
    \centering
    \includegraphics[width=\linewidth]{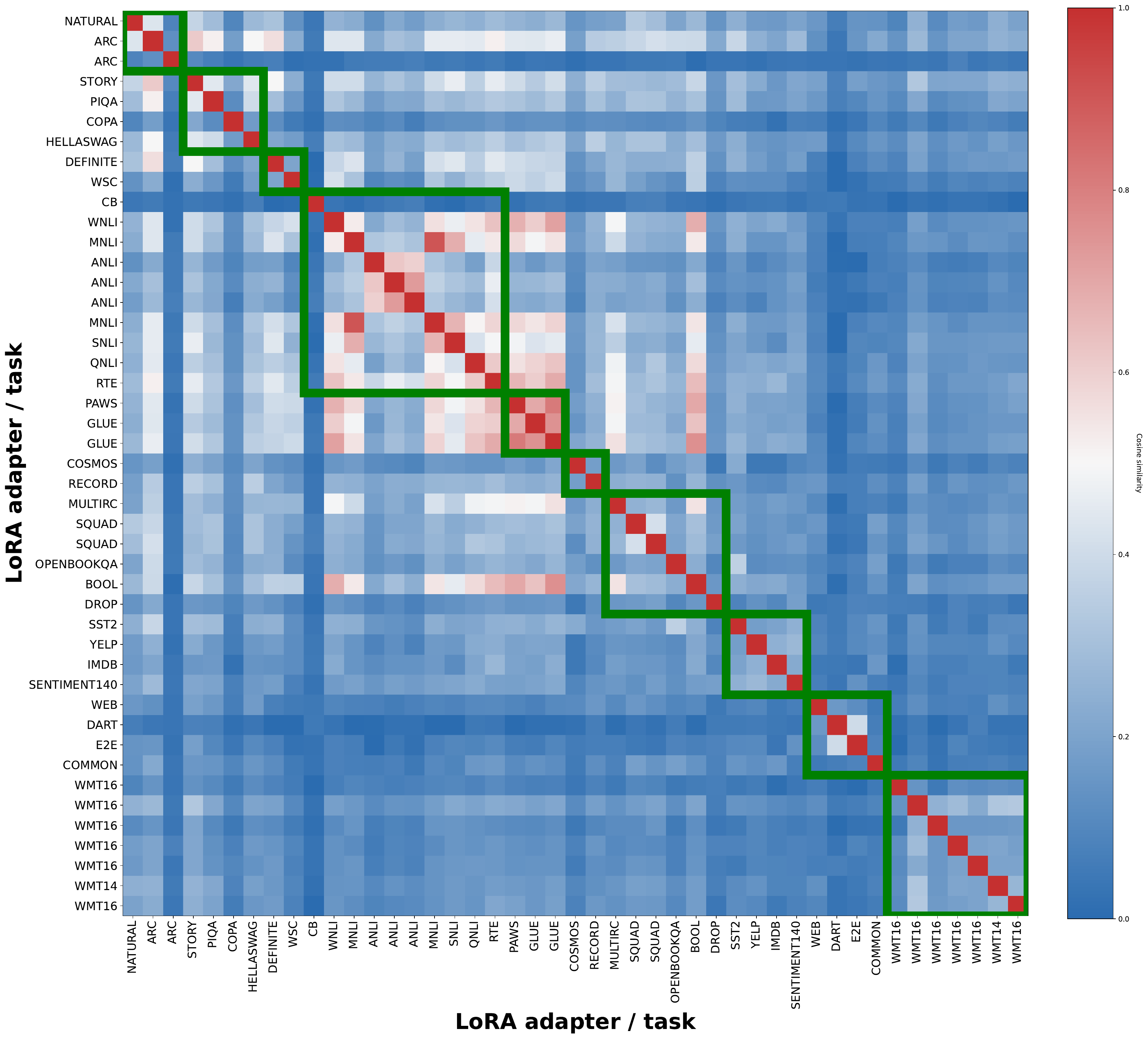}
    \caption{Layer 25}
\end{subfigure}
\hfill
\begin{subfigure}[t]{0.48\textwidth}
    \centering
    \includegraphics[width=\linewidth]{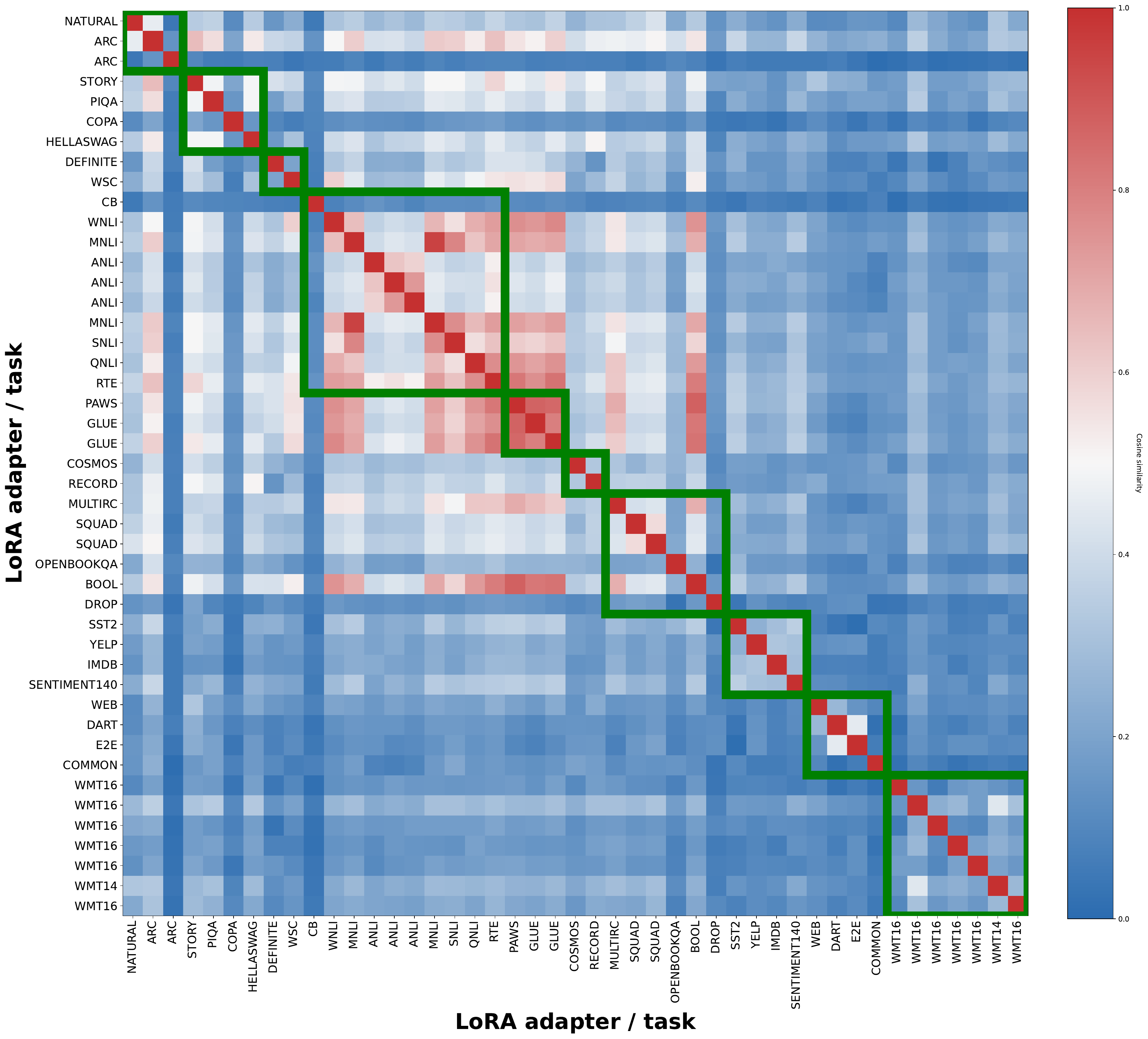}
    \caption{Layer 32}
\end{subfigure}

\caption{Layer-wise cosine similarity between LoRA adapter prototypes in the adapter-output space for LLaMA-2-7B adapters. Each heatmap shows a task-by-task prototype similarity matrix for one transformer layer. The changing similarity patterns across layers indicate that adapter relationships are depth-dependent, supporting independent layer-wise retrieval rather than a single global adapter ranking.}
\label{fig:heatmaps-7b}
\end{figure*}

\begin{figure*}[t]
\centering

\begin{subfigure}[t]{0.48\textwidth}
    \centering
    \includegraphics[width=\linewidth]{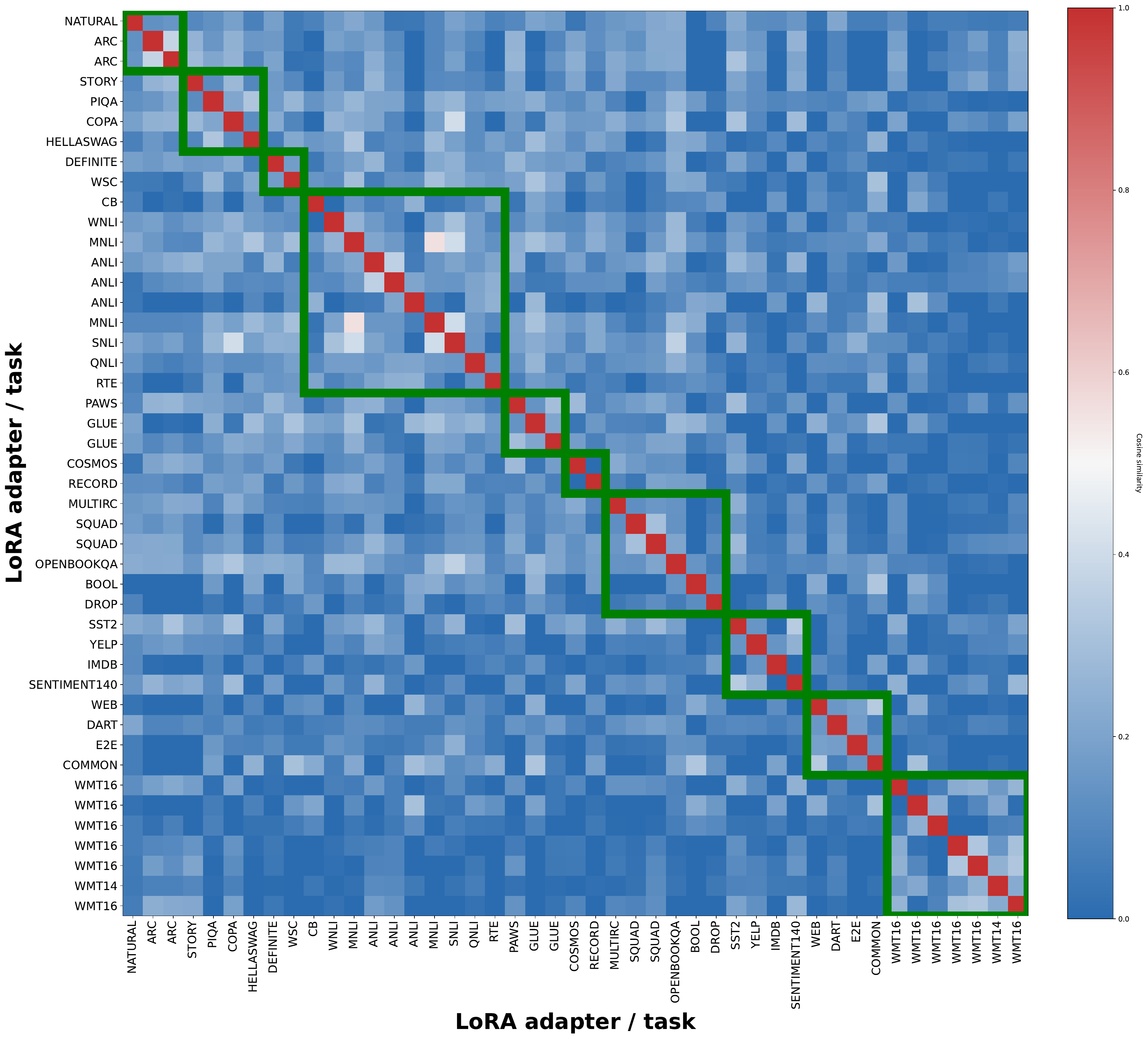}
    \caption{Layer 1}
\end{subfigure}
\hfill
\begin{subfigure}[t]{0.48\textwidth}
    \centering
    \includegraphics[width=\linewidth]{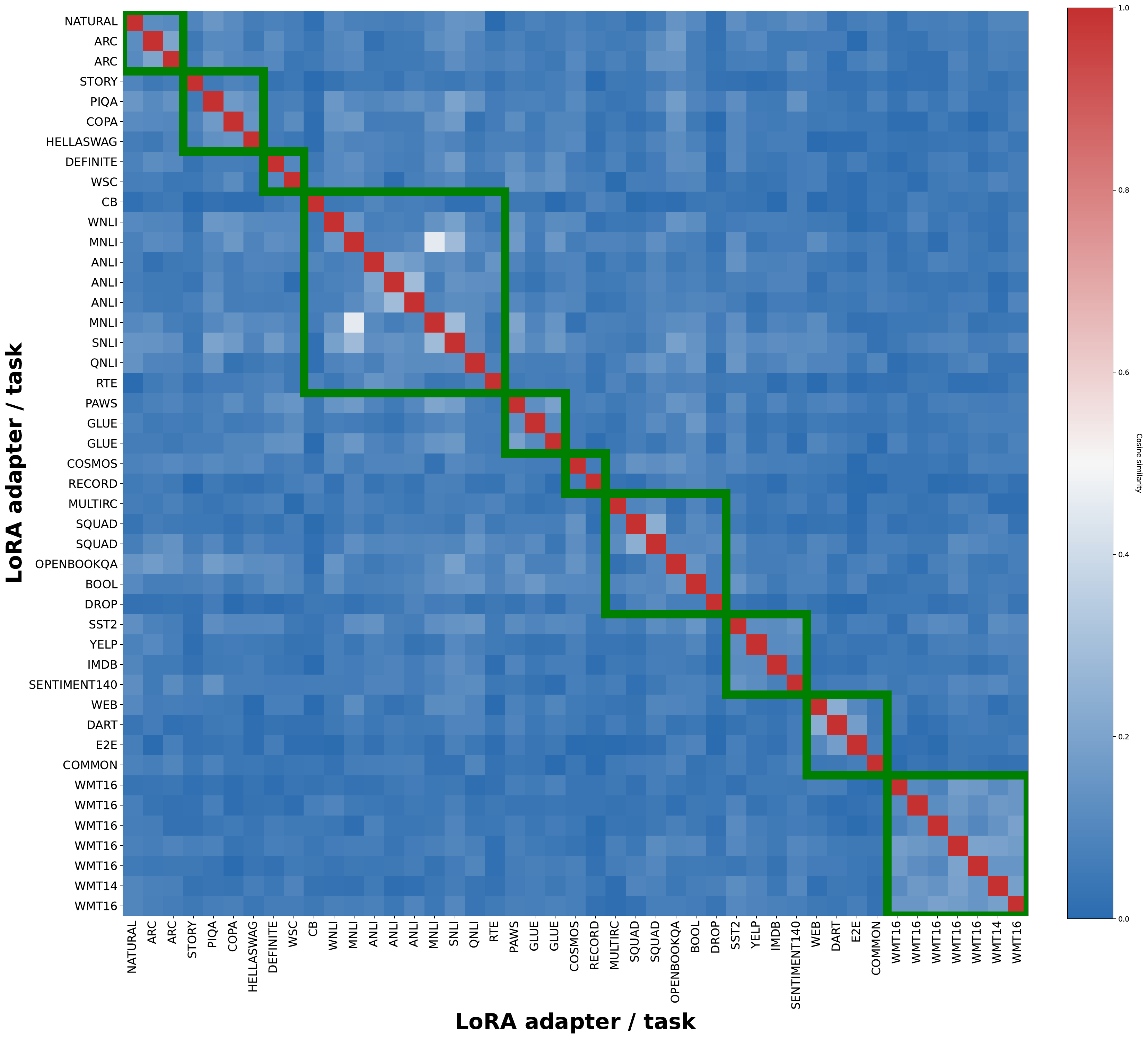}
    \caption{Layer 6}
\end{subfigure}

\vspace{0.5em}

\begin{subfigure}[t]{0.48\textwidth}
    \centering
    \includegraphics[width=\linewidth]{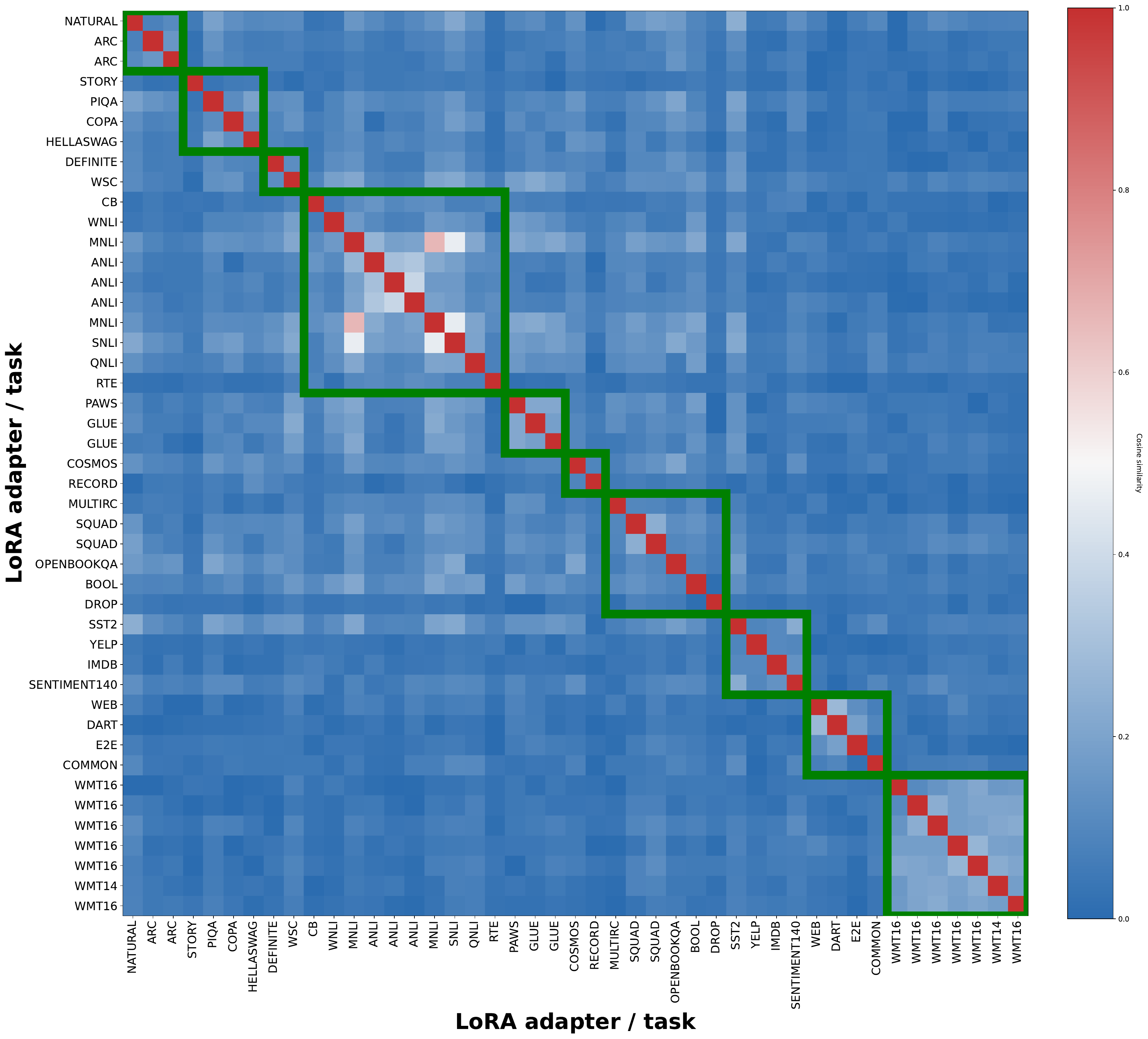}
    \caption{Layer 11}
\end{subfigure}
\hfill
\begin{subfigure}[t]{0.48\textwidth}
    \centering
    \includegraphics[width=\linewidth]{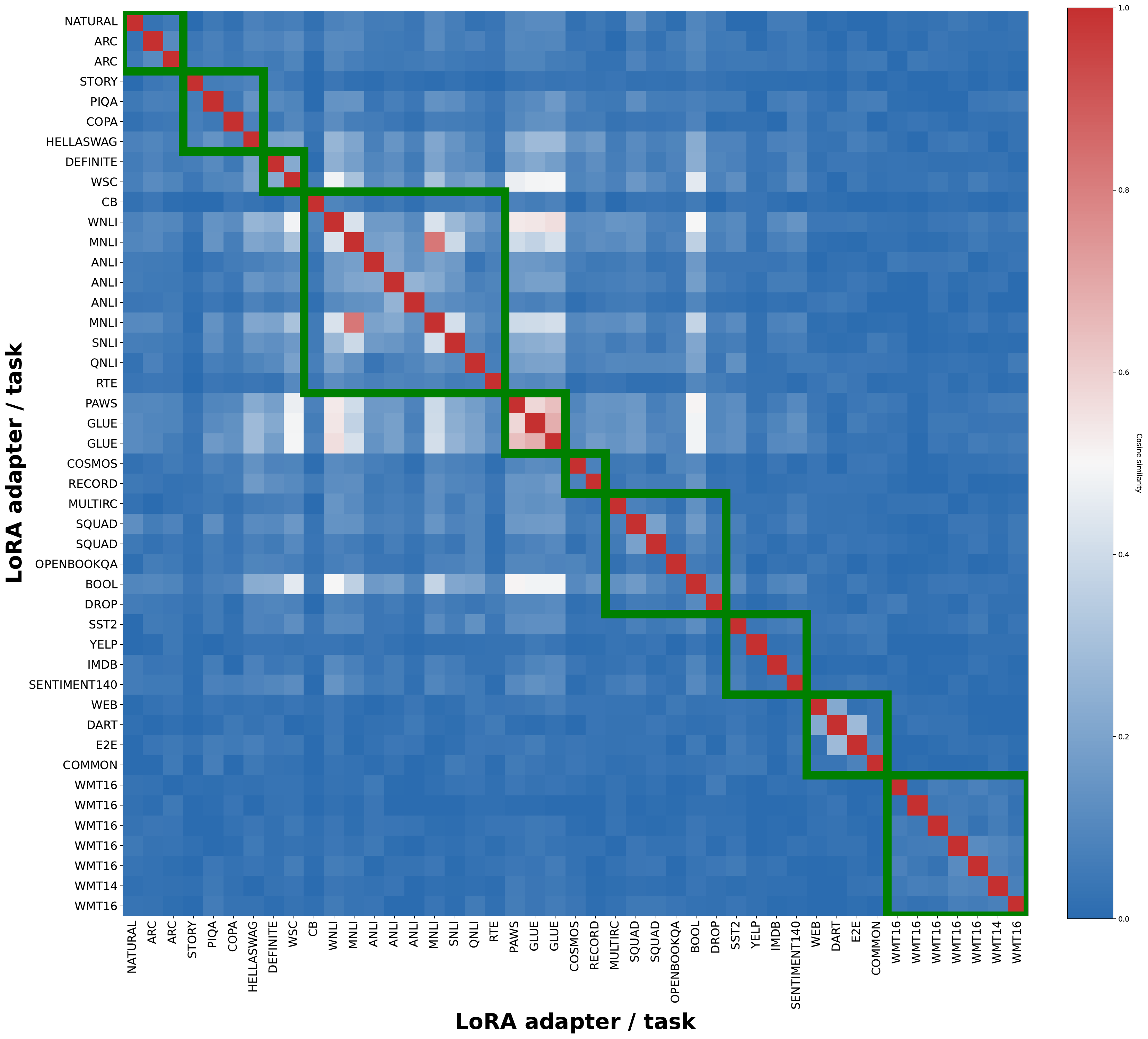}
    \caption{Layer 19}
\end{subfigure}

\vspace{0.5em}

\begin{subfigure}[t]{0.48\textwidth}
    \centering
    \includegraphics[width=\linewidth]{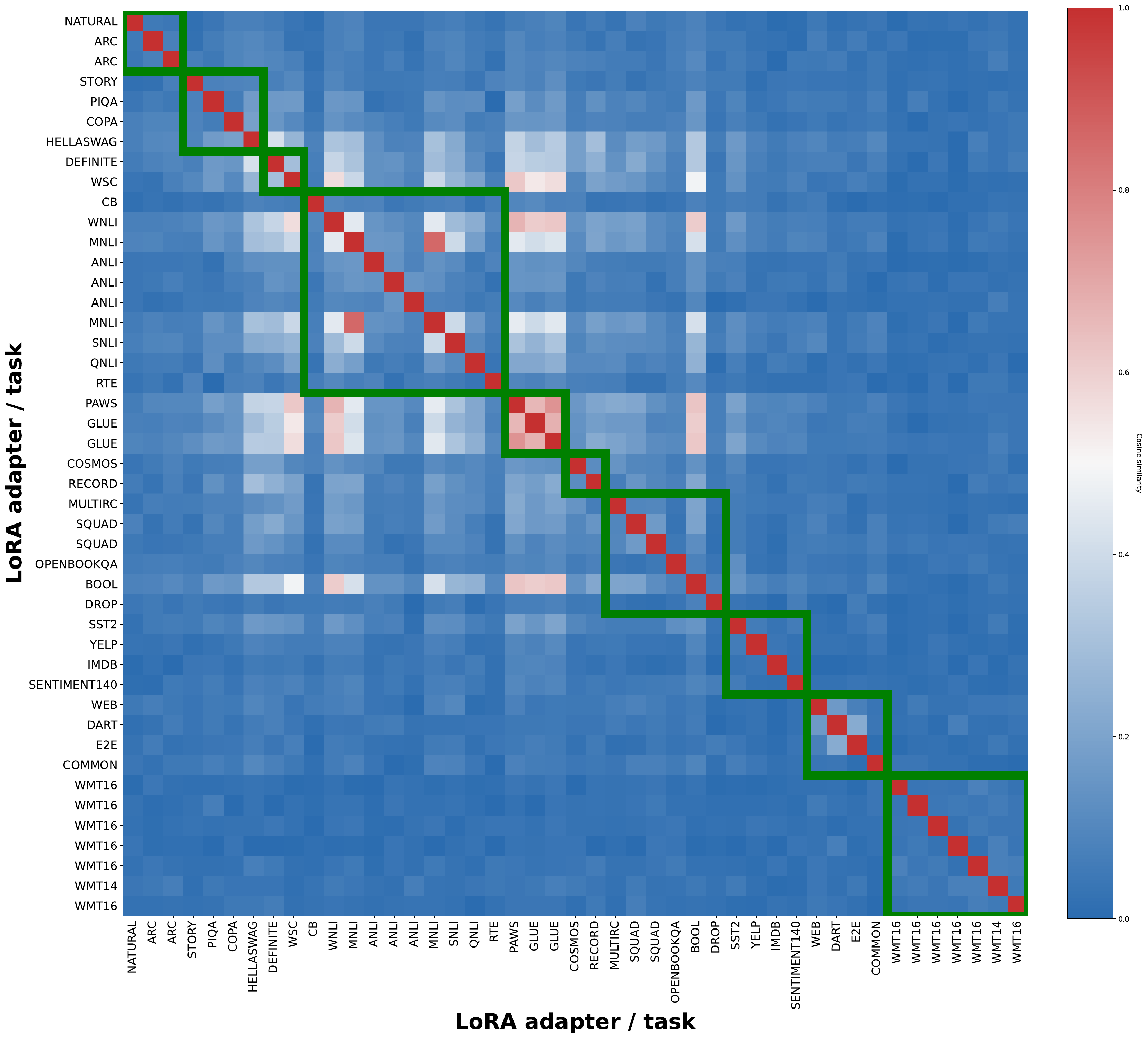}
    \caption{Layer 30}
\end{subfigure}
\hfill
\begin{subfigure}[t]{0.48\textwidth}
    \centering
    \includegraphics[width=\linewidth]{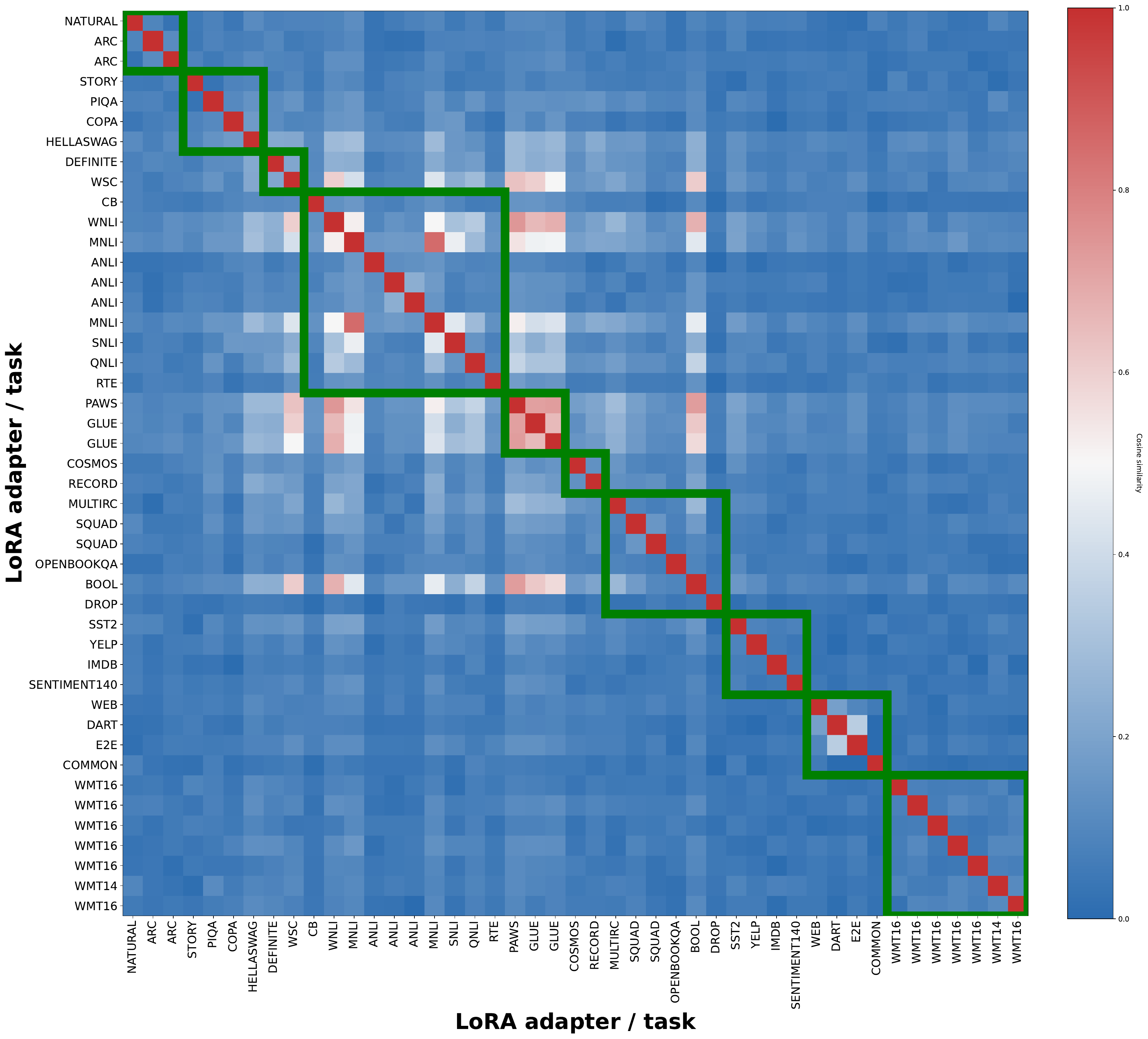}
    \caption{Layer 39}
\end{subfigure}

\caption{Layer-wise cosine similarity between LoRA adapter prototypes in the adapter-output space for LLaMA-2-13B adapters. Each heatmap shows a task-by-task prototype similarity matrix for one transformer layer. The changing similarity patterns across layers indicate that adapter relationships are depth-dependent, supporting independent layer-wise retrieval rather than a single global adapter ranking.}
\label{fig:heatmaps-13b}
\end{figure*}

Figures \ref{fig:heatmaps-7b} and \ref{fig:heatmaps-13b} visualise layer-wise task similarity in the LoRA adapter-output space for LLaMA-2-7B and LLaMA-2-13B, respectively. Each heatmap corresponds to the \texttt{q\_proj} target module of a single transformer block; the index denotes the block index $b \in \{1,\dots,32\}$, rather than the target-module index $\ell \in \{1,\dots,64\}$ defined in Section 3.1. The \texttt{v\_proj} modules exhibit the same qualitative behaviour and are omitted for space. For each selected layer, the analysis computes a task-by-task cosine similarity matrix from adapter prototypes. Each heatmap reports the degree to which task adapters produce aligned responses at a given model depth.

The heatmaps indicate that task relationships vary across layers. Some layers assign higher cosine values to related tasks, whereas other layers produce lower values or distinct grouping patterns. This layer-level variation indicates that adapter relationships depend on the representation geometry at each transformer layer. The observed structure supports the design of LoRAWeaver, which retrieves adapters independently per layer instead of selecting one fixed adapter set for the full network.

\end{document}